\documentclass[acmsmall, authorversion=true, nonacm=true]{acmart}

\usepackage{graphicx}
\usepackage{booktabs}
\usepackage{blindtext}
\usepackage{array}
\usepackage{pifont}
\usepackage{soul, color}
\usepackage{url}
\usepackage{caption}
\usepackage{subcaption}
\newcommand{\cmark}{\ding{51}}%
\newcommand{\xmark}{\ding{55}}%

\AtBeginDocument{%
  \providecommand\BibTeX{{%
    \normalfont B\kern-0.5em{\scshape i\kern-0.25em b}\kern-0.8em\TeX}}}

\begin{document}

\title[Explainable AI is Responsible AI]{Explainable AI is Responsible AI: How Explainability Creates Trustworthy and Socially Responsible Artificial Intelligence}
\author{Stephanie Baker}
\email{stephanie.baker@jcu.edu.au}
\orcid{0000-0003-0467-7791}
\affiliation{%
  \institution{James Cook University}
  \streetaddress{1/14-88 McGregor Road}
  \city{Cairns}
  \state{QLD}
  \country{Australia}
  \postcode{4878}
}

\author{Wei Xiang}
\affiliation{%
  \institution{La Trobe University}
  \streetaddress{Corner Plenty Road \& Kingsbury Drive}
  \city{Melbourne}
  \state{VIC}
  \country{Australia}
  \postcode{3083}}
\email{w.xiang@latrobe.edu.au}

\renewcommand{\shortauthors}{Baker and Xiang}

\begin{abstract}
Artificial intelligence (AI) has been clearly established as a technology with the potential to revolutionize fields from healthcare to finance - if developed and deployed responsibly. This is the topic of responsible AI, which emphasizes the need to develop trustworthy AI systems that minimize bias, protect privacy, support security, and enhance transparency and accountability. Explainable AI (XAI) has been broadly considered as a building block for responsible AI (RAI), with most of the literature considering it as a solution for improved transparency. This work proposes that XAI and responsible AI are significantly more deeply entwined. In this work, we explore state-of-the-art literature on RAI and XAI technologies. Based on our findings, we demonstrate that XAI can be utilized to ensure fairness, robustness, privacy, security, and transparency in a wide range of contexts. Our findings lead us to conclude that XAI is an essential foundation for every pillar of RAI.
\end{abstract}



\received{dd Month 2023}
\received[revised]{dd Month 2023}
\received[accepted]{dd Month 2023}

\maketitle

\section{Introduction}
In our increasingly data-driven society, artificial intelligence (AI) has become pervasive for decision-making in a wide range of fields. AI has immense potential to make positive change in fields ranging from agriculture \cite{Saiz-Rubio2020} to healthcare \cite{bakerAIOT}. However, as with any tool, AI can also cause great harm if developed and used improperly or inappropriately; regardless of whether this is done unintentionally or maliciously. The consequences of bad AI practice can range from bias affecting underprivileged or underrepresented groups \cite{Ensign2018, Dastin2019}, to financial harm \cite{Rinta-Kahila2022}, to physical and mental harm \cite{Norori2021, Laestadius2022}.

The need to develop AI in a way that benefits human life and societies has lead to the emergence of responsible AI (RAI). RAI is fundamentally the field of applying ethics to the development and utilization of AI, to ensure that AI systems and used for the good of humanity \cite{Anagnostou2022, Trocin2021}. As with all moral philosophy, there is no single consensus on what makes AI ethical and therefore responsible. However, efforts have been made to establish RAI characteristics, frameworks, and guidelines in academia \cite{Anagnostou2022, kaur2022, Li2023}, industry \cite{Microsoft2023, Google2023, Samsung2023, MetaAI2021, IBM2023}, and goverments and political bodies \cite{CSIRO2023, USGovRAI2022, ChinaRAI2022, EuropeanCommission2021, SaudiArabia2022, JapanMIST2022, UKGovernment2021, IndiaAI2022}. Frequently identified pillars of RAI across these diverse sources include fairness, robustness, transparency, accountability, privacy, and safety.

Explainability is also considered as a pillar of RAI by many works \cite{BarredoArrieta2020, Li2023, kaur2022}, often connected to transparency. Explainability is broadly considered by the field of explainable AI (XAI), which is focused on providing humans with insight into the reasoning of AI models during their decision-making processes. Researchers in the XAI domain have developed methods for producing explanations that include text-based \cite{Chun2022, Arous2021}, visual \cite{Selvaraju2017, shap}, and feature importance \cite{shap, lime} methods.

\begin{figure}[h]
    \centering
    \includegraphics[width=0.9\linewidth]{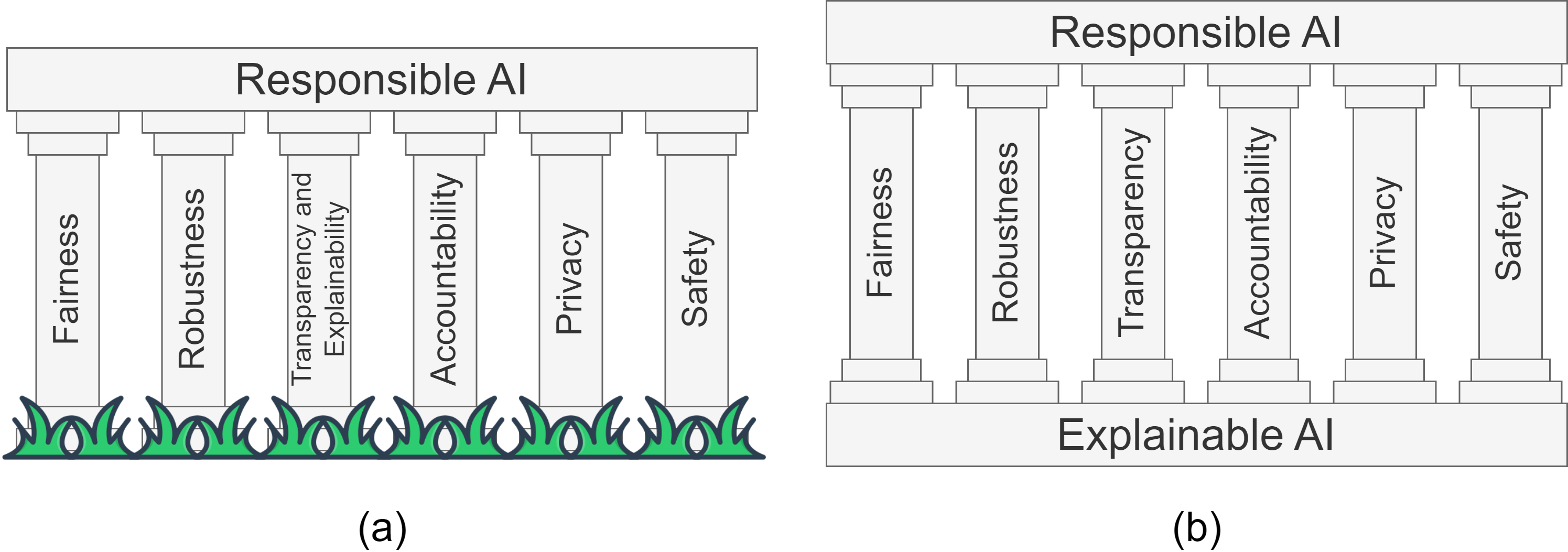}
    \caption{Comparison of RAI frameworks (a) as commonly defined in the literature with explainability as a pillar, and (b) as proposed by this work with explainability as the foundational concept.}
    \label{fig:pillars}
\end{figure}

Given that XAI techniques improve transparency \cite{Shin2021,Alam2021}, it is understandable that some researchers have classed explainability as a pillar of RAI. However, our review of the literature finds that considering XAI as one discrete pillar of RAI, separate to all others, is insufficient to capture the potential impact of explainability on RAI. In this review, we find substantial evidence that XAI should instead be considered as the foundation of RAI; this key difference is shown in Fig. \ref{fig:pillars}.

In our review, XAI is shown to support assessment of fairness \cite{Miron2021, Nakao2022, Pradhan2022}, robustness \cite{SharmaCERTIFAI, malik2022xai}, and safety \cite{Atif2022, Sado2023, Jia2022xaihealthcare}, thus supporting responsible AI operation and enabling improvement as needed. The transparency offered by XAI is shown to improve interpretability and understability of AI through human evaluation in several studies \cite{Aechtner2022, Ding2022, Singla2023}, which in turn supports accountability for decision making in critical sectors such as finance \cite{Bucker2022} and law \cite{Padovan2023}. Finally, it is illustrated that XAI can be combined with privacy-preserving techniques to ensure that the requirement of privacy is met without compromising on other RAI requirements \cite{Bozorgpanah2022, Gaudio2023, Raza2022,Saldanha2022}.

\subsection{Original Contributions}
The literature contains several high-quality reviews on the topics of XAI and RAI; however, these are typically considered as separate topics. In this work, we fill a key gap in the literature by illustrating that XAI is foundational to RAI. The key contributions of this work are as follows:

\begin{itemize}
    \item We conduct a broad-scoping review of the literature on both XAI and RAI, highlighting the technologies, principals, and frameworks developed by previous works.
    \item We propose a novel framework that considers explainable AI as the foundation of responsible AI. This is in contrast to previous works, which have considered XAI as a single pillar of RAI.
    \item We demonstrate that XAI is foundational to the RAI principles of fairness, robustness, transparency, accountability, privacy and safety through rigorous exploration of the XAI literature.
    \item Building upon our findings, we illustrate real-world use cases where XAI directly supports RAI in applications within the key fields of generative AI, healthcare, and transportation. 
\end{itemize}

\subsection{Comparison to Other Works}
The field of XAI has been extremely active in recent years, and responsible AI has also become a hot topic. As such, several high-quality surveys have examined these two topics. However, the concept of XAI as the foundational technology of RAI has not been considered in previous works. This is shown in Table \ref{tab:literatureComparison}, where we compare several recent surveys in the literature. We exclude surveys that review XAI and/or RAI in a domain-specific context (i.e., healthcare or finance).

\begin{table}[h]
\footnotesize
    \caption{Comparison of recent surveys on explainable or responsible AI. Key: \hspace{0.05cm} \cmark\cmark \hspace{0.05cm} The topic is explored in depth. \hspace{0.05cm} \cmark \hspace{0.05cm} The topic is briefly mentioned. \hspace{0.05cm} \xmark \hspace{0.05cm} The topic is not discussed.}    \begin{tabular}    
    {p{2.2cm}p{2.2cm}p{2.2cm}p{2.2cm}p{2.9cm}}
        \hline
        \centering \textbf{Reference} & \centering \textbf{XAI} & \centering \textbf{RAI} & \centering \textbf{Use Cases} & \centering \textbf{XAI as the foundation of RAI} \arraybackslash \\
         \hline
         Barredo Arrieta \textit{et al.} (2020) \cite{BarredoArrieta2020} & \centering \cmark\cmark & \centering \cmark & \centering \xmark & \centering \xmark \arraybackslash \\
         \hline
         Ahmed \textit{et al.} (2022) \cite{Ahmed2022} & \centering \cmark\cmark & \centering \xmark & \centering \cmark & \centering \xmark \arraybackslash \\
         \hline         
         Anagnostou \textit{et al.} (2022) \cite{Anagnostou2022} & \centering \cmark & \centering \cmark\cmark & \centering \cmark & \centering \xmark \arraybackslash \\
         \hline
         Ashok \textit{et al.} (2022) \cite{Ashok2022} & \centering \xmark & \centering \cmark & \centering \xmark & \centering \xmark \arraybackslash \\
         \hline
         Kaur \textit{et al.} (2022) \cite{kaur2022} & \centering \cmark & \centering \cmark\cmark & \centering \xmark & \centering \xmark \arraybackslash \\
         \hline
         Minh \textit{et al.} (2022) \cite{Minh2022} & \centering \cmark\cmark & \centering \xmark & \centering \cmark & \centering \xmark \arraybackslash \\
         \hline
         Dwivedi \textit{et al.} (2023) \cite{dwivedi2023} & \centering \cmark\cmark & \centering \xmark & \centering \xmark & \centering \xmark \arraybackslash \\
         \hline
         Li \textit{et al.} (2023) \cite{Li2023} & \centering \cmark & \centering \cmark\cmark & \centering \xmark & \centering \xmark \arraybackslash \\
         \hline
         Saeed \textit{et al.} (2023) \cite{Saeed2023} & \centering \cmark\cmark & \centering \xmark & \centering \xmark & \centering \xmark \arraybackslash \\
         \hline
         This work & \centering \cmark\cmark & \centering \cmark\cmark & \centering \cmark\cmark & \centering \cmark\cmark \arraybackslash \\ 
         \hline
    \end{tabular}
    \vspace{1em}
    \label{tab:literatureComparison}
\end{table}

Responsible AI has been considered in recent surveys by Kaur \textit{et al.} \cite{kaur2022} and Li \textit{et al.} \cite{Li2023}. In both of these works, the literature on RAI was reviewed and findings were used to propose principals and guidelines for the development and deployment of RAI. In each work, explainability was identified as a characteristic of RAI but was not explored in depth. Characteristics of transparency, accountability, fairness, and robustness were identified by each work.

Another work by Ashok \textit{et al.} \cite{Ashok2022} sought to outline broader ethical principles for digital technology development, including but not limited to AI. They ultimately proposed 14 ethical characteristics, which included accountability, fairness, and privacy. Explainability was not discussed.

RAI was explored through the lens of multiple critical use cases in another recent survey by Anagnostou \textit{et al.} \cite{Anagnostou2022}. Their work first explored the problems in critical sectors including healthcare and transportation, before identifying several characteristics required of RAI. These characteristics were similar to other works and included transparency, accountability, and fairness. Explainability was mentioned as a strategy for supporting transparency, but was not explored in depth.

Dwivedi \textit{et al.} \cite{dwivedi2023} recently conducted a survey that explored explainability techniques and provides guidance on Python programming packages that allow for implementation. Similarly, a recent review by Minh \textit{et al.} \cite{Minh2022} explored XAI techniques in depth, with some discussion of applications. Furthermore, a recent survey by Ahmed \textit{et al.} \cite{Ahmed2022} discussed AI, XAI, and use cases of XAI in Industry 4.0. However, these three works did not discuss responsible or ethical AI concepts.

Another recent survey conducted by Saeed \textit{et al.} \cite{Saeed2023} explored XAI with a particular focus on future challenges. They also briefly discuss how XAI can be used to create trustworthy AI systems, which supports RAI. Perhaps the survey that most considered the relationship between XAI and RAI was conducted by Barredo Arrieta \textit{et al.} \cite{BarredoArrieta2020}. They comprehensively reviewed the literature on XAI, before discussing RAI as a separate but related topic. They briefly highlight how XAI could support RAI in terms of transparency and fairness. However, their review overlooked the usefulness of XAI in areas of RAI including robustness.

Overall, the literature is rich with surveys on RAI and XAI separately. However, there is relatively little discussion in the literature of the strong relationship between these two topics. In this review, we fill that gap in the literature by investigating how XAI can be used to create AI systems that align with existing RAI frameworks and characteristics.

\subsection{Structure of the Paper}
\begin{figure}[h]
    \centering
    \includegraphics[width=0.95\linewidth]{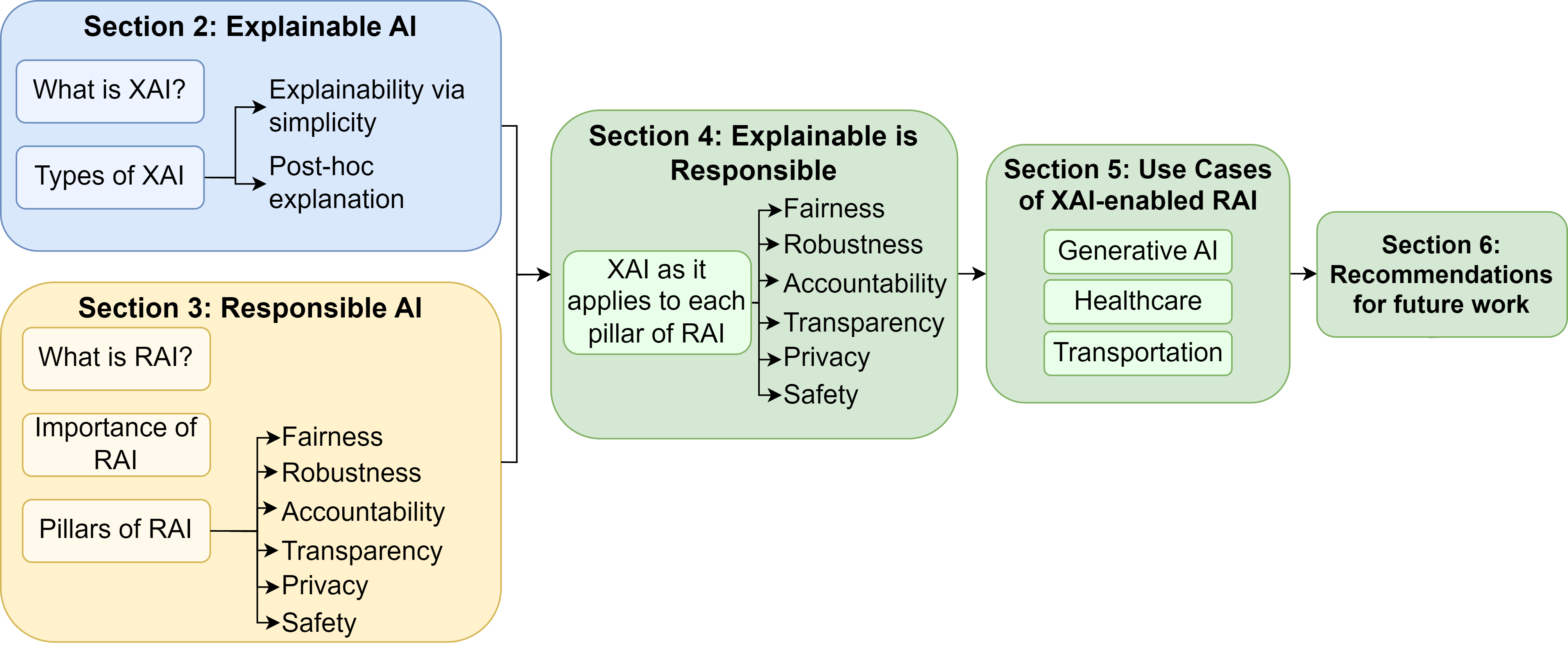}
    \caption{Overview of the key contents of this paper}
    \label{fig:structure}
\end{figure}

The remainder of this paper is structured as follows: Section 2 addresses the question of what makes AI explainable, and explores the types and applications of explainable AI. Section 3 then discusses responsible AI, including why responsibility is important and the characteristics that make AI responsible. Section 4 then examines how the pillars of responsible AI can be achieved through explainability. In Section 5, we present several real-world use cases to highlight the importance of explainability to produce responsible and trustworthy AI for sectors including healthcare and transportation. We present recommendations for future works in the AI domain in Section 6, before concluding the paper in Section 7. The key contents of this paper are illustrated in Fig. \ref{fig:structure}.

\section{Explainable AI}
\subsection{What is Explainable AI?}
Explainable AI (XAI) describes strategies and processes used to make AI models more understandable and interpretable to their developers and end-users, without significant compromise on performance \cite{BarredoArrieta2020, Adadi2018, Ahmed2022}. Audience is critical to consider; strategies utilised to understand a model as the developer may differ from those needed to explain a model to non-technical end users. The key motivations for explainability are improved transparency and trust in AI models, particularly those used in critical applications \cite{Saeed2023}. 

The need for explainable AI is primarily due to the increasing complexity of AI models. Early AI models, such as linear and logistic regression algorithms, can be considered as explainable AI models due to being understandable by design. In such models, the internal mechanisms linking input to output can be readily visualised and interpreted. However, these models are often limited in terms of performance; their simplicity prevents them from accurately interpreting complex data.

As computing resources have improved, increasingly complex AI algorithms have been developed to support better interpretation of complex data. Advanced AI algorithms that continue to grow in prevalence include deep neural networks (DNNs) and random forest. Such models have demonstrated high performance in many domains \cite{Piccialli2021, Himeur2021, Muhammad2021}, however their complexity limits insight into what happens in between input and output; they are not inherently understandable. The increase of high-performing yet opaque models, often termed \textit{black-box} models \cite{Adadi2018}, has lead to the substantially increasing interest in XAI for interpreting complex models.

The following subsections dive deeper into the existing literature on XAI, focusing on both explainability by design, where models are intentionally designed to be inherently interpretable, and post-hoc explainability, whereby the model is retrospectively explained after its development or use. Within both types of explainability, there are methods designed to provide local explanations (i.e., explain a single prediction) or global explanations (i.e., explain the general decision-making process of the model). Additionally, some strategies are model-agnostic (i.e., they can be applied to all or many models), while others are model-specific (they can only be applied to a select model). We introduce these terms here as they are relevant to our discussions in the next two subsections.

\subsection{Explainability by Design}
\label{subsec:xaiByDesign}
Several AI models can be considered explainable by design due to their simplicity; their inner workings are transparent and thus inherently interpretable. Models that fall into this category include linear regression, generalised additive models, logistic regression, decision trees, k-nearest neighbour, and low-dimensionality support vector machines. 

\subsubsection{Linear Regression}
\begin{figure}
    \centering
    \begin{subfigure}[b]{0.32\textwidth}
        \includegraphics[width=\textwidth]{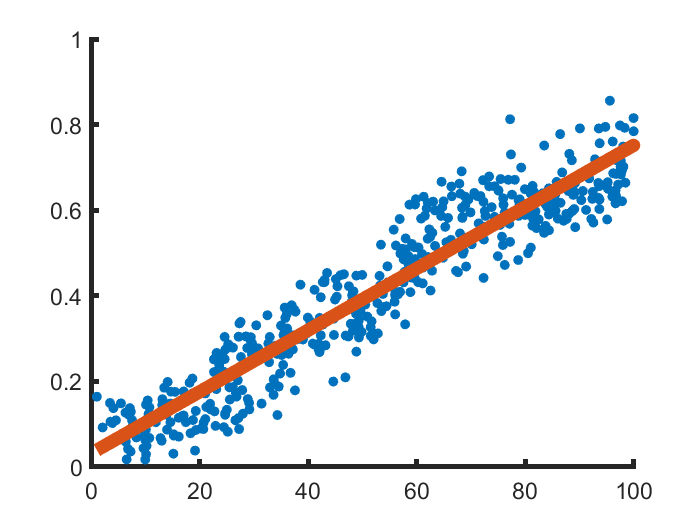}
        \caption{}
        \label{fig:linReg}
    \end{subfigure}
    \begin{subfigure}[b]{0.32\textwidth}
        \includegraphics[width=\textwidth]{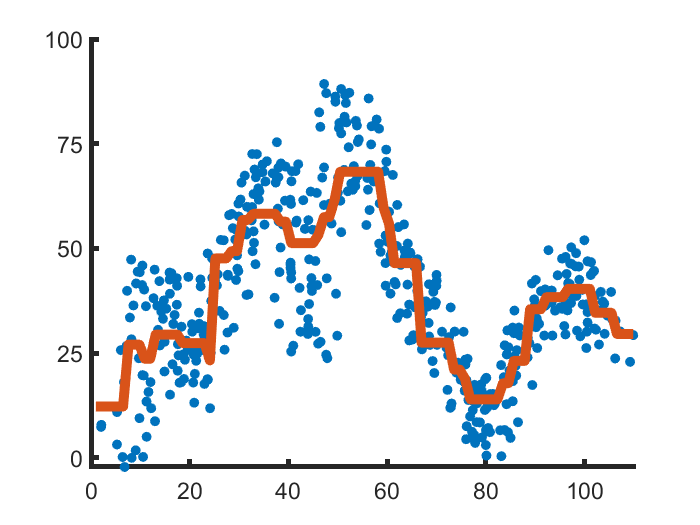}
        \caption{}
        \label{fig:gamReg}
    \end{subfigure}
    \begin{subfigure}[b]{0.32\textwidth}
        \includegraphics[width=\textwidth]{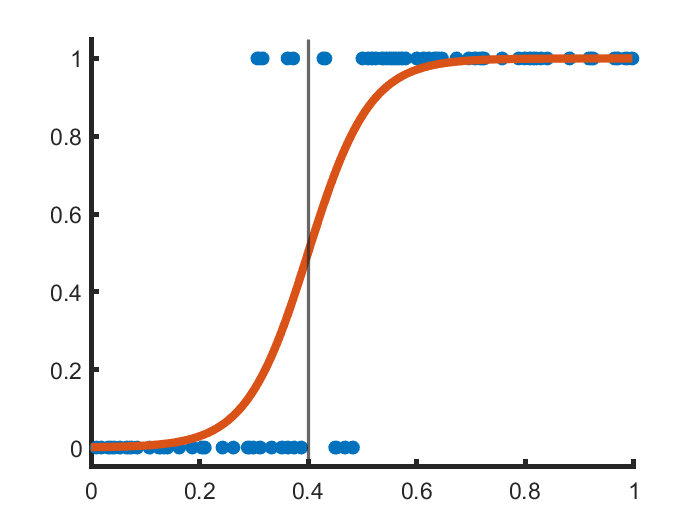}
        \caption{}
        \label{fig:logReg}
    \end{subfigure}
    \vspace{0.5em}
    
    \begin{subfigure}[b]{0.32\textwidth}
        \includegraphics[width=\textwidth]{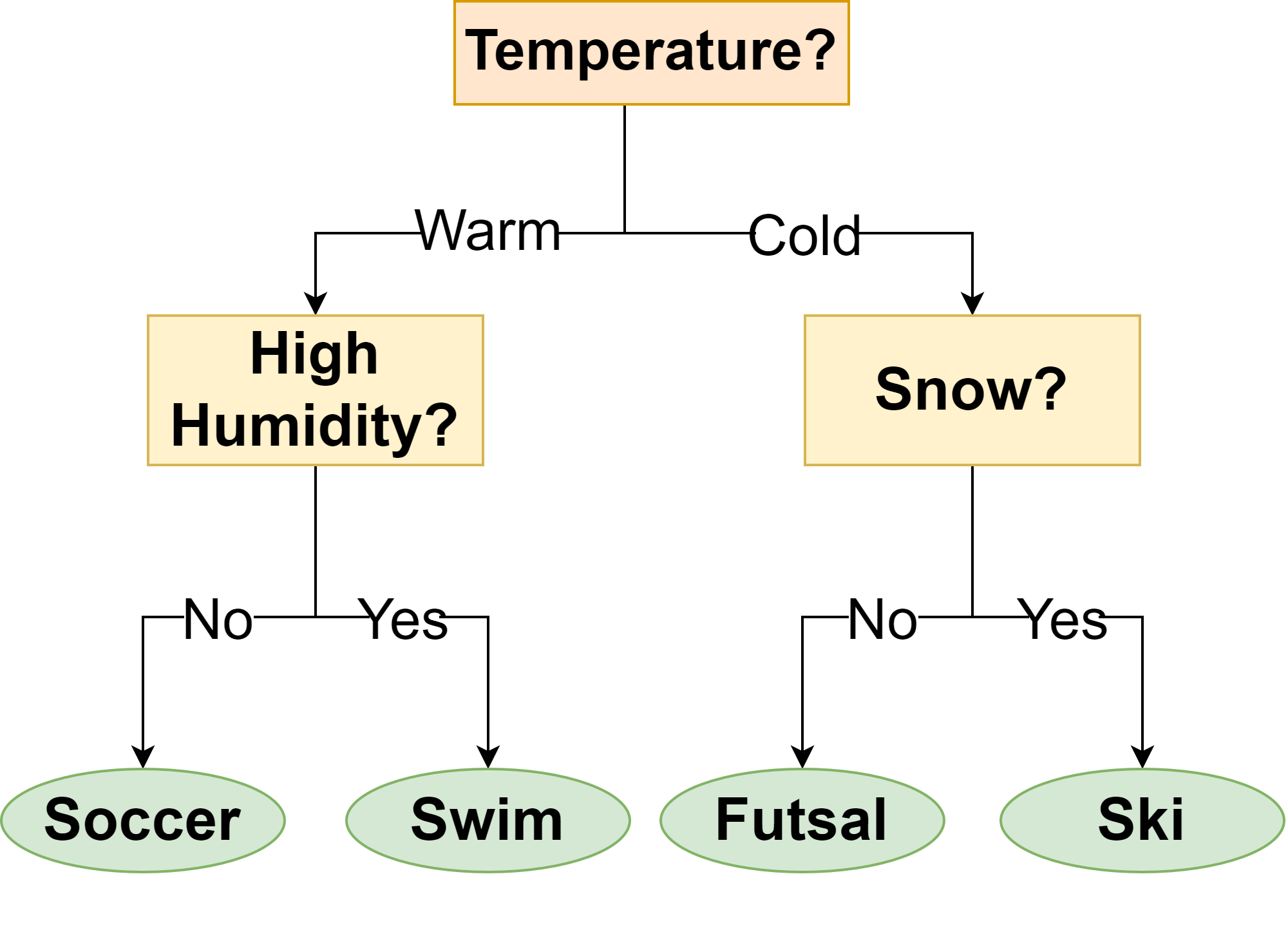}
        \caption{}
        \label{fig:dt}
    \end{subfigure}
    \begin{subfigure}[b]{0.32\textwidth}
        \includegraphics[width=\textwidth]{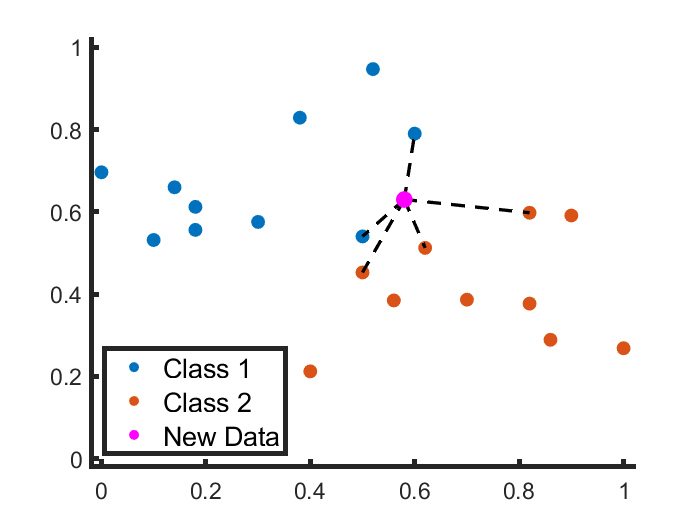}
        \caption{}
        \label{fig:simpleKnn}
    \end{subfigure}
    \begin{subfigure}[b]{0.32\textwidth}
        \includegraphics[width=\textwidth]{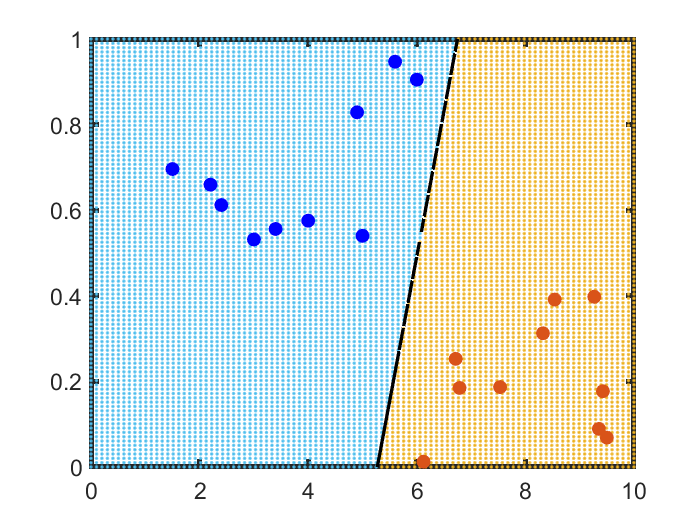}
        \caption{}
        \label{fig:simpleSVM}
    \end{subfigure}
    \caption{Examples of models that are explainable by design, namely (a) linear regression, (b) generalised additive models, (c) logistic regression, (d) decision trees, (e) k-nearest neighbour,  and (f) low-dimensionality SVM.}
\end{figure}

Linear regression (LR) seeks to find a line of best fit to continuous data, as illustrated in Fig. \ref{fig:linReg}. In its simplest form with one independent variable, this type of regression is widely understood and thus inherently explainable. Multivariate LR is more common in the literature as it can consider multiple independent variables, each of which can be explained via individual LR plots. This is still relatively interpretable for lower numbers of variables, however weightings of variables may need to be explained. Despite their advantages in explainability, the performance of LR is limited by its inability to capture non-linear relationships, or complex relationships between variables. Nonetheless, multivariate LR remains in use with recent works utilising it to identify radiomics features associated with clinical outcomes \cite{Granata2022}, water and wastewater forecasting \cite{Jia2022}, and pandemic preparedness assessment \cite{Bollyky2022}.

\subsubsection{Generalised Additive Models}
Generalised additive models (GAMs) overlap with linear regression in that they attempt to fit a line to the data. However, they differ in that they are not constrained to linear features. They are comprised of multiple `splines' - non-linear smooth functions - added together to give an overall line describing the data, as illustrated in Fig. \ref{fig:gamReg}. While there is some complexity in their development, their decisions can be readily explained to the end-user through illustrations of the fitted GAM lines for each variable considered; as with linear regression, interpretabiilty decreases as the number of variables increases. Due to their ability to capture non-linear relationships, GAMs have recently been used for applications including predicting time to crop maturity \cite{Marcillo2021}, electrical load forecasting \cite{Pachauri2022}, and bushfire prediction \cite{Phelps2021}.

\subsubsection{Logistic Regression} Commonly used for classification problems, logistic regression finds a logistic function that best fits to a given set of data. The binary classification case is shown in Fig. \ref{fig:logReg}; the threshold (grey) line splits the outputs into the two classes. This can again be clearly illustrated and explained. Multiple logistic regression can be performed where multiple variables exist, remaining explainable for lower numbers of variables. Multiclass logistic regression is more complex, however recent papers have attempted to improve explainability through weight heat-maps \cite{Bennetot2022}. Logistic regression has been relatively popular in XAI literature, used for applications including image recognition \cite{Bennetot2022}, telehealth record sentiment analysis \cite{Obeid2020}, adverse reaction to medication prediction \cite{Zhan2020}, and predicting student outcomes \cite{Jang2022, Kostopoulos2021}. However, logistic regression is often outperformed by more advanced models \cite{Aslam2022, Anand2021}.

\subsubsection{Decision Trees}
Decision trees (DTs) utilise multiple input features to make a series of decisions resulting in an output. The features are typically `branching' points, with the tree splitting into separate paths after each feature is considered. The is illustrated by Fig. \ref{fig:dt}, which shows a simple DT for choosing a sport to play based on weather conditions. DTs are explainable due to their clear logical flow, resembling flowcharts that are common in many industries. However, they are limited in their implementation and have become less popular in the literature. DT ensemble methods such as random forest (RF) are more common in the literature, however explainability decreases as the number of trees increases; many RF papers explain their models using post-hoc methods \cite{Peng2021, Liang2022, Wang2021} discussed in the next section.

\subsubsection{k-Nearest Neighbour}
In $k$-nearest neighbour (KNN) models, a prediction is made by identifying the \textit{k} closest datapoints to the new datapoint of interest, and using their values to determine what output value should be assigned to a new input. In the classification example illustrated in Fig. \ref{fig:simpleKnn}, this would result in the new data being classified as Class 2. KNN can also be used for regression tasks. The premise of using proximate data to make a decision is intuitive, and can be easily visualised where \textit{k} is small. However, as the value of \textit{k} increases, the interpretability of the model decreases. The simplicity of KNN also leads to limitations, including susceptibility to unimportant features and ignoring the distances of each neighbour \cite{Uddin2022}. Classic KNN is primarily seen in the literature for benchmarking or validating feature selection algorithms \cite{Tubishat2021, Song2021}, however KNN variants have been recently used in applications including severity assessment for Parkinson's disease \cite{Zhao2022} and prediction of cardiovascular disease \cite{Ghosh2021}.

\subsubsection{Support Vector Machines}
Support vector machines (SVMs) are generally not considered to be explainable, as they are challenging to visualise or explain for higher dimensionality inputs. However, we argue that this is also true of algorithms such as linear regression. Low-dimensionality SVMs split data into classes based on a line or hyperplane. For simple binary classification scenarios with up to three features (i.e., three dimensions), the clusters and separating line or plane can be readily illustrated to the end user; an example for 2D binary classification is given in Fig. \ref{fig:simpleSVM}. Due to the interpretability of low-dimensionality SVM, some early studies have sought to represent complex data in lower-dimensional feature spaces so that explainable 3D SVM can be utilised \cite{Zhang2022}.

\subsection{Post-Hoc Explainability}
\label{subsec:postHocXai}

\begin{figure}
    \centering
    \begin{subfigure}[b]{0.28\textwidth}
        \includegraphics[width=\textwidth]{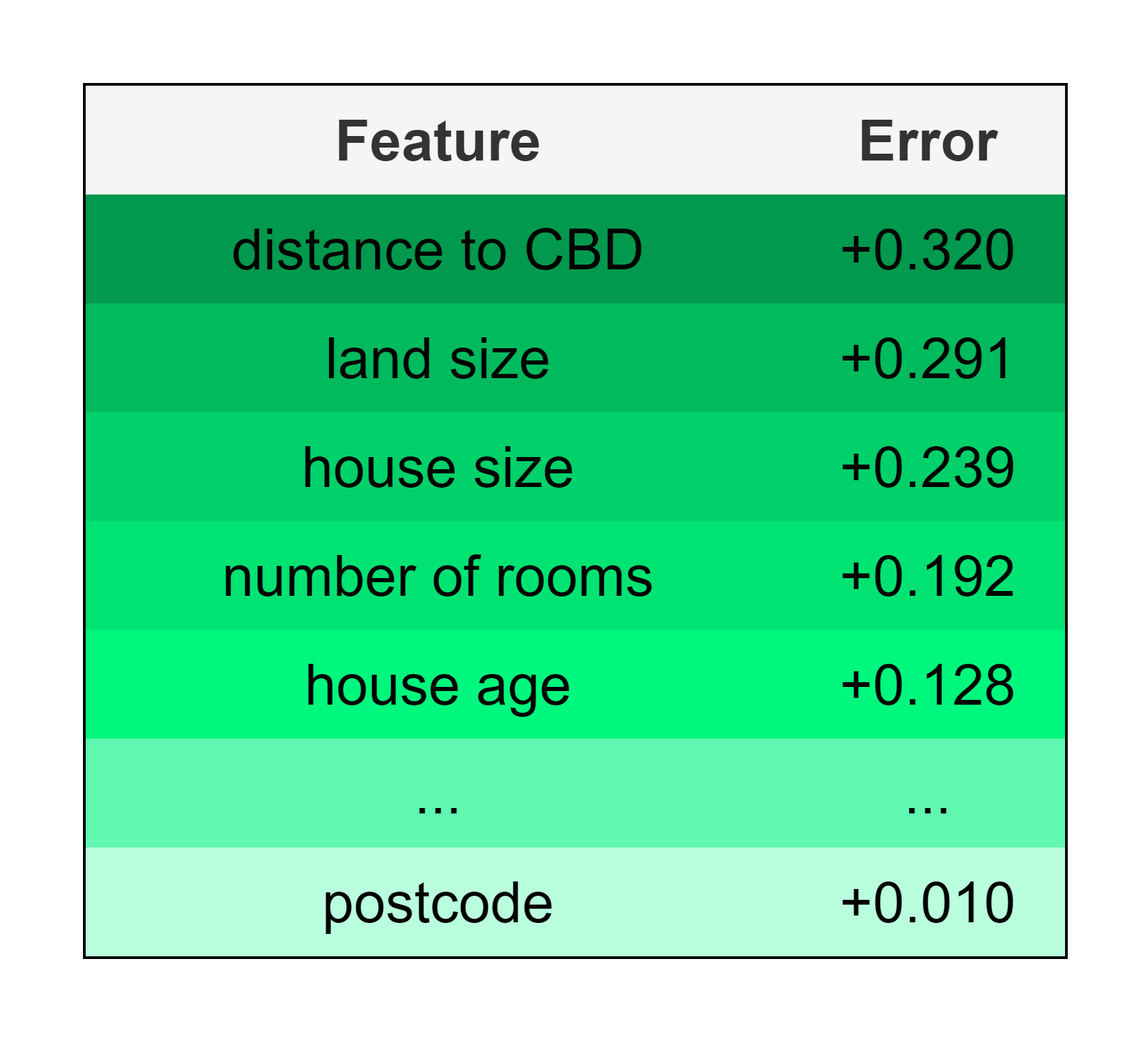}
        \caption{}
        \label{fig:permutation}
    \end{subfigure}
    \begin{subfigure}[b]{0.32\textwidth}
        \includegraphics[width=\textwidth]{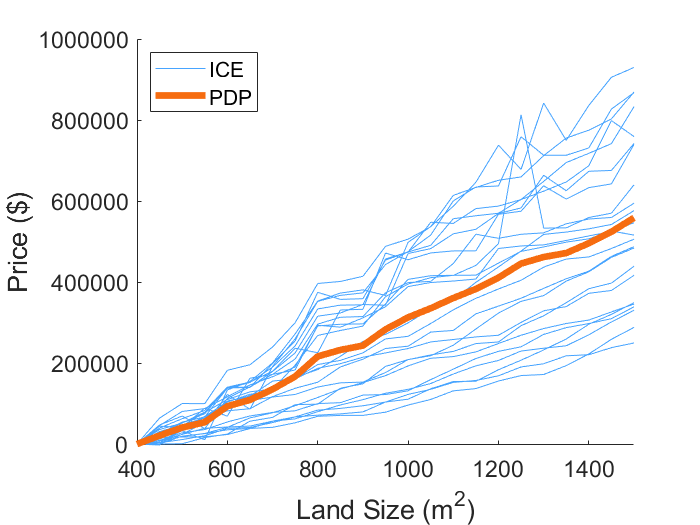}
        \caption{}
        \label{fig:iceAndPdp}
    \end{subfigure}
    \begin{subfigure}[b]{0.32\textwidth}
        \includegraphics[width=\textwidth]{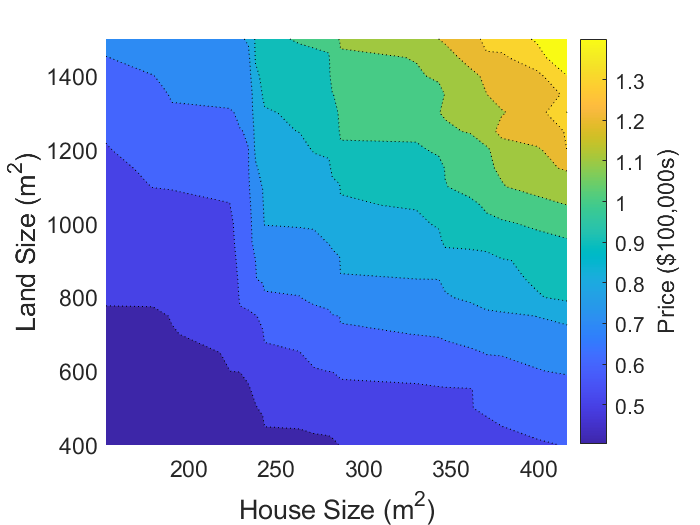}
        \caption{}
        \label{fig:2dpdp}
    \end{subfigure}
    
    \begin{subfigure}[b]{0.4\textwidth}
        \includegraphics[width=\textwidth]{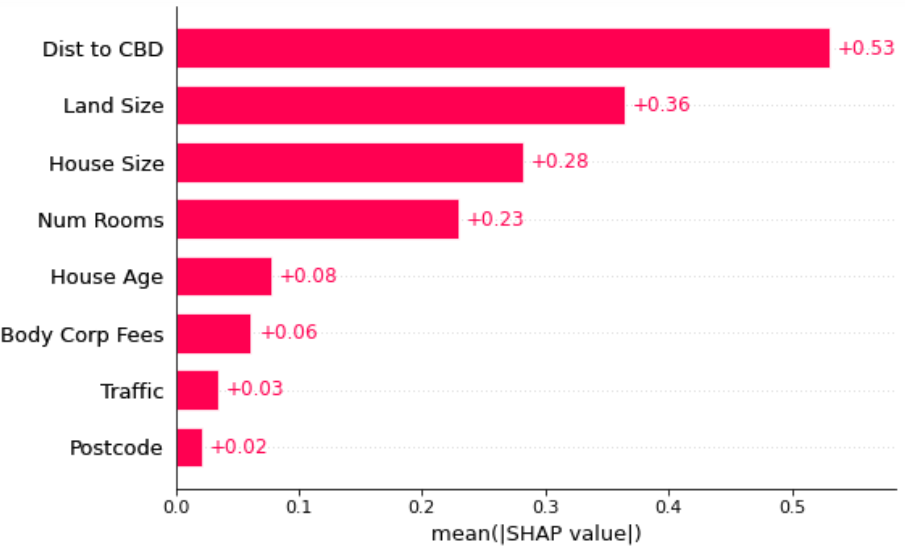}
        \caption{}
        \label{fig:shapBar}
    \end{subfigure}
        \begin{subfigure}[b]{0.45\textwidth}
        \includegraphics[width=\textwidth]{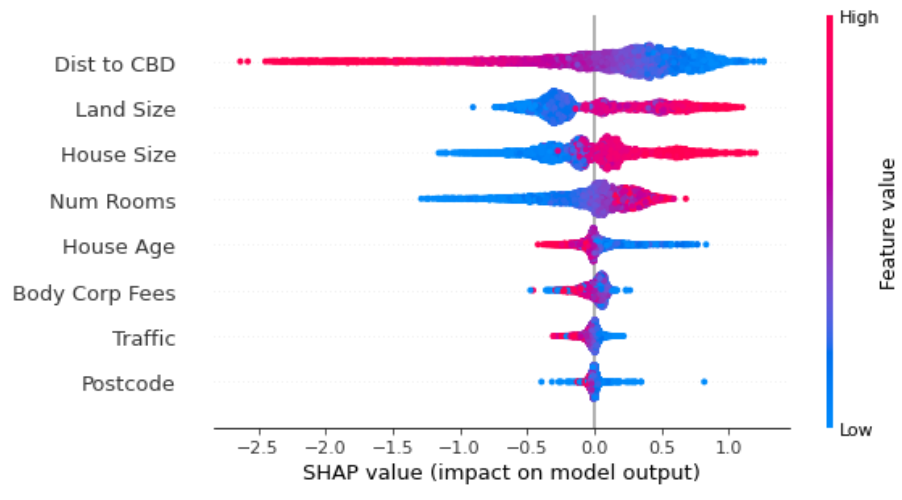}
        \caption{}
        \label{fig:shapBee}
    \end{subfigure}
    \caption{Global feature importance approaches to post-hoc explainability, namely (a) permutation importance, (b) single-feature ICE and PDP, (c) two-feature PDP, (d) SHAP bar plot, and (e) SHAP beeswarm plot.}
\end{figure}

\subsubsection{Permutation Importance}
Permutation importance is a model-agnostic explainability technique that seeks to quantify feature importances in a global context, applied after a model is trained. It involves randomly shuffling all values for a particular feature and assessing whether this has an impact on model performance. The output of permutation importance is a list of weights indicating how important each feature is. These weights represent the impact of shuffling the particular feature on a metric of interest. An example output for a house pricing task is shown in Fig. \ref{fig:permutation}, with features that caused the highest increase in prediction error considered to be the most important. Permutation importance is a computationally expensive technique, however both the output and the method itself are easily understood. This has lead to it being widely used in applications including disease biomarker discovery \cite{Mi2021}, digital soil mapping \cite{Taghizadeh-Mehrjardi2021}, wildfire susceptibility analysis \cite{Zhang2021}, and prediction of biohydrogen production efficiency \cite{Hosseinzadeh2022}. A key limitation is that its results can be misleading where two variables are highly correlated; shuffling one may have little effect on the output due to the presence of the other.

\subsubsection{Partial Dependence Plots and Individual Conditional Expectation}
Partial dependence plots (PDPs) and individual conditional expectation (ICE) plots are model-agnostic methods which seek to understand the changes in prediction if a single feature is modified in the global context. After a model is trained, the particular feature is modified to assess the impact on the output. ICE plots illustrate the impact of the feature for every instance in the database; e.g., every instance is plotted as a single line. PDP plots find the average of the impact across all features, and therefore result in a single line. An example for the impact of land size on house pricing is illustrated in Fig. \ref{fig:iceAndPdp}.

ICE and PDP each have advantages and disadvantages. ICE can reveal heterogeneous effects, such as when changing a feature has a negative effect on half of the instances, but a positive effect in the other half. In such a case, PDP would display a flat, horizontal line - a result that is misleading and not useful. However, the interpretability of ICE decreases as the number of instances increases; plots become crowded, and it becomes harder to separate instances or identify atypical instances. PDP avoids this problem by illustrating an average. 

PDP is also useful in in graphing the dependencies of two features. As shown in Fig. \ref{fig:2dpdp}, PDP can be used to generate contour plots that illustrate the relationship of two features to the output of interest - in our example, the relationship of house and land size to house price. This cannot be achieved with ICE, as the large number of overlapping contours would not be interpretable.

As they each have their respective benefits, ICE and PDP are commonly applied together; they have seen use in applications including acute kidney injury risk assessment \cite{Kim2021}, identification of factors in crash severity \cite{Afshar2022}, water stream degradation assessment \cite{Maloney2022}, and mortality risk prediction in intensive care settings \cite{Safaei2022}. PDP has also been commonly applied alone in the literature. Recent works have applied PDP to applications including biochar yield prediction \cite{Khan2022}, concrete strength prediction \cite{QuanTran2022}, crop yield assessment \cite{Zhi2022}, and deterioration in hepatitis patients \cite{Peng2021}.

\subsubsection{Shapley Additive Explanations}
Shapley Additive Explanations (SHAP) \cite{shap} is a model-agnostic explainability approach that uses a game theory approach to assess the impact on model performance when different combinations of features (`players') participate in the prediction (`game'). It assesses all combinations of features and uses an additive approach to summarise the impact of each feature. 

SHAP can be utilised in several different ways. Examples in the global context are illustrated by Figs. \ref{fig:shapBar} and \ref{fig:shapBee}. In Fig. \ref{fig:shapBar}, feature importances as determined by SHAP are simply graphed as bar charts. Meanwhile, Fig. \ref{fig:shapBee} shows a `beeswarm' chart. Features are listed in order of determined importance. Each dot represents a single instance, and its colour indicates whether the feature had a low (blue) through to high (red) value; for example, large distances to the CBD are shown to have a negative impact on the output of house prices.

\begin{figure}[ht]
    \centering
    \begin{subfigure}[b]{0.98\textwidth}
        \includegraphics[width=\textwidth]{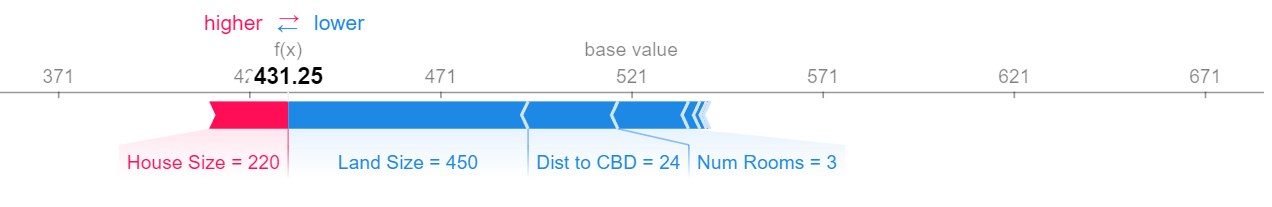}
        \includegraphics[width=\textwidth]{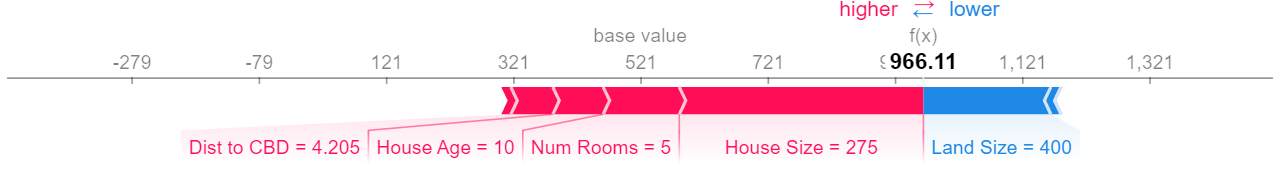}
        \caption{}
        \label{fig:shapForce}
    \end{subfigure}
    
    \begin{subfigure}[b]{0.98\textwidth}
        \includegraphics[width=\textwidth]{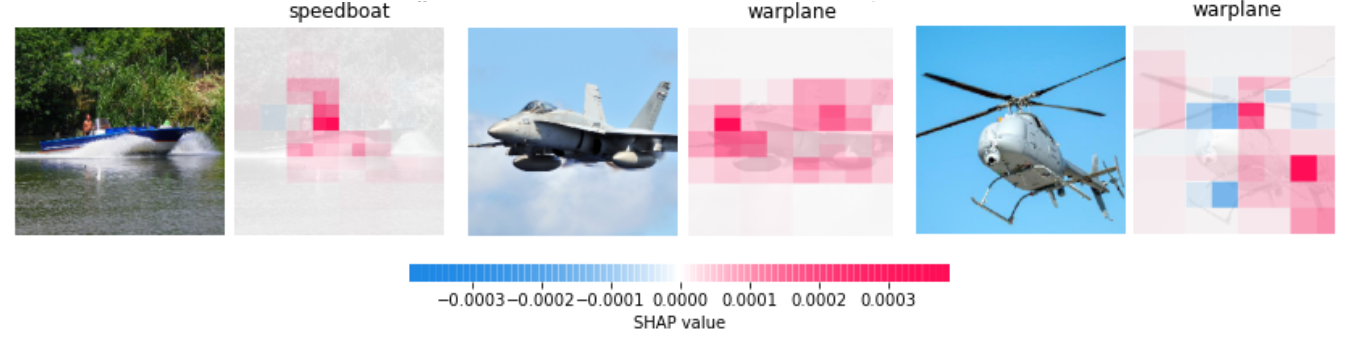}
        \caption{}
        \label{fig:shapImage}
    \end{subfigure}
    
    \caption{Local usage of SHAP explanations for (a) data with discrete features, and (b) image-based data.}
\end{figure}

SHAP can also be used in a local context. For an individual prediction using feature-based data, force plots can be generated to explain which features contributes the final output, including the direction in which they contributed - as shown in Fig. \ref{fig:shapForce}. For image-based data, heatmaps can be generated to illustrate the regions of an image that SHAP considers most important when making a classification. Examples of these are seen in Figs. \ref{fig:shapImage}; these examples were generated using the pre-trained ResNet model \cite{Keras2023} for ImageNet data \cite{StanfordVisionLab2020}. The helicopter was incorrectly labelled as a warplane, but the presence of blue squares indicates a lower level of confidence. This transparency can help maintain trust even where predictions are wrong.

SHAP has been extremely popular in recent literature due to the range of clearly interpretable graphical methods that can be implemented to understand its results. SHAP has been used to explain feature importances in applications including mortality risk assessment \cite{Baker2021}, concrete creep prediction \cite{Liang2022}, plastic waste forecasting \cite{Fan2022}, wastewater processing analytics \cite{Wang2022}, electric vehicle route planning \cite{Ullah2022}, and fault detection for rotating machinery \cite{Brito2022}. SHAP has also been applied for image explanations in a wide range of settings, including brain tumour diagnosis \cite{Gaur2022}, offshore oil slick identification \cite{Amri2022}, and spam image detection \cite{Zhang2022Spam}.


\subsubsection{Local Interpretable Model-Agnostic Explanations}
Local Interpretable Model-Agnostic Explanations (LIME) \cite{lime} is an explainability tool that does what its name suggests - offers model-agnostic explanations in a local context. It modifies a single data sample by altering features, and then fits a linear model to the pertubed dataset in the vicinity of the original data sample. This is then used to generate explanations that are accurate for the local context, but invalid in the global context.

\begin{figure}[ht]
    \centering
    \begin{subfigure}[b]{0.9\textwidth}
    \includegraphics[width=\textwidth]{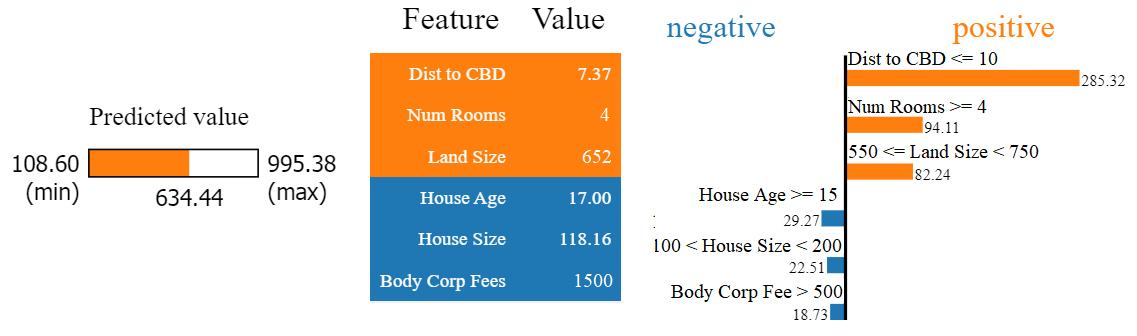}
    \caption{}
    \label{fig:limeFeature}
    \end{subfigure}
    
    \begin{subfigure}[b]{0.98\textwidth}
    \includegraphics[width=\textwidth]{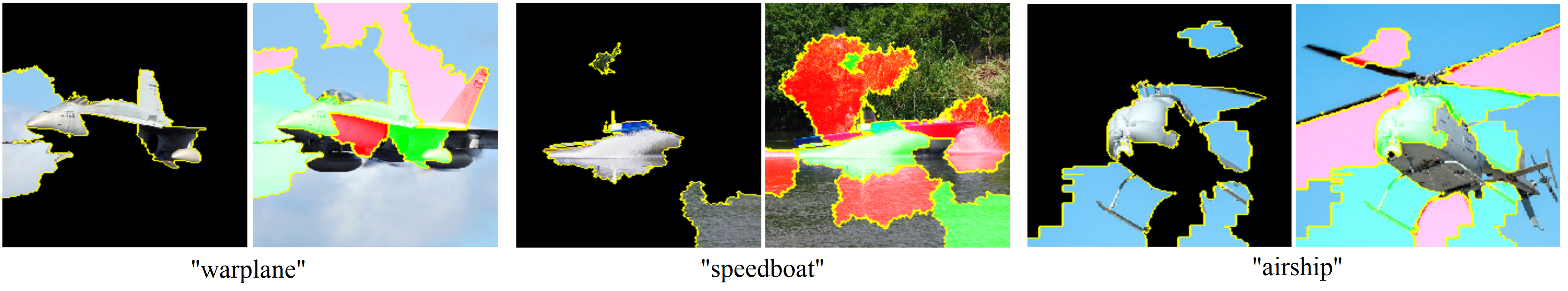}
    \caption{}
    \label{fig:limeImage}
    \end{subfigure}
    \caption{Local explanations with LIME for (a) feature-based data, and (b) image-based data.}
\end{figure}

LIME returns scores for feature importances in the local context, and thus many graphical methods for interpretation resemble those seen with previous feature importance methods. Fig. \ref{fig:limeFeature} illustrates LIME's graphical outputs when assessing a single prediction; it displays the overall prediction, a list of the most important features, and a `waterfall' graph of the most important features. Features on the right pushed the predicted house price higher, while those on the left pushed the prediction lower. LIME can also be used on image and text data, highlighting regions of an image or words within a document that contributes to an overall output prediction, as shown in Fig. \ref{fig:limeImage}. For each example in Fig. \ref{fig:limeImage}, the visible section in the left image indicates the areas that contributed towards the prediction. The green and red areas highlighted in the right image are those that had positive and negative impacts on the prediction, respectively.

Alongside SHAP, LIME is one of the most popular explainability methods in the literature. LIME has been used on feature-based data for applications including activity recognition with wearable sensors \cite{Uddin2021}, steel beam damage prediction \cite{Onchis2021}, bankruptcy prediction \cite{Park2021}, and stock market forecasting \cite{Wu2022}. It has also been broadly used for image and waveform classification tasks including heartbeat detection from electrocardiogram \cite{Neves2021}, COVID-19 detection in chest x-rays \cite{Ahsan2021}, and quality assessment for augmented synthetic aperture radar images \cite{Zhu2022}. Finally, LIME has been used for text analysis tasks including depression identification from symptom self-reporting \cite{Uddin2022depression} and identification of resource needs following natural disaster based on Twitter posts \cite{Behl2021}.

\subsubsection{Class Activation Mapping}
\begin{figure}[ht]
    \centering
    \includegraphics[width=0.65\textwidth]{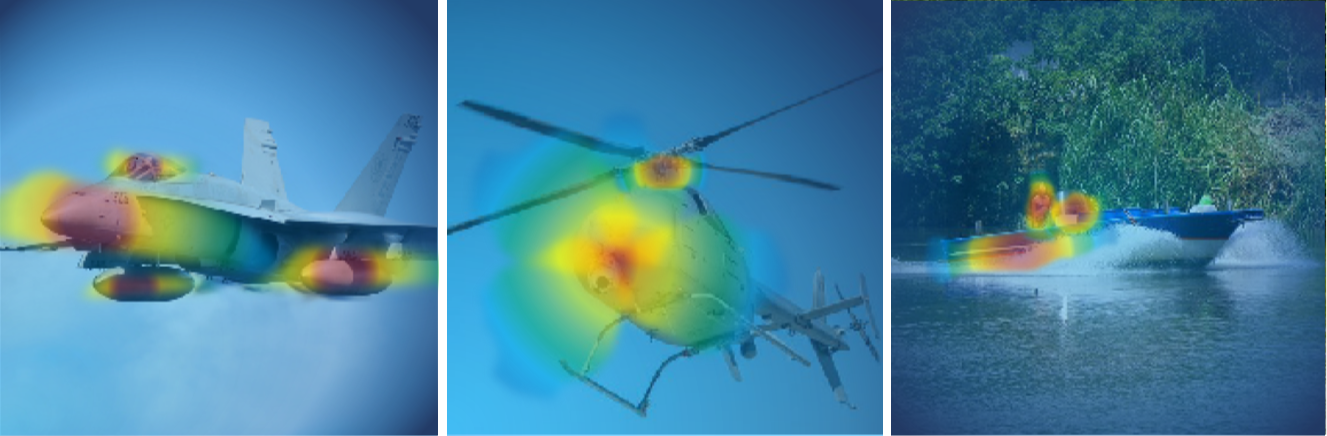}
    \caption{Example of outputs generated by CAM, Grad-CAM, and other tools utilising a heatmapping approach.}
    \label{fig:camFig}
\end{figure}

Class activation mapping (CAM) is a model-specific method for providing local explanations of image-based predictions made by CNNs. Classical CAM by inserting a global average pooling (GAP) layer after the final convolutional layer of a convolutional neural network (CNN), and weighting the GAP outputs to generate a heatmap. An example of outputs is illustrated in Fig. \ref{fig:camFig}, with the red `hotter' areas contributing the most to a prediction, through to blue `cooler' areas which contribute the least. 

Classical CAM has also inspired a large number of variants in recent years, including Score-CAM \cite{Wang2020}, Gradient-weighted CAM (Grad-CAM) \cite{Selvaraju2017}, and HiResCAM \cite{Draelos2020}. One recent paper proposed a CAM method using SHAP scores \cite{Zheng2022}. The key purpose of each variant has been to produce visualisations that more faithfully illustrate a model's decision making process. However, it is challenging to prove superiority of one method against another without extensive manual comparison of explanations by end-users; this has not been broadly conducted in the literature.

Variants of CAM have been utilised in a wide range of image classification papers, with applications including tree mapping for forest management \cite{Onishi2021}, Alzheimer's disease diagnosis from structural magnetic resonance imagery \cite{Zhang2022cam}, manufacturing cost prediction from computer-aided designs \cite{Yoo2021}, and guided rehabilitation training \cite{Qiu2022}.

\subsubsection{Other Techniques}
In the above subsections, we have examined some of the most prevalent XAI techniques in the literature. However, there are many additional techniques that are gaining interest in the literature. One example is counterfactuals, a group of model-agnostic local explanation techniques that seek to explain what would need to be different for the outcome to have changed; these are particularly useful for binary classification tasks \cite{Guidotti2022}.

Anchors focus on building rule-based model simplifications that capture the key features and identify \textit{if-then} rules that explain predictions \cite{Ribeiro2018}. Rules are developed in the global context, and can then be used to explain local predictions. Anchors have also inspired related approaches such as the recently proposed CASTLE (cluster-aided space transformation for local explanations), which has been shown to outperform its predecessor on a number of databases \cite{LaGatta2021}.

Emerging AI systems such as generative AI have also posed challenges for explainability, as the underlying models have high complexity. In the natural language processing (NLP) context, one recent work adapted SHAP scores for the language context, highlighting text that indicated COVID-19 misinformation \cite{Ayoub2021}. Another novel method determined layer-wise relevance propagation (LRP) for each component of a transformer model, using this to generate heatmaps with higher accuracy than CAM approaches \cite{Chefer2021}. This approach was also demonstrated to work in the NLP context \cite{Chefer2021}, and has been applied to problems such as COVID-19 screening using chest radiography \cite{Mondal2022}. 

\section{Responsible AI}
\subsection{What is Responsible AI?}
Responsible AI (RAI) describes the principles by which AI systems are developed, deployed, and utilised in order to comply with legal and ethical standards. As laws and ethics vary from region to region, there is no singular definition of what makes AI responsible; however, many stakeholders have sought to define frameworks to guide responsible AI development \cite{kaur2022, Anagnostou2022, Li2023}.

In much recent literature, RAI has been used synonymously with related terms such as trustworthy AI \cite{kaur2022}, ethical AI \cite{Ayling2022}, and fair or equitable AI \cite{Robert2020, Ti2021}. However, we suggest that RAI encompasses these concepts; that truly responsible AI must be trustworthy, fair, and adhere to ethical norms, amongst other characteristics. Based on our review of the literature on RAI and related terms, this section illustrates the importance of responsibility in AI and defines six essential pillars of responsibility.

\subsection{Importance of Responsibility}
AI systems are becoming increasingly prevalent in everyday life, used in applications ranging from media recommendation systems to self-driving vehicles. The dangers of irresponsibly developed AI in safety-critical applications are evident: if autonomous vehicles, weapon systems, or automated medication delivery devices fail to operate as intended, then lives can be lost or irreparably altered. It is essential that AI systems making safety-critical decisions are designed responsibly to ensure that they are reliable, accountable, and free from biases. 

In other domains, the dangers of irresponsible AI design are less obvious but can be equally serious. One notable example is facial recognition, which is used by law enforcement \cite{NewSouthWalesPoliceForce2023}, retail stores \cite{CHOICE2022}, and sporting venues \cite{Bavas2019}. In each of these settings, facial recognition technology is primarily being used to identify perpetrators of crime or misdemeanours. However, researchers have found that commercial facial recognition tools have higher error rates when identifying women and racial or ethnic minorities compared to light-skinned men \cite{buolamwini2018, raji2022}. Thus, innocent civilians belonging to marginalized groups are more likely to be wrongly criminalized due to mistaken identification. Additionally, predictive policing algorithms trained on images of previously arrested persons have been shown to exhibit racial biases \cite{Bowyer2020}. It is widely theorised that biases in facial recognition algorithms are largely due to data imbalance \cite{Bowyer2020, buolamwini2018}; minority groups are often underrepresented in data used to train commercial facial recognition models \cite{buolamwini2018}, and overrepresented in data used to train predictive policing models \cite{Bowyer2020}. Thus, irresponsible AI can reproduce and reinforce human biases in justice systems, contributing to a cycle of discriminatory arrest practices \cite{Malek2022}.

Facial recognition is just one example of where AI systems can have life-altering consequences. Medical AI diagnosis systems can fail to accurately diagnose conditions in gender, racial, and ethnic minorities, largely due to these groups being underrepresented in medicine and medical data until the last century \cite{Straw2020}. Hiring algorithms have also been shown to exhibit biases based on gender, ethnicity, age, disability, and sexuality due to human biases \cite{Yam2021, Tilmes2022, Dastin2019}. Even where these characteristics are removed from a resume, NLP approaches can still discriminate against minority groups based on sociolinguistic patterns \cite{Deshpande2020}. Automated lending algorithms are also known to exhibit biases in terms of approval rates and maximum loan sizes for minority groups \cite{Zou2022}.

Aside from the potentially life-altering consequences of biased AI, there are also risks of physical, mental and social harm associated directly with some types of AI. Generative AI systems are increasingly popular in a range of fields, ranging from education to psychology. However, there are many recent and historical examples of generative AIs producing biased musings and hate speech \cite{Alba2022, Wakefield2016}, perpetuating misinformation \cite{Hsu2022}, or responding inappropriately to mental health crises \cite{Martinengo2022, Laestadius2022}. There is also some evidence that human-chatbot relationships can become dysfunctional and lead to social or mental harm \cite{Laestadius2022}. Even state-of-the-art models such as ChatGPT are already being shown to have dangers; one recent work demonstrated that ChatGPT provided misleading information about mental health disorders and treatments \cite{VanDis2023}, while another recent work suggests that ChatGPT provides ``correct but inadequate'' information about medical conditions \cite{Yeo2023}.

Even seemingly innocuous recommendation systems can have dangers. Some studies have found that recommendation systems on some media platforms can lead to users being guided towards problematic content based on their previous media engagement \cite{Alam2021, Yesilada2022Systematic}. Amplification of misinformation and extremism has been found to be particularly prevalent on YouTube \cite{Yesilada2022Systematic}, creating `filter bubbles' where users are predominantly recommended content that confirms their existing views. Research has found that the filter bubble can be easily burst through providing more balanced recommendations, however this is not being implemented by all platforms \cite{Tomlein2021}.

Overall, there are many ways that AI can cause harm - socially, financially, mentally, and physically. The potential of AI to cause harm is the key motivation for RAI. Each of the substantial risks outlined in this section can be mitigated through strategies to open the black box of AI in order to better develop, deploy and use AI responsibly - for the good of society and environments. 

\subsection{Pillars of Responsibility}
In this section, we propose a framework comprised of six critical characteristics for responsible AI: fairness, robustness, transparency, accountability, privacy, and safety. Each of these pillars has been identified as critical to RAI based on our review of RAI policies and frameworks recently proposed in academia, industry, and governance, as highlighted in the following subsections. The identification of these key pillars then guides our discussions in Section 4, where we examine how XAI is foundational to all pillars of RAI.

To overview the prevalence of each pillar in the literature, Table \ref{tab:RAIpillars} summarises the frequency with which various sources include a particular pillar in their own responsible AI framework or guidelines. In terms of academic sources, we focus on four recent frameworks published in reputable journals \cite{Ashok2022, kaur2022, Li2023, Anagnostou2022}. In terms of government and political sources, we have considered AI frameworks and guidelines developed by government and political bodies in eight diverse regions - namely Australia's Commonwealth Scientific and Industrial Research Organisation (CSIRO) \cite{CSIRO2023}, the United States' (US) National Institute of Standards and Technology (NIST) \cite{USGovRAI2022}, China's Ministry of Science and Technology (MOST) \cite{ChinaRAI2022}, India's INDIAai \cite{IndiaAI2022}, Japan's Ministry of Economy Trade and Industry (METI) \cite{JapanMIST2022}, Saudi Arabia's Saudi Authority for Data and Artificial Intelligence (SADAI) \cite{SaudiArabia2022}, the United Kingdom's (UK) Office for Artificial Intelligence \cite{UKGovernment2021}, and the European Union's (EU) High-Level Expert Group on Artificial Intelligence (AI HLEG) \cite{EuropeanCommission2021}. Lastly, we considered RAI principles from six industry sources that are prevalent in the AI space, namely Google \cite{Google2023}, Microsoft \cite{Microsoft2023}, Meta \cite{MetaAI2021}, Samsung \cite{Samsung2023}, IBM \cite{IBM2023}, and Amazon \cite{AmazonWebServices2022}.

\begin{table}[]
    \centering
    \caption{Quantifying the number of considered sources that included each pillar in their respective RAI frameworks or guidelines. A total of 4 academic sources, 8 government and political sources, and 6 industry sources were consulted. Academic sources were papers recently published in reputable journals, government and political sources were selected to be representative of a wide range of global regions, and industry sources were selected based on their prevalence in the AI domain.}
    \footnotesize
    \begin{tabular}{|p{1.3cm}|p{1.65cm}|p{1.65cm}|p{1.65cm}|p{1.65cm}|p{1.65cm}|p{1.65cm}|}
    \hline
    \centering{\textbf{Source}} & \multicolumn{6}{c|}{\textbf{Pillar}} \\
    \cline{2-7} 
    \centering{\textbf{Type}} & \centering{Fairness} & \centering{Robustness} & \centering{Transparency} & \centering{Accountability} & \centering{Privacy} & \centering{Safety} \arraybackslash \\
    \hline
    Academia & \centering{4} & \centering{2} & \centering{4} & \centering{4} & \centering{4} & \centering{4} \arraybackslash \\ 
    \hline
    Government & \centering{8} & \centering{8} & \centering{8} & \centering{8} & \centering{8} & \centering{8} \arraybackslash \\ 
    \hline
    Industry & \centering{6} & \centering{4} & \centering{5} & \centering{5} & \centering{6} & \centering{4} \arraybackslash \\
    \hline
    \end{tabular}
    \vspace{0.5em}
    \label{tab:RAIpillars}
\end{table}

\subsubsection{Fairness}
Fairness of an AI system ensures that the model treats people and scenarios equitably, without discrimination. Several recent works on RAI have identified fairness as a key attribute of responsibility \cite{Ashok2022, kaur2022, Li2023, Anagnostou2022}. Another work by Mehrabi \textit{et al.} \cite{Mehrabi2021} conducted a focused survey on bias and fairness in AI, arguing that fairness is the most critical issue facing successful development and deployment of AI systems. 

Four of the considered industry sources specifically named fairness as a fundamental principle of RAI \cite{MetaAI2021, Microsoft2023, AmazonWebServices2022, IBM2023}. Of the two that didn't explicitly name fairness, the concept is still present - Google's related principle is to ``avoid creating or reinforcing unfair bias'' \cite{Google2023}, while Samsung discusses fairness under their principle of ``Diversity \& Inclusion''.

Government sources from around the world unanimously agree that fairness and non-discrimination is compulsory for AI systems \cite{USGovRAI2022, CSIRO2023, UKGovernment2021, SaudiArabia2022, ChinaRAI2022, JapanMIST2022, EuropeanCommission2021, IndiaAI2022}. However, it is worth noting that not all sources define what fairness means in their regional context.

Overall, the principle of fairness is perhaps the most agreed upon of all the pillars of RAI. Fairness is explicitly mentioned by all consulted sources across academia, governing bodies, and industry.

\subsubsection{Robustness}
Robustness of an AI system ensures that the model is accurate across all potential use cases, and resilient against malicious attacks. This concept is perhaps the most familiar to AI developers and users; there has long been significant value placed on ensuring that AI systems are robust, reliable, and resilient.

The pillar of robustness has been identified by most academic sources. One work directly names robustness as a key principle of RAI, focusing on robustness against adversarial attacks \cite{Li2023}. Meanwhile, another work identifies the related concept of ``technical robustness and safety'' as critical; they suggest that a system meets this criteria if it performs as expected, is resilient against attacks, and can recover from failure without causing harm \cite{kaur2022}. Additionally, the related concept of reliability is mentioned in one work as a means of ensuring that AI is understandable \cite{Ashok2022}.

In industry, robustness is widely considered but terminology varies. Meta \cite{MetaAI2021}, IBM \cite{IBM2023}, and Amazon \cite{AmazonWebServices2022} all name robustness as a key principal of RAI, while Microsoft uses the terminology of reliability \cite{Microsoft2023}. Google \cite{Google2023} do not directly discuss reliability or responsibility, but under the principle of ensuring safety they mention the need to ``avoid unintended results that create risks of harm''. This is somewhat related to robustness, but does not cover all elements of the concept. Interestingly, Samsung \cite{Samsung2023} do not discuss robustness, reliability, or accuracy in their RAI framework.

Most government and political bodies note the need for robustness in RAI systems. Robustness is noted as key for responsibility by government bodies in the US \cite{USGovRAI2022}, EU \cite{EuropeanCommission2021}, Japan \cite{JapanMIST2022}, and Saudi Arabia \cite{SaudiArabia2022}. Those that did not discuss robustness instead noted that the strongly related concept of reliability is critical for responsible AI systems \cite{UKGovernment2021, ChinaRAI2022, CSIRO2023, IndiaAI2022}.

Overall, robustness and reliability are considered across academia, industry, and governing bodies to be essential for the development of ethical and responsible AI.

\subsubsection{Transparency}
Transparency is the principle of ensuring that the decisions made by an AI system are able to be described and reproduced. For transparency to be achieved, descriptions should be accessible and understandable to users, developers, and other stakeholders.

The principle of transparency in RAI is largely agreed upon by academic, government, and industry sources. In terms of academic sources, three works name transparency as an essential pillar of RAI \cite{kaur2022, Li2023, Anagnostou2022}. Another work also discussed the need for transparency in the context of the overarching principle of ``intelligibility'' \cite{Ashok2022}. Each of these works note that transparency also supports the identification of issues within a model that might cause failures or biases.

In industry, transparency is often mentioned but individual definitions vary. Microsoft \cite{Microsoft2023}, IBM \cite{IBM2023}, and Samsung \cite{Samsung2023} each note transparency as a key RAI principle in the sense of a model being understandable to users, however Samsung stipulates that this applies only where it does not compromise corporate competitiveness. Amazon \cite{AmazonWebServices2022} state that models should be transparent to the extent that users can decide whether or how to use them, while Meta's policy \cite{MetaAI2021} focuses on transparency around how data is used. Google's stance on transparency is vague, mentioning that AI systems should feature ``relevant explanations'' and ``provide appropriate transparency and control over the use of data'', but this is in the context of accountability and privacy principles \cite{Google2023}.

Government agencies are in consensus regarding transparency - all eight considered government and political bodies noted transparency as a critical feature of responsible AI \cite{CSIRO2023, USGovRAI2022, UKGovernment2021, ChinaRAI2022, JapanMIST2022, SaudiArabia2022, EuropeanCommission2021, IndiaAI2022}. Most noted the need for transparency primarily for supporting auditability and for ensuring that end users know how decisions are being made about them by an AI system.

\subsubsection{Accountability}
Accountability of an AI system ensures that a model can justify its decisions. An AI system is accountable if its functionality and decisions can be explained to users, governing bodies, and other stakeholders to ensure compliance with laws and ethical standards. Accountability also means that AI should be subject to feedback and appeals from users and governing bodies, with mechanisms for remediation in place where issues arise.

In terms of academic literature, accountability is discussed extensively in all considered works \cite{Ashok2022, kaur2022, Li2023, Anagnostou2022}. One work notes that the principle of transparency can support accountability, however does not guarantee it \cite{kaur2022}. Academic literature emphasises the need for AI systems to be made accountable to ensure that they can be made liable for their decisions.

Industry sources also commonly include accountability in their principles, but implementations vary. Meta notes accountability and governance as critical for RAI, with internal self-governance measures in place along with processes for user appeals \cite{MetaAI2021}. Google explicitly states that RAI should ``be accountable to people'', further clarifying that this includes providing explanations and opportunities for feedback and appeal \cite{Google2023}. Conversely, Microsoft states that people should be accountable for AI systems \cite{Microsoft2023}. Amazon mentions the related principle of governance as key for RAI, but provides no clear guidelines on how this should be implemented \cite{AmazonWebServices2022}. IBM discusses accountability and governance extensively in their stance on AI ethics, noting that governance to ensure accountability should occur both internally and externally; however, accountability is not one of their five primary principles of RAI \cite{IBM2023}. Samsung does not include accountability or governance in their RAI framework \cite{Samsung2023}.

Accountability is also considered as a key principle of RAI by all government and political bodies considered, but descriptions of accountability vary. RAI principles from governing bodies in the US, China, India, Japan and Australia focus on holding people accountable for AI systems \cite{USGovRAI2022, JapanMIST2022, CSIRO2023, ChinaRAI2022, IndiaAI2022}. Meanwhile, holding AI itself accountable is the focus of accountability principles proposed by political bodies in the EU \cite{EuropeanCommission2021} and the UK \cite{UKGovernment2021}. RAI principles from Saudi Arabia's SADAI considers accountability of both people and technology, indicating that people across the AI supply chain should be accountable for AI systems, as should the system itself \cite{SaudiArabia2022}. RAI criteria from the US \cite{USGovRAI2022}, UK \cite{UKGovernment2021}, and Japan \cite{JapanMIST2022} consider transparency and accountability as a combined principle of RAI, while all other bodies consider accountability as an individual principle.

Overall, accountability is another prevalent pillar of RAI in the literature. It is clear that AI systems must be accountable for their decisions in order for systems and developers to meet legal and ethical requirements.

\subsubsection{Privacy}
Privacy in AI systems relates to ensuring that sensitive information used in training, validating, testing, or using the model remains private. An AI system should not be able to be `reverse engineered' to reveal private information about persons in the training data, nor should it be manipulable to reveal sensitive information about a person through malicious inputs.

Privacy is named as a key principle of RAI by all considered academic sources \cite{kaur2022, Li2023, Ashok2022, Anagnostou2022}. The definitions of privacy in all instances refer to ensuring that sensitive data are not revealed. Emphasis is broadly placed on ensuring that data is protected at all stages of the AI pipeline.

Similarly, all considered industry sources mention the concepts of privacy, data protection, or security in their RAI principles \cite{Google2023, AmazonWebServices2022, MetaAI2021, Samsung2023, Microsoft2023, IBM2023}. Implementations of privacy vary somewhat between companies. Meta specify that their privacy approach requires them ``assess privacy risks that involve the collection, use, or sharing of people’s information'' \cite{MetaAI2021}, however they do not specify how data sharing or collection will be disclosed. Conversely, Google notes the need for consent and notice of how data is used or shared \cite{Google2023}, and IBM notes the need for at least notice \cite{IBM2023}. Policies from Microsoft, Amazon and Samsung focus on preventing data breaches \cite{Microsoft2023, Samsung2023}.

All considered government and political sources also noted the need for privacy in RAI \cite{JapanMIST2022, ChinaRAI2022, CSIRO2023, UKGovernment2021, USGovRAI2022, SaudiArabia2022, EuropeanCommission2021, IndiaAI2022}. Privacy is directly named as an RAI principle by six bodies \cite{SaudiArabia2022, JapanMIST2022, ChinaRAI2022, CSIRO2023, EuropeanCommission2021, IndiaAI2022}, while the US NIST used the terminology `privacy-enhancing' \cite{USGovRAI2022}. The UK government emphasised the related concepts of data protection and security. All definitions focused on preserving individual privacy with strong emphasis on data security. The US NIST's documentation \cite{USGovRAI2022} also notes that ``privacy-related
risks may overlap with security, bias, and transparency,'' highlighting the challenges of ensuring transparency and fairness without compromising privacy.

Overall, the principle of privacy is well agreed upon - all considered academic, government, and industry sources note it as essential. However, it is worth noting that there are varied descriptions of \textit{how} privacy should be considered amongst industry sources.

\subsubsection{Safety}
Safety of an AI system is focused on ensuring that it does not harm people, environments, and societies. The need for safety to prevent harm is clear in a range of applications, from mental health chatbots to autonomous driving systems.

All considered literature sources discussed safety, to varying degrees \cite{kaur2022, Ashok2022, Anagnostou2022, Li2023}. One work \cite{kaur2022} considered the principles of robustness and safety together - arguing that a robust system would prevent harm by preventing failures. Other works discussed safety separately, with one noting the need for real-time protections \cite{Anagnostou2022} to mitigate safety issues should an AI fail.

Industry also broadly agrees that safety is a critical issue. Meta \cite{MetaAI2021} and Microsoft \cite{Microsoft2023} both group safety with either robustness or reliability in their frameworks. Guidelines presented by Samsung do not explicitly name safety, but do discuss limiting potential harm to humans, environments and ecosystems \cite{Samsung2023}. Google states that AI should ``be built and tested for safety'' \cite{Google2023}. Interestingly, neither IBM nor Amazon include safety in their guidelines \cite{IBM2023, AmazonWebServices2022}.

All government and political bodies discuss safety. The US's NIST names safety as an attribute of RAI \cite{USGovRAI2022}, while guidelines from China \cite{ChinaRAI2022}, Japan \cite{JapanMIST2022}, and the UK \cite{UKGovernment2021} also discuss safety or minimisation of harm. Australia's CSIRO \cite{CSIRO2023} addresses safety under the principle of ``Do no harm'', stating that civilian AI systems should be designed to minimise harm and negative outcomes. Government and political bodies from India \cite{IndiaAI2022}, the EU \cite{EuropeanCommission2021}, and Saudi Arabia \cite{SaudiArabia2022} all group safety with robustness or reliability, again highlighting a link between these two pillars.

Overall, safety is widely agreed to be critical for RAI. Interestingly, many sources grouped safety with robustness or reliability. However, an AI model could be robust without being safe if it is designed to do harm. As such, we argue that safety should be considered as a discrete pillar of RAI.

\subsubsection{Summary}
Our review of current academic, industrial, and government frameworks and policies on RAI has revealed that there are six critical pillars for RAI: Fairness, Robustness, Transparency, Accountabiilty, Privacy, and Safety.

Interestingly, we found that government and political bodies are leading the push for responsible AI. In all instances, government and political bodies had substantial guidelines in place for the development of responsible and ethical AI. Additionally, government bodies from a diverse group of regions all agreed on these six fundamental principles of RAI. 

Overall, this section has highlighted that these six principles are critical in ensuring that AI is developed, deployed, and used responsibly. However, a significant question remains: how do we address these pillars to create responsible AI systems? In the next section, we propose an answer: the use of XAI to underpin truly responsible AI.

\section{Explainable is Responsible}
In this section, we investigate the use of explainability to enable responsibility. In particular, we have presented strong evidence illustrating that explainability is foundational for each of the six pillars of responsibility identified in the previous section. Through our analysis, we determine that XAI is foundational for RAI; it would not be possible for RAI to exist without it.

\subsection{Fairness}
In order to achieve fairness in AI systems, developers must first be able to identify when bias or discrimination occur so that the problem can be rectified. Explainability techniques are highly suited to revealing a model's biases, both at global and local levels. Consider a home loan example: if explainability tools such as SHAP or LIME are applied and reveal that a person's gender was the largest contributor towards an AI system deciding to decline their home loan application, then this reveals a clear bias that must be corrected. This intuition has led to explainability being broadly considered in the literature for promoting development of fair AI systems.

One recent work examined the fairness of AI models including RF, KNN, and DTs trained to predict recidivism in juveniles convicted of a crime \cite{Miron2021}. Their analysis used LIME scoring alongside accuracy and fairness metrics, comparing models trained with only static inputs (those that cannot be changed, such as sex and nationality), only dynamic inputs (those that can be changed, such as social support and school performance), and a combination of both. Using LIME scoring, they identified that AI models made decisions largely based on static variables where these were available; the models were learning to be biased. Further analysis showed that disparity metrics also increased where static variables were included - specifically, the likelihood of someone being falsely identified as recidivist was significantly higher amongst men and foreigners. A trade-off between accuracy and fairness was identified, with the feature encoding strategy of learning fair representations (LFR) identified as a candidate for mitigating bias \cite{Zemel2013}. In this instance, the explanations provided by LIME were essential in revealing the bias of all developed AI models, and thus enabling bias mitigation strategies to be applied and assessed.

In the medical domain, one recent work proposed a framework based on their custom XAI tool for auditing medical diagnostics systems \cite{Panigutti2021}. Their proposed XAI tool was inspired by both LIME and anchors, generating a local `neighbourhood' of pertubations, and thereafter extracting rules to provide as output explanations. The overall framework initially calculates disparity scores for different groups (including groups based on sex, ethnicity, and insurance status) based on the Wasserstein distance between predictions and the ground truth. Their custom XAI tool is then utilised to explain system misclassifications. They suggest that this allows for assessment of fairness at two stages: firstly, by visualising the disparity scores and assessing whether any one group suffers from higher disparity than another, and secondly, but inspecting local explanations of misclassifications to assess whether the model demonstrated biases in making those predictions. Overall, their proposed system supports fairness in an auditable way.

In another recent work \cite{Nakao2022}, an interactive and explainable tool was developed for loan decision making. The prototype was based on a logistic regression model with interfaces that explain feature importances at global and local levels based on model weights, as well as enabling comparison to similar applications. The interactive component of the prototype allows non-technical end users to mark a decision as `fair' or `not fair', and adjust weightings on different features to see if a decision changes; these labels and suggested weight changes were then utilised to adjust the model. Prior `fair' and `not fair' labels are incorporated into explanations; users can see how many times the model's predictions were labelled as `fair' and `not fair' overall, as well as seeing whether similar applications were deemed to be fairly assessed by the model or not. Additionally, when the suggested weight changes were incorporated into the model, it was found to increase fairness based on disparate impact (DI) assessment. Overall, this is an interesting approach that shows promise in incorporating human feedback from end-users to improve the fairness of models, however the authors did caution that there is always a risk of people introducing their own biases to the model.

Another work that sought to enable human `fairness debugging' through a proposed system called Gopher \cite{Pradhan2022}, which seeks to quantify and explain model bias, utilising an approach of data subset removal and fairness metrics to analyse which types of data are causing biases and make suggestions on how to correct these biases. Explanations are made in the form of lists of patterns (groups of one or more co-occurring features) responsible for biases, with additional explanations provided as to possible updates or data removals that would reduce the bias of the model. The effectiveness of this approach was demonstrated on multiple well-known databases. 

Overall, the use of explainability for the purpose of supporting fairness has become established in the literature. Several works used explanation tools like LIME and SHAP to inspect whether particular features are being unfairly emphasised by a model. Meanwhile, several works propose systems that instead assess and explain fairness to developers and end-users, with the intention of receiving feedback and iterating the model to improve fairness. It is clear from the literature that XAI strongly supports the fairness principle of RAI.

\subsection{Robustness}
Robustness has been a key focus for many works utilising explainable AI. Explainability can help to reveal whether a model behaves consistently with similar inputs, as well as being robust against adversarial attacks \cite{Bai2021}.

In one recent work, a novel explainability technique entitled Similarity Difference and Uniqueness (SIDU) was proposed for producing heatmap explanations of image data that are robust against adversarial attacks \cite{Muddamsetty2022}. SIDU extracts the final layer of a CNN and calculates similarity and difference masks relating  to different feature applications. These are then fused to generate a final heatmap. In their experiments, SIDU produced explanations that were strongly aligned with human-grounded evaluation via eye tracking, outperforming prevalent literature models including Grad-CAM. To assess robustness, noise was introduced to input images, with CNN model predictions then explained using SIDU and competing literature models. SIDU explanations remained consistent as noise levels increased, while Grad-CAM explanations rapidly declined with increased noise. Their work demonstrates that explanations can be used to assess whether a model is robust against noise, and that explainability tools themselves can be enhanced to improve robustness.

Another recent work examined the relationship between robustness and explainability from a different angle, exploring whether robust models are inherently explainable \cite{Noack2021}. Their study used clean and adversarial data to train models for improved robustness, as well as training equivalent non-robust models with only clean data. They then examined how closely salience maps produced by the models matched target salience maps. Across two open-access databases, it was shown that salience maps for the robust models were more closely aligned with target salience maps than those of the non-robust models. This indicates that models designed to be robust have more inherently interpretable salience maps, potentially eliminating the need for post-hoc heatmap explanations where robustness is achieved.

Explainability has also been used to quantify robustness, with one study proposing a robustness metric based on counterfactual explanations \cite{SharmaCERTIFAI}. Their proposed Counterfactual Explanation-based Robustness Score (CERScore) operates on the principal that when comparing two AI models, the model where counterfactuals are further away on average from the input instances is the more robust model. CERScore is also model-agnostic and can be applied with no knowledge of the underyling architecture, a significant advantage over previous benchmark robustness metrics. Through a series of experiments on established AI models, the CERScore authors demonstrated that their proposed metric produced robustness scores consistent with literature benchmarks, indicating that explainability can be used to quantify robustness.

XAI has also been used to fine-tune models to improve their robustness. In one recent study, a DNN model was trained to identify malware using adversarial training and XAI-based fine tuning \cite{malik2022xai}. Following initial adversarial training, SHAP scores were calculated for the considered input features. A new database was then developed using input features and their corresponding SHAP scores to create a second adversarial database. This was then utilised to train the final DNN model. Testing showed that the XAI-based fine-tuning approach improved the accuracy of the model on unseen data, compared to using solely traditional adversarial training approaches.

Overall, recent literature has demonstrated that explainability can be utilised both to quantify robustness and to provide robustness. Both of these areas of robustness are critical to RAI. Quantification of robustness is a useful metric for RAI, as it allows comparison between candidate models. Additionally, the use of XAI to make models that are more robust is essential to RAI, as it supports development of models are resilient against adversarial attacks and invalid inputs.

\subsection{Transparency}
Transparency is perhaps the most self-evident characteristic of RAI that XAI can provide; the key purpose of XAI is to open the black-box of AI to improve transparency and comprehensibility \cite{Ding2022}. To achieve transparency, models must provide explanations that are interpretable and understandable to end-users. Transparency may include clarifying the inner workings of a black-box model, or explaining how a model made a decision in a way that users find acceptable and meaningful.

Several recent studies have sought to understand whether XAI explanations are transparent, interpretable, or understandable to end-users. In one such study, a counterfactual-based explanation strategy was developed to explain decisions made by a chest X-ray CNN classifier to expert radiologists \cite{Singla2023}. To assess the transparency of their approach, they utilised survey questions and free-text responses to assess whether explanations improved the understandability and trust in their proposed XAI system, compared to no explanation, saliency maps, and cycleGAN \cite{Zhu2017}. Based on a 5-point Likert scale, experts rated the counterfactual explanations the highest for understandability and justifiability. Additionally, all explainability tools trialled were rated significantly higher for understability than the no-explanation case. Several free-text comments also verified that the experts had a better understanding of the AI model following counterfactual explanation.

In another study, text-based explanations were generated by analysing the decision pathways of a DT model trained to assess behavioural anomalies and cognitive status in a smart home residential care setting \cite{Khodabandehloo2021}. Clinicians were asked to complete a Likert scale survey to rate their experience with the provided explanations. Questions about transparency were rated highly, with all clinicians agreeing that explanations were easily understandable and essential to understanding how the model classified anomalies for individual patients. Most clinicians reported that the explanations would assist them in determining whether assessments made by the model were correct.

In a study focusing on the education sector \cite{Aechtner2022}, a RF model was developed to make decisions on admissions to graduate school, with the decisions then explained using LIME, local SHAP, global SHAP, and PDP. University students with a range of AI experience were then surveyed using a Likert scale approach to determine whether explanations improved the understandability of the model. PDP and LIME were both found to significantly improve understandability amongst both AI novices and students with some AI experience. SHAP showed some improvement in understandability, but to a lesser extent. However, it is unclear which graphical format was used to present SHAP scores, so the chosen method may have contributed to the lower ranking.

Another recent study sought to understand what level of transparency gave users the most confidence in an autonomous vehicle's driving decisions \cite{Waa2020}. They developed a confidence metric derived from case-based reasoning, which utilised prior situations to determine an autonomous vehicle's confidence in a current situation. The model's confidence in a given situation  was presented to a non-expert cohort, using several different text structures to explain the confidence score. Survey participants were asked to rank which text structure they found most useful. Results of the survey found that the best-perceived explanation included the confidence level along with general information about prior situations. Explanations that provided confidence in addition to current or future situation information were also well received. The worst-ranked explanation was one that provided the confidence level only, with no transparency about how that confidence level was reached. These results indicate that transparency via text explanations of model confidence were critical to improving acceptability of the proposed AI approach.

Overall, recent literature that has sought to quantify transparency, understandability, or interpretability of AI systems have indicated a strong link between explanations and transparency. A wide range of explanation approaches have been examined by the literature, including text-based explanations, visual explanations, and feature-based explanations; each has been shown to improve transparency to some degree. Explanations have also been shown to support transparency for a wide range of users, with varied levels of domain and AI expertise.

\subsection{Accountability}
Accountability requires AI models to justify their decisions to ensure that they can be audited and assessed for compliance with legal and governance requirements. This principle is related to transparency, as models must be able to explain their decisions to be accountable.

One sector where accountability is critical is finance. In one recent study, an auditable pipeline based on various XAI techniques was proposed to provide transparent and accountable credit scoring \cite{Bucker2022}. The authors note that logistic regression is popular in credit risk scoring due to the inherent explainability of these models that has lead to acceptance amongst regulators, however note that logistic regression does not perform as strongly as advanced AI techniques for this task. To address this issue, they proposed a pipeline that begins with global explanations using feature permutation importance, followed by assessment of the most important features with PDP and ICE plots. From there, local instances are then able to be explained using LIME, SHAP, and another variant of SHAP. They indicate that this pipeline explains advanced AI algorithms in a manner that satisfies prevalent regulatory and legal requirements for transparency, accountability, and privacy. 

There is also a need for accountable AI systems to establish liability in legal cases. One recent work identifies XAI as highly suitable for providing a `forensic toolset' that can help establish liability of AI systems in a range of settings \cite{Padovan2023}. In particular, they illustrate a case study where an autonomous vehicle changes lanes to avoid a collision with car A, ultimately causing a collision with car B. The authors demonstrate that the use of XAI tools can help establish \textit{why} the AI model made the decision, thus establishing factual and legal causation as is required by many common law systems. Ultimately, this enables legal systems to hold AI systems accountable for their actions, and thereafter to determine whether the AI system in question or another party is at fault for an accident. It is suggested that LIME, ICE, and PDP are critical XAI tools for understanding the global model, while SHAP is essential to understand individual decisions; this is consistent with suggestions made by the credit scoring study discussed above \cite{Bucker2022}.

In the medical sector, an auditable pipeline was recently developed for prediction of oxygen requirement in COVID-19 patients, based on GradCAM explanations of features identified in chest x-ray input images \cite{Chung2022cxr}. Their technique included the development of model-derived atlases for similar feature patches and similar overall predictions, based on areas that GradCAM highlighted as important. When a prediction is made, the clinician is presented with 8 similar patches from the atlas alongside the test patch, a GradCAM heatmap of the overall x-ray, a table comparing the test x-ray with several similar x-rays both visually and with feature-based similarity scores, and finally a table showing the model's confidence in each feature it detects. Clinicians can also select feature labels of interest to receive further characterization of the given disease. The use of example-based explanations and numerical scores for confidence makes the decisions made by AI highly auditable and accountable, both to expert clinicians and regulatory bodies.

Overall, it is clear that accountability is critical in a wide range of sectors to ensure that AI models can be held to the same standards as humans. The literature indicates that XAI can provide an appropriate level of accountability and auditability across several key sectors, which contributes significantly towards the development of RAI systems.

\subsection{Privacy}
Privacy is critical for AI systems, as no sensitive information should be accessible to users of an AI system. Achieving privacy while maintaining transparency has previously been identified as a key challenge in developing responsible AI systems \cite{felzmann2020towards}. 

In standard use, most techniques for global and local explanation are inherently privacy-preserving as they do not explicitly reveal identifiable information about data in the training set. However, limited recent studies have identified that targeted adversarial attacks on CAM-based XAI models have a nominally increased chance of success in breaching privacy through training image reconstruction \cite{zhao2021} or model duplication \cite{Yan2023} compared to equivalent models without explanations. 

Critically, these attacks were successful against both explainable and non-explainable models; thus data protection and privacy-preserving techniques are essential for developing responsible AI. However, such techniques make AI more opaque and thus are in conflict with transparency and accountability principles for RAI. This brings us full-circle back to XAI: several recent works have demonstrated that XAI techniques continue to provide robust and meaningful explanations where data protection and privacy-preserving techniques have been applied, thus supporting privacy to be incorporated into RAI systems without compromising on other RAI principles. 

In one such work, federated learning was applied to develop a model for heart arrhythmia classification from electrocardiogram (ECG) signals before explainability was applied \cite{Raza2022}. Federated learning preserves privacy by enabling individual insitutions to train local models, with these models then shared to a central node for aggregation. The aggregated model is then returned to institutions for use. This enables the development of more robust models without direct data sharing, and thus protects against data inversion attacks. In the ECG study, federated learning was shown to improve accuracy in detecting key ECG features and classifying arrhythmia compared to local models, even where signals were noisy. Explanations were provided using GradCAM, providing heatmaps of the most important regions in a signal. Model duplication is still feasible, however reconstruction of local models is unlikely if weight sharing is protected.

In another work, swarm learning was applied to develop a model for cancer diagnosis from histopathology slides \cite{Saldanha2022}. Swarm learning is similar to federated learning in that models are trained locally first, however aggregation of models occurs by sharing trained local models in a peer-to-peer structure; this distributed and decentralized architecture ensures that the failure of a single node does not lead to failure of the entire network. The cancer histopathology work demonstrated that this enabled the development of swarm models that exceeded the performance of local models across all criteria of interest. To explain model predictions, heatmaps were generated to illustrate regions of interest that lead to a prediction. As with federated learning, this approach is robust against data inversion attacks as the final aggregate model has no knowledge of training data. The decentralized nature of swarm learning also ensures higher robustness in the event of a node failure.

Another approach to privacy preservation is the notion of differential privacy, where data is anonymized by injecting noise. One recent study explored whether this approach had a significant impact on SHAP values, using various open-access databases, data pertubation methods, and AI models \cite{Bozorgpanah2022}. Their findings showed that SHAP values were minimally affected by data protection techniques. Simple linear models were more affected than more advanced models such as support vector regression, however overall still provided meaningful explanations. Based on their findings, the authors concluded that data protection and explanation are not mutually exclusive.

Differential privacy was also considered in an image context by a recent work that used image compression to remove spatial and edge information in order to privatise data prior to training \cite{Gaudio2023}. Their work considered three image-based use cases: chest x-ray classification, cervix type classification, and glaucoma detection. Their work indicated that high accuracy could still be achieved with compressed images. Salience map explanations were also able to provide meaningful information about relevant areas of an image without access to the original uncompressed image. Overall, this approach improves privacy as original images are not used for model training.

Overall, XAI enables privacy in an indirect manner. AI systems are vulnerable to many adversarial attacks, and privacy-preserving techniques can compromise RAI principles such as transparency and accountability. XAI thus supports privacy by enabling privacy-preserving learning and data protection techniques to be implemented without violating the other key requirements of RAI.


\subsection{Safety}
Safety of humans, environments and societies is critical for AI systems, particularly in high-risk settings such as autonomous driving and healthcare. As XAI helps reveal the inner workings of an AI model, it can similarly assist in revealing safety risks. Additionally, XAI systems can be developed to sit above existing AI systems to provide safety in terms of detecting network intrusion or other attacks, protecting the overall pipeline from adversarial attacks and thus supporting RAI.

In one recent work, the need for XAI in ensuring safety in healthcare systems was examined through a case study on predicting when a patient should be extubated from mechanical ventilation in intensive care \cite{Jia2022xaihealthcare}. In this application, safety is of the utmost importance as early extubation can lead to emergency re-intubation, while late extubation increases risk of complications including pneumonia. Their work identified that XAI techniques are useful for verifying safety in several stages of AI development. In particular, they highlight that feature importance and counterfactual example methods support safety in AI. Feature importances can allow clinicians to assess whether the model utilised features that meet clinical expectations, improving confidence that the model makes decisions that will benefit the patient. Meanwhile, counterfactual explanations can be used to determine how many features would need to change (and by how much) in order to change the decision. This allows clinicians to evaluate the confidence of the model in its decision, which in turn supports patient safety as clinicians can make informed decisions about whether the model can be trusted in a particular instance or whether further investigation is needed.

In the autonomous driving domain, XAI has been applied in one recent work to compare the performance of three AI models classifying traffic signs where camera failures such as broken lens, ice coverage, or dead pixels occur \cite{Atif2022}. Their study evaluated the performance decreases for several types of lens failure, identifying that the model based on AlexNet \cite{AlexNet} was the strongest performer across most lens failure modes. To understand this further, LIME heatmaps were applied to traffic sign classification outputs. This illustrated that predictions made by the AlexNet model used features distributed over a larger section of the inputted image compared to the other models. The authors hypothesise that this leads to the higher camera failure tolerance of AlexNet, as the features of interest are less likely to be obscured by the defect if they are widely spread throughout the image. In this context, XAI supports safe autonomous driving by enabling identification of the traffic sign model that performs most reliably under adverse conditions.

Another recent work reviewed the field of goal-driven AI systems (GDAIs) \cite{Sado2023}, which include robots and other agents operating independently to fulfil their individual goals. In particular, they examined the need for explainable GDAIs in safety-critical applications to ensure that GDAIs do not cause harm to humans or environments. Their work identified that communication of actions via explanations such as CAM, SHAP, and textual description are essential for communicating the decisions of GDAIs. Through their analysis, the authors identify explanations support human-computer interaction, and thus support safety by enabling early detection of problems and thus early intervention to prevent failures, which is particularly crucial for safety-critical applications.

XAI can also be used to create a layer of protection against adversarial attacks on digital systems that employ AI. In one recent study, explainability techniques were utilised as part of an intrusion detection system for recognising attacks on Internet of Things (IoT) networks \cite{Houda2022}. Using a pipeline including rule-based explanations, LIME, and SHAP, the proposed intrusion detection system accurately detects intruder or compromised nodes on the IoT network, with global and local explanations provided to show how the decision was made so that human experts can decide on a response path. The use of XAI in this context helps to protect any AI models operating on the IoT network data from attack, as rapid and interpretable detection of intruders greatly reduces the risk of the IoT network being poisoned by malicious data injection.

Overall, explainability strongly supports safety across diverse AI applications. XAI techniques improve safety by enabling human experts to review decisions and decide whether they were made reasonably. Additionally, XAI can be applied on top of other AI systems to detect security risks such as network intrusions, reducing the risk of adversarial attacks. XAI is thus an important tool for ensuring that the RAI principle of safety is met. 

\subsection{Lessons Learned}
Through our exploration of the literature, it was found that XAI is foundational to each key pillar of RAI. The relationship of XAI to the principles of transparency and accountability was found to be straightforward. The primary intention of explanation is to make models more interpretable, and several works have established that users perceive XAI to be more transparent, understandable, and trustworthy \cite{Aechtner2022, Ding2022, Singla2023}. Transparency is also the first step towards accountability, where AI models need to be able to explain and justify their decisions to relevant stakeholders. Works to date have established that XAI is essential in making AI accountable and, where necessary, liable for its actions in critical sectors from autonomous driving to healthcare \cite{Padovan2023, Chung2022cxr}.

Accountability is also connected to the principles of fairness and safety. In our exploration of the literature, it was found that XAI tools can highlight societal biases learned from historic data \cite{Panigutti2021, Nakao2022, Miron2021, Yogarajan2022}. Human-in-the-loop fairness debugging was also discussed \cite{Pradhan2022, Nakao2022}, with human users able to provide feedback on potential biases based on model explanations, as a form of fairness accountability. Human-in-the-loop debugging was also considered in the safety context \cite{Sado2023}, with explanations provided by goal-driven AIs to human safety auditors. In turn, humans could provide feedback to the AI models that would improve the safety and failure tolerance of the systems. Similarly, safety-critical applications in health \cite{Jia2022xaihealthcare} and autonomous driving \cite{Atif2022} were supported by XAI through the provision of explanations to domain experts, who could then assess how and why the model is making a decision and determine whether that decision is appropriate.

Safety is also tied to the principle of robustness; a model that is less likely to fail, is less likely to cause harm in safety-critical applications. XAI was found to support the RAI principle of robustness in several direct and indirect ways. In particular, XAI has been used quantify robustness \cite{SharmaCERTIFAI}, enable robustness comparison between models \cite{Muddamsetty2022}, and support fine-tuning of models to enhance robustness \cite{malik2022xai}. Interestingly, one study also identified that robust models have a higher degree of inherent explainability \cite{Noack2021}, indicating a strong relationship between these principles.

In our analysis of XAI robustness literature, one study highlighted that certain explanations remain accurate in the face of adversarial attacks that utilise perturbed inputs to compromise a model \cite{Muddamsetty2022}. The resilience of XAI to adversaries was considered by several works seeking to improve privacy. In some works, it was found that XAI could still successfully explain decisions where user privacy was protected by perturbing the dataset such that adversarial attacks would not be able to unveil genuine and identifiable information \cite{Bozorgpanah2022, Gaudio2023}. Another key approach for preserving privacy was distributed learning, both in federated \cite{Raza2022} and swarm learning structures \cite{Saldanha2022}, where local models are trained on private data, shared securely, and then aggregated into a final model for the end-user. In this context, XAI supported privacy indirectly by enabling models to continue meeting other RAI requirements whilst using an otherwise opaque method of learning.

In this section, we identified that research into XAI and RAI is an active field that remains in its infancy. The works highlighted have firmly established that XAI and RAI are inseparable concepts, however much remains to be explored in how best to utilise XAI to create truly responsible AI systems. Notably, relatively few works sought to develop frameworks or metrics for assessing the quality of explanations as they pertain to AI responsibility. Additionally, many XAI works to date considered only a subset of the six principles of responsible AI. Thus, there is still a need to create AI systems that are truly responsible in all attributes. XAI for RAI is a significant and growing field, and one which remains open for further exploration.

\section{Use Cases}
Having now established that XAI is foundational to RAI, this section presents several short case studies that illustrate how explainability can be utilised to create responsible AI. In particular, we present case studies in content generation, healthcare, and transport - domains where responsibility is essential to ensure that AI acts for the benefit of humanity and minimising risk of harm.

\subsection{Generative AI}
Generative AI (GAI) has recently become a prevalent topic in the literature and media, largely due to the emergence of large-language generative models such as ChatGPT. GAI has previously been found to exhibit biases \cite{Alba2022, Wakefield2016}, produce disinformation \cite{Hsu2022} or misleading information \cite{VanDis2023}. These issues are barriers to GAI meeting RAI requirements, and can pose substantial harm to users.

Explainable AI could be leveraged to make GAI systems more responsible. Consider a user who asks a GAI model a question about clinical depression. The model might provide information about symptoms and treatment paths, however without explanation the user cannot ascertain that the information is correct. XAI could be used to explain GAI outputs by providing local explanations that highlight how it understood the prompt, similar to previous research that used LIME for text analysis tasks \cite{Uddin2022depression}. This would allow users to assess whether their query has been appropriately understood by the GAI model. 

In terms of the generated output, text-based explanations where the GAI model generates an output \textit{and} an explanation of how it chose this output are suitable. This approach has been highlighted by one early work in the GAI/XAI space \cite{Khurana2021}. Case-based reasoning can also be used to evaluate the similarity between the users prompts, previous prompts, and legitimate sources to assign confidence weightings to different sections of its generated output. For example, a GAI model responding to a question about clinical depression may have higher confidence in some symptoms than others based on its training set, and this could be highlighted using colour coding - similar to a previous approach utilised in decision-making systems \cite{Waa2020}. Additionally, example-based approaches could be utilised by GAI models to effectively cite their sources. Example-based approaches have previously been utilised in healthcare applications such as chest x-ray assessment to provide examples of similar chest x-rays and their diagnoses \cite{Chung2022cxr}. In the GAI context, example-based explainability could highlight relevant excerpts from the literature - in our clinical depression example, the GAI model could list five sources that describe the symptom of `fatigue' when a user hovers over the word. 

Having GAI models explain their outputs supports responsibility across all key pillars. For the user, explanation of prompt understanding and output generation provides transparency that supports them in identifying whether information is correct. This can enhance user safety and interpretation in many settings. From the developer perspective, explanations make the model auditable for RAI attributes so that it can be held accountable for its actions and continuously improved. For example, explanations of prompts would allow model auditors to phrase a question in many different ways, assessing model robustness to phrasing changes. Additionally, auditors could examine explanations of prompts and outputs for signs of unfair bias, dangerous information or advice, and privacy breaches such as reconstructing sensitive data from the training set. Using XAI to ensure transparency and accountability in GAI can thus support iterative development to enhance the fairness, privacy, safety, and robustness of these models.

Overall, GAI has significant potential to cause harm if not developed responsibly. In this case study, we have shown how XAI can serve as the foundation for responsible GAIs that better serve humanity. This field is novel with few works published to date, offering much opportunity for future research. 

\subsection{Medical Diagnostics}
The field of medical diagnostics is one where AI holds much potential. In some scenarios, AI has been shown to outperform expert clinicians \cite{Jiang2023}, partly due to their ability to learn from large datasets. Responsible AI development is critical in healthcare due to the sensitive nature of the underlying data, and the potential for significant harm to humans if incorrect decisions are made. 

One potential source for harm in medical AI models arises from historic databases reflecting societal biases. Without caution, AI models can learn the same biases and thus make potentially harmful decisions. Consider the case of attention-deficit/hyperactivity disorder (ADHD). Due to societal biases, ADHD has historically been underdiagnosed in women and racial minorities \cite{Garb2021}. ADHD diagnosis with AI is an active field of research that has shown promising results, however a recent work cautioned that certain inputs can introduce unfair bias into models \cite{Loh2022}. XAI can assist in ensuring fairness in such models, with global and local explanations helping to reveal if the model places weight on parameters that perpetuate societal biases. For feature-based data types, explanations such as counterfactuals, SHAP, and LIME are strong candidates, while CAM would be suitable for image data. 

Similarly, model robustness can be assessed using XAI techniques. By quantifying how significantly the data would need to change for the prediction to change using a technique such as CERSCore \cite{SharmaCERTIFAI}, the robustness of a model can be assessed. This also supports fairness and safety, as robust models will not make decisions based on one or two excessively weighted features.

Global and local explanations also make the model inherently more transparent, and support accountability. Clinicians and regulatory bodies can review the decisions made by AI models to ensure that the model operates in accordance with clinical knowledge and legal requirements. For individual patients, clinicians can utilise local explanations to understand the confidence of the model, and thus make assessment on the correctness of the decision. This also improves patient safety, as the interaction between AI and clinicians further supports accurate ADHD diagnosis.

In the medical context, the concepts of safety and privacy are intertwined due to the sensitive nature of medical data. Federated learning and swarm learning are of interest in the medical space \cite{Raza2022, Saldanha2022} as these methods eliminate the need for data sharing between institutions; however, distributed learning makes models more opaque. Fortunately, XAI still provides robust, transparent and meaningful explanations where distributed learning has been utilised - enabling privacy-preserving training to be implemented without compromising other responsibility requirements. 

Overall, XAI is critical for RAI in healthcare, due to the direct impact of AI decisions on human life. In this case study, we have illustrated how XAI can be utilised to develop an RAI system for ADHD diagnosis. The concepts discussed here are applicable to many other diagnostic use cases.

\subsection{Autonomous Vehicles}
Autonomous vehicles (AVs), vessels and aircraft pose significant risk to human life if not developed responsibly. AI systems in this domain must be able to make rapid decisions, and poor decisions can lead to injury or loss of life - both for passengers and bystanders. 

XAI is essential for ensuring that AVs are piloted by RAIs. Transparency about decision making on the road can help to support user trust, and enable the human driver to regain manual control if needed. For example, if an AV speeds up after seeing a new speed limit sign, this decision could be explained to the user by showing the sign and decision, and highlighting the regions that lead to the decision using an approach such as CAM. If the decision is incorrect, the user could make manual speed adjustment and report the error to the car manufacturer. Similarly, feature-based explanations of speed changes based on weather conditions using LIME or a similar method would support the user in assessing the suitability of the decision made.

This transparency makes the AV accountable to the user and to car manufacturers. XAI can further ensure accountability when events such as collisions resulting in injury occur, as explanations of how and why parameters were used to make a decision can aid in determining whether the AI model acted reasonably and in line with legal requirements \cite{Padovan2023}.

Explanations used to ensure transparency and accountability also support robustness and safety. Users can make informed decisions about whether the AV has made a good choice, and feedback can be provided to vehicle manufacturers for iterative improvement to enhance robustness and safety. Legal accountability further encourages manufacturers to ensure that their systems are resilient and safe. During the development stage, XAI can also be utilised to compare the robustness of AI models, to identify the system that makes the safest decisions in various contexts \cite{Atif2022}.

Fairness in the AV context has primarily been considered in terms of the Moral Machine Experiment (MME) \cite{Awad2018}, which placed humans in the metaphorical driver's seat of an AV that was guaranteed to collide with someone - and asked them to choose which person or group of people the car should hit. This experiment revealed human preferences contain biases based on attributes such as sex, age, and societal status. Thus, AVs are at risk of inheriting biases that could put life at risk. While the MME presents an extreme case, it is clear that there is a need to explain the decisions made by AVs - both for legal accountability purposes in extreme events, and to ensure that we are not sharing the road with vehicles that are prejudiced. Explained decisions can be evaluated for any potential biases, and this in turn can be used to iteratively improve the fairness of systems.

The final attribute to consider for RAIs in autonomous driving is privacy. Large datasets are essential for ensuring that AI models for autonomous driving perform well, however many companies would be unwilling to share proprietary data. Swarm learning may be more acceptable to companies, as they could share local models in a peer-to-peer structure with only companies they choose. Another aspect of privacy is user privacy; users will have varying preferences regarding the sharing of sensitive information such as location. Fortunately, this can largely be addressed by implementing differential privacy techniques. XAI again supports privacy by ensuring that all other attributes of RAI continue to be met when privacy-preserving techniques are implemented.

Overall, the need for RAI to drive AVs is clear. AVs are making continuous and rapid decisions, and many of their decisions can mean life or death for passengers and bystanders. The transparency provided by XAI is essential for investigating decisions that AVs make to ensure that they operate robustly, safely, and fairly. This also enables AVs to be held accountable for their actions should accidents occur. XAI also responsible AVs by enabling privacy-preserving techniques to be implemented without compromising on transparency and accountability. In this case study, we further illustrate that XAI and RAI are inseparable concepts. When AI is behind the wheel, it is essential that XAI techniques are in the passenger seat. 

\section{Opportunities for Future Work}
This review of the literature has clearly demonstrated that XAI is foundational to the development of RAI systems. However, this field remains in its infancy, with much opportunity for future research. In this section, we outline several key directions for future research in this critical domain.

\textit{Quantifying responsibility:} One notable direction for future research is in quantifying responsibility to enable comparison between different AI systems. Some early works have proposed metrics for assessing attributes such as fairness \cite{Pradhan2022} and robustness \cite{SharmaCERTIFAI}, however these have not been widely validated. Similarly, some works have sought to assess transparency by surveying users of XAI models \cite{Singla2023, Khodabandehloo2021, Aechtner2022, Waa2020}, however each of these works used different surveys. A significant gap remains in terms of developing clear frameworks, guidelines, or metrics for assessing transparency. Similarly, metrics or methods for assessing accountability, privacy, and safety remain open research topics. Development of clear metrics or methods for quantifying responsibility would be a valuable addition to the literature. Validated strategies for quantifying responsibility could then be utilised to assess different XAI techniques to determine which are the most responsible, and subsequently develop new XAI techniques that enhance responsibility further.

\textit{Improving explanations:} The field of XAI has been rapidly growing, however relatively few studies have sought to analyse whether the explanations provided by their proposed techniques are understandable and acceptable to humans users. These characteristics are essential in ensuring that XAI tools are contributing to responsibility. There remains a significant gap in the literature in determining how humans interpret data, and subsequently developing XAI systems that are in line with human understanding. The development of databases and methods for capturing human attention to data would greatly support research in this area. 

\textit{Considering all pillars of responsibility:} In this review, it is clear that most works on XAI for RAI focus on only one pillar of responsibility. Future research would benefit from utilising XAI techniques to improve responsibility in more than one area. The case studies presented have highlighted that this is feasible, particularly given certain XAI techniques and approaches can address many pillars of RAI simultaneously. While there certainly remains a need for ongoing research that seeks to improve responsibility with respect to certain pillars, developers of AI systems should seek to meet all pillars of RAI using XAI tools, as AI cannot be truly responsible if it addresses some pillars of RAI at the expense of others.

\textit{Responsible GAI:} In terms of domain-specific research, our review indicates that explainability and responsibility of generative AI is under-explored. Given the prevalence and controversy surrounding generative AI systems such as ChatGPT in the literature and popular media, there is a significant research opportunity in implementing and continuing to develop XAI techniques that will lay the foundation for responsible GAI models that incorporate all pillars of RAI.

Overall, the field of XAI for RAI offers significant opportunities to future researchers. Research to date has illustrated that XAI is foundational to RAI, however there remains much research opportunity in areas such as quantifying responsibility, enhancing explanations to be more responsible, developing RAI models that address all pillars of responsibility, and finally applying XAI techniques to support development of RAI in rapidly evolving and exciting fields such as generative AI.

\section{Conclusion}
In this work, we conducted a scoping review of current XAI and RAI literature, and presented an evidence-based argument that XAI is foundational to RAI across all key pillars of responsibility. Our review began with XAI, exploring methods for explainability by design and post-hoc explainability. We identified approaches for feature-based, image-based, and text-based explanations at both local and global levels, with illustrative examples provided for each prevalent method.

We then examined the literature on RAI, considering sources from academic literature, government and political bodies, and large technology companies. Based on our findings, we identified six key pillars of responsibility: fairness, robustness, transparency, accountability, privacy, and safety. Interestingly, it was also identified that governments are leading academia and industry on the push for responsible AI, with diverse government and political bodies from across the globe having substantial guidelines in place for the development of responsible AI.

With the fields of XAI and RAI established, our review then explored the relationship between XAI and RAI. Based on the evidence found, we identified that XAI is foundational to all pillars of responsibility. XAI supports AI systems to be more transparent and accountable without compromising privacy, which in turn supports the principles of fairness, robustness, and safety. It is therefore clear that XAI is the essential foundation for truly responsible AI.

To further illustrate the importance of XAI for RAI in practical contexts, we present several timely use cases. We first showed how XAI is essential for developing responsible GAI models that can explain their understanding of a prompt and how they produced their outputs. Next, we turned to the medical domain and highlighted how XAI can be utilised to develop RAI diagnostics models that overcome historical biases and better serve humanity. Finally, we explored the use of XAI to ensure that autonomous vehicles act responsibly, highlighting how they can use XAI to communicate their decisions with passengers, manufacturers, and legal authorities.

Based on our thorough review of cutting-edge research on XAI as the foundational tool of RAI, we next presented a series of future research directions informed by our lessons learned. It was identified that this emerging field has much research potential, with opportunities present in developing methods and metrics to quantify responsibility, further improvements to XAI for RAI, development of systems that are responsible across all key pillars of RAI, and lastly in the application of XAI techniques to novel and exciting domains, including generative AI.



\bibliographystyle{ACM-Reference-Format}
\bibliography{references.bib}


\begin{thebibliography}{168}


\ifx \showCODEN    \undefined \def \showCODEN     #1{\unskip}     \fi
\ifx \showDOI      \undefined \def \showDOI       #1{#1}\fi
\ifx \showISBNx    \undefined \def \showISBNx     #1{\unskip}     \fi
\ifx \showISBNxiii \undefined \def \showISBNxiii  #1{\unskip}     \fi
\ifx \showISSN     \undefined \def \showISSN      #1{\unskip}     \fi
\ifx \showLCCN     \undefined \def \showLCCN      #1{\unskip}     \fi
\ifx \shownote     \undefined \def \shownote      #1{#1}          \fi
\ifx \showarticletitle \undefined \def \showarticletitle #1{#1}   \fi
\ifx \showURL      \undefined \def \showURL       {\relax}        \fi
\providecommand\bibfield[2]{#2}
\providecommand\bibinfo[2]{#2}
\providecommand\natexlab[1]{#1}
\providecommand\showeprint[2][]{arXiv:#2}

\bibitem[Adadi and Berrada(2018)]%
        {Adadi2018}
\bibfield{author}{\bibinfo{person}{A Adadi} {and} \bibinfo{person}{M Berrada}.} \bibinfo{year}{2018}\natexlab{}.
\newblock \showarticletitle{{Peeking Inside the Black-Box: A Survey on Explainable Artificial Intelligence (XAI)}}.
\newblock \bibinfo{journal}{\emph{IEEE Access}}  \bibinfo{volume}{6} (\bibinfo{year}{2018}), \bibinfo{pages}{52138--52160}.
\newblock
\showISSN{2169-3536 VO - 6}
\urldef\tempurl%
\url{https://doi.org/10.1109/ACCESS.2018.2870052}
\showDOI{\tempurl}


\bibitem[Aechtner et~al\mbox{.}(2022)]%
        {Aechtner2022}
\bibfield{author}{\bibinfo{person}{J Aechtner}, \bibinfo{person}{L Cabrera}, \bibinfo{person}{D Katwal}, {and} others.} \bibinfo{year}{2022}\natexlab{}.
\newblock \showarticletitle{{Comparing User Perception of Explanations Developed with XAI Methods}}. In \bibinfo{booktitle}{\emph{2022 IEEE International Conference on Fuzzy Systems}}. \bibinfo{pages}{1--7}.
\newblock
\showISBNx{1558-4739 VO -}
\urldef\tempurl%
\url{https://doi.org/10.1109/FUZZ-IEEE55066.2022.9882743}
\showDOI{\tempurl}


\bibitem[Afshar et~al\mbox{.}(2022)]%
        {Afshar2022}
\bibfield{author}{\bibinfo{person}{F Afshar}, \bibinfo{person}{S Seyedabrishami}, {and} \bibinfo{person}{S Moridpour}.} \bibinfo{year}{2022}\natexlab{}.
\newblock \showarticletitle{{Application of Extremely Randomised Trees for exploring influential factors on variant crash severity data}}.
\newblock \bibinfo{journal}{\emph{Scientific Reports}} \bibinfo{volume}{12}, \bibinfo{number}{1} (\bibinfo{year}{2022}), \bibinfo{pages}{11476}.
\newblock
\urldef\tempurl%
\url{https://doi.org/10.1038/s41598-022-15693-7}
\showDOI{\tempurl}


\bibitem[Ahmed et~al\mbox{.}(2022)]%
        {Ahmed2022}
\bibfield{author}{\bibinfo{person}{I Ahmed}, \bibinfo{person}{G Jeon}, {and} \bibinfo{person}{F Piccialli}.} \bibinfo{year}{2022}\natexlab{}.
\newblock \showarticletitle{{From Artificial Intelligence to Explainable Artificial Intelligence in Industry 4.0: A Survey on What, How, and Where}}.
\newblock \bibinfo{journal}{\emph{IEEE Transactions on Industrial Informatics}} \bibinfo{volume}{18}, \bibinfo{number}{8} (\bibinfo{year}{2022}), \bibinfo{pages}{5031--5042}.
\newblock
\showISSN{1941-0050 VO - 18}
\urldef\tempurl%
\url{https://doi.org/10.1109/TII.2022.3146552}
\showDOI{\tempurl}


\bibitem[Ahsan et~al\mbox{.}(2021)]%
        {Ahsan2021}
\bibfield{author}{\bibinfo{person}{M~M Ahsan}, \bibinfo{person}{R Nazim}, \bibinfo{person}{Z Siddique}, {and} \bibinfo{person}{P Huebner}.} \bibinfo{year}{2021}\natexlab{}.
\newblock \showarticletitle{{Detection of covid-19 patients from ct scan and chest x-ray data using modified mobilenetv2 and lime}}.
\newblock \bibinfo{journal}{\emph{Healthcare (Switzerland)}} \bibinfo{volume}{9}, \bibinfo{number}{9} (\bibinfo{year}{2021}), \bibinfo{pages}{1099}.
\newblock
\urldef\tempurl%
\url{https://doi.org/10.3390/healthcare9091099}
\showDOI{\tempurl}


\bibitem[Alam and Mueller(2021)]%
        {Alam2021}
\bibfield{author}{\bibinfo{person}{Lamia Alam} {and} \bibinfo{person}{Shane Mueller}.} \bibinfo{year}{2021}\natexlab{}.
\newblock \showarticletitle{{Examining the effect of explanation on satisfaction and trust in AI diagnostic systems}}.
\newblock \bibinfo{journal}{\emph{BMC Medical Informatics and Decision Making}} \bibinfo{volume}{21}, \bibinfo{number}{1} (\bibinfo{year}{2021}), \bibinfo{pages}{178}.
\newblock
\showISSN{1472-6947}
\urldef\tempurl%
\url{https://doi.org/10.1186/s12911-021-01542-6}
\showDOI{\tempurl}


\bibitem[Alba(2022)]%
        {Alba2022}
\bibfield{author}{\bibinfo{person}{Davey Alba}.} \bibinfo{year}{2022}\natexlab{}.
\newblock \bibinfo{title}{{OpenAI Chatbot Spits Out Biased Musings, Despite Guardrails}}.
\newblock
\newblock
\urldef\tempurl%
\url{https://www.bloomberg.com/news/newsletters/2022-12-08/chatgpt-open-ai-s-chatbot-is-spitting-out-biased-sexist-results}
\showURL{%
\tempurl}


\bibitem[{Amazon Web Services}(2022)]%
        {AmazonWebServices2022}
\bibfield{author}{\bibinfo{person}{{Amazon Web Services}}.} \bibinfo{year}{2022}\natexlab{}.
\newblock \bibinfo{booktitle}{\emph{{Responsible Use of Machine Learning}}}.
\newblock \bibinfo{type}{{T}echnical {R}eport}.
\newblock
\urldef\tempurl%
\url{https://d1.awsstatic.com/responsible-machine-learning/responsible-use-of-machine-learning-guide.pdf}
\showURL{%
\tempurl}


\bibitem[Amri et~al\mbox{.}(2022)]%
        {Amri2022}
\bibfield{author}{\bibinfo{person}{E Amri}, \bibinfo{person}{P Dardouillet}, \bibinfo{person}{A Benoit}, {and} others.} \bibinfo{year}{2022}\natexlab{}.
\newblock \showarticletitle{{Offshore Oil Slick Detection: From Photo-Interpreter to Explainable Multi-Modal Deep Learning Models Using SAR Images and Contextual Data}}.
\newblock \bibinfo{journal}{\emph{Remote Sensing}} \bibinfo{volume}{14}, \bibinfo{number}{15} (\bibinfo{year}{2022}), \bibinfo{pages}{3565}.
\newblock
\urldef\tempurl%
\url{https://doi.org/10.3390/rs14153565}
\showDOI{\tempurl}


\bibitem[Anagnostou et~al\mbox{.}(2022)]%
        {Anagnostou2022}
\bibfield{author}{\bibinfo{person}{Marianna Anagnostou}, \bibinfo{person}{Olga Karvounidou}, \bibinfo{person}{Chrysovalantou Katritzidaki}, {and} others.} \bibinfo{year}{2022}\natexlab{}.
\newblock \showarticletitle{{Characteristics and challenges in the industries towards responsible AI: a systematic literature review}}.
\newblock \bibinfo{journal}{\emph{Ethics and Information Technology}} \bibinfo{volume}{24}, \bibinfo{number}{3} (\bibinfo{year}{2022}), \bibinfo{pages}{37}.
\newblock
\showISSN{1572-8439}
\urldef\tempurl%
\url{https://doi.org/10.1007/s10676-022-09634-1}
\showDOI{\tempurl}


\bibitem[Anand et~al\mbox{.}(2021)]%
        {Anand2021}
\bibfield{author}{\bibinfo{person}{Deepak Anand}, \bibinfo{person}{Kumar Yashashwi}, \bibinfo{person}{Neeraj Kumar}, {and} others.} \bibinfo{year}{2021}\natexlab{}.
\newblock \showarticletitle{{Weakly supervised learning on unannotated H\&E-stained slides predicts BRAF mutation in thyroid cancer with high accuracy}}.
\newblock \bibinfo{journal}{\emph{The Journal of Pathology}} \bibinfo{volume}{255}, \bibinfo{number}{3} (\bibinfo{date}{nov} \bibinfo{year}{2021}), \bibinfo{pages}{232--242}.
\newblock
\showISSN{0022-3417}
\urldef\tempurl%
\url{https://doi.org/10.1002/path.5773}
\showDOI{\tempurl}


\bibitem[Arous et~al\mbox{.}(2021)]%
        {Arous2021}
\bibfield{author}{\bibinfo{person}{I Arous}, \bibinfo{person}{L Dolamic}, \bibinfo{person}{J Yang}, {and} others.} \bibinfo{year}{2021}\natexlab{}.
\newblock \showarticletitle{{MARTA: Leveraging Human Rationales for Explainable Text Classification}}. In \bibinfo{booktitle}{\emph{35th AAAI Conference on Artificial Intelligence, AAAI 2021}}, Vol.~\bibinfo{volume}{7}. \bibinfo{publisher}{AAAI Press}, \bibinfo{address}{Vancouver, Canada}, \bibinfo{pages}{5868--5876}.
\newblock
\urldef\tempurl%
\url{https://www.scopus.com/inward/record.uri?eid=2-s2.0-85129363314&partnerID=40&md5=56f214b67454396f0aea9aa933bad621}
\showURL{%
\tempurl}


\bibitem[Ashok et~al\mbox{.}(2022)]%
        {Ashok2022}
\bibfield{author}{\bibinfo{person}{Mona Ashok}, \bibinfo{person}{Rohit Madan}, \bibinfo{person}{Anton Joha}, {and} \bibinfo{person}{Uthayasankar Sivarajah}.} \bibinfo{year}{2022}\natexlab{}.
\newblock \showarticletitle{{Ethical framework for Artificial Intelligence and Digital technologies}}.
\newblock \bibinfo{journal}{\emph{International Journal of Information Management}}  \bibinfo{volume}{62} (\bibinfo{year}{2022}), \bibinfo{pages}{102433}.
\newblock
\showISSN{0268-4012}
\urldef\tempurl%
\url{https://doi.org/10.1016/j.ijinfomgt.2021.102433}
\showDOI{\tempurl}


\bibitem[Aslam et~al\mbox{.}(2022)]%
        {Aslam2022}
\bibfield{author}{\bibinfo{person}{Nida Aslam}, \bibinfo{person}{Irfan~Ullah Khan}, \bibinfo{person}{Aisha Alansari}, {and} others.} \bibinfo{year}{2022}\natexlab{}.
\newblock \showarticletitle{{Anomaly Detection Using Explainable Random Forest for the Prediction of Undesirable Events in Oil Wells}}.
\newblock \bibinfo{journal}{\emph{Applied Computational Intelligence and Soft Computing}}  \bibinfo{volume}{2022} (\bibinfo{year}{2022}), \bibinfo{pages}{1558381}.
\newblock
\showISSN{1687-9724}
\urldef\tempurl%
\url{https://doi.org/10.1155/2022/1558381}
\showDOI{\tempurl}


\bibitem[Atif et~al\mbox{.}(2022)]%
        {Atif2022}
\bibfield{author}{\bibinfo{person}{M Atif}, \bibinfo{person}{A Ceccarelli}, \bibinfo{person}{T Zoppi}, {and} others.} \bibinfo{year}{2022}\natexlab{}.
\newblock \showarticletitle{{Robust Traffic Sign Recognition Against Camera Failures}}.
\newblock \bibinfo{journal}{\emph{IEEE Open Journal of Intelligent Transportation Systems}}  \bibinfo{volume}{3} (\bibinfo{year}{2022}), \bibinfo{pages}{709--722}.
\newblock
\showISSN{2687-7813 VO - 3}
\urldef\tempurl%
\url{https://doi.org/10.1109/OJITS.2022.3213183}
\showDOI{\tempurl}


\bibitem[Awad et~al\mbox{.}(2018)]%
        {Awad2018}
\bibfield{author}{\bibinfo{person}{Edmond Awad}, \bibinfo{person}{Sohan Dsouza}, \bibinfo{person}{Richard Kim}, {and} others.} \bibinfo{year}{2018}\natexlab{}.
\newblock \showarticletitle{{The Moral Machine experiment}}.
\newblock \bibinfo{journal}{\emph{Nature}} \bibinfo{volume}{563}, \bibinfo{number}{7729} (\bibinfo{year}{2018}), \bibinfo{pages}{59--64}.
\newblock
\showISSN{1476-4687}
\urldef\tempurl%
\url{https://doi.org/10.1038/s41586-018-0637-6}
\showDOI{\tempurl}


\bibitem[Ayling and Chapman(2022)]%
        {Ayling2022}
\bibfield{author}{\bibinfo{person}{Jacqui Ayling} {and} \bibinfo{person}{Adriane Chapman}.} \bibinfo{year}{2022}\natexlab{}.
\newblock \showarticletitle{{Putting AI ethics to work: are the tools fit for purpose?}}
\newblock \bibinfo{journal}{\emph{AI and Ethics}} \bibinfo{volume}{2}, \bibinfo{number}{3} (\bibinfo{year}{2022}), \bibinfo{pages}{405--429}.
\newblock
\showISSN{2730-5961}
\urldef\tempurl%
\url{https://doi.org/10.1007/s43681-021-00084-x}
\showDOI{\tempurl}


\bibitem[Ayoub et~al\mbox{.}(2021)]%
        {Ayoub2021}
\bibfield{author}{\bibinfo{person}{J Ayoub}, \bibinfo{person}{X~J Yang}, {and} \bibinfo{person}{F Zhou}.} \bibinfo{year}{2021}\natexlab{}.
\newblock \showarticletitle{{Combat COVID-19 infodemic using explainable natural language processing models}}.
\newblock \bibinfo{journal}{\emph{Information Processing and Management}} \bibinfo{volume}{58}, \bibinfo{number}{4} (\bibinfo{year}{2021}), \bibinfo{pages}{102569}.
\newblock
\urldef\tempurl%
\url{https://doi.org/10.1016/j.ipm.2021.102569}
\showDOI{\tempurl}


\bibitem[Bai et~al\mbox{.}(2021)]%
        {Bai2021}
\bibfield{author}{\bibinfo{person}{Xiao Bai}, \bibinfo{person}{Xiang Wang}, \bibinfo{person}{Xianglong Liu}, {and} others.} \bibinfo{year}{2021}\natexlab{}.
\newblock \showarticletitle{{Explainable deep learning for efficient and robust pattern recognition: A survey of recent developments}}.
\newblock \bibinfo{journal}{\emph{Pattern Recognition}}  \bibinfo{volume}{120} (\bibinfo{year}{2021}), \bibinfo{pages}{108102}.
\newblock
\showISSN{0031-3203}
\urldef\tempurl%
\url{https://doi.org/10.1016/j.patcog.2021.108102}
\showDOI{\tempurl}


\bibitem[Baker and Xiang(2023)]%
        {bakerAIOT}
\bibfield{author}{\bibinfo{person}{Stephanie Baker} {and} \bibinfo{person}{Wei Xiang}.} \bibinfo{year}{2023}\natexlab{}.
\newblock \showarticletitle{Artificial Intelligence of Things for Smarter Healthcare: A Survey of Advancements, Challenges, and Opportunities}.
\newblock \bibinfo{journal}{\emph{IEEE Communications Surveys \& Tutorials}} \bibinfo{volume}{25}, \bibinfo{number}{2} (\bibinfo{year}{2023}), \bibinfo{pages}{1261--1293}.
\newblock
\urldef\tempurl%
\url{https://doi.org/10.1109/COMST.2023.3256323}
\showDOI{\tempurl}


\bibitem[Baker et~al\mbox{.}(2021)]%
        {Baker2021}
\bibfield{author}{\bibinfo{person}{Stephanie Baker}, \bibinfo{person}{Wei Xiang}, {and} \bibinfo{person}{Ian Atkinson}.} \bibinfo{year}{2021}\natexlab{}.
\newblock \showarticletitle{{Hybridized neural networks for non-invasive and continuous mortality risk assessment in neonates}}.
\newblock \bibinfo{journal}{\emph{Comp. Biol. Med.}}  \bibinfo{volume}{134} (\bibinfo{year}{2021}), \bibinfo{pages}{104521}.
\newblock
\showISSN{0010-4825}
\urldef\tempurl%
\url{https://doi.org/10.1016/j.compbiomed.2021.104521}
\showDOI{\tempurl}


\bibitem[{Barredo Arrieta} et~al\mbox{.}(2020)]%
        {BarredoArrieta2020}
\bibfield{author}{\bibinfo{person}{Alejandro {Barredo Arrieta}}, \bibinfo{person}{Natalia D{\'{i}}az-Rodr{\'{i}}guez}, \bibinfo{person}{Javier {Del Ser}}, {and} others.} \bibinfo{year}{2020}\natexlab{}.
\newblock \showarticletitle{{Explainable Artificial Intelligence (XAI): Concepts, taxonomies, opportunities and challenges toward responsible AI}}.
\newblock \bibinfo{journal}{\emph{Information Fusion}}  \bibinfo{volume}{58} (\bibinfo{year}{2020}), \bibinfo{pages}{82--115}.
\newblock
\showISSN{1566-2535}
\urldef\tempurl%
\url{https://doi.org/10.1016/j.inffus.2019.12.012}
\showDOI{\tempurl}


\bibitem[Bavas(2019)]%
        {Bavas2019}
\bibfield{author}{\bibinfo{person}{Josh Bavas}.} \bibinfo{year}{2019}\natexlab{}.
\newblock \bibinfo{title}{{Facial recognition quietly switched on at Queensland stadiums, sparking privacy concerns}}.
\newblock
\newblock
\urldef\tempurl%
\url{https://www.abc.net.au/news/2019-06-05/facial-recognition-quietly-switched-on-at-queensland-stadiums/11178334}
\showURL{%
\tempurl}


\bibitem[Behl et~al\mbox{.}(2021)]%
        {Behl2021}
\bibfield{author}{\bibinfo{person}{Shivam Behl}, \bibinfo{person}{Aman Rao}, \bibinfo{person}{Sahil Aggarwal}, {and} others.} \bibinfo{year}{2021}\natexlab{}.
\newblock \showarticletitle{{Twitter for disaster relief through sentiment analysis for COVID-19 and natural hazard crises}}.
\newblock \bibinfo{journal}{\emph{International Journal of Disaster Risk Reduction}}  \bibinfo{volume}{55} (\bibinfo{year}{2021}), \bibinfo{pages}{102101}.
\newblock
\showISSN{2212-4209}
\urldef\tempurl%
\url{https://doi.org/10.1016/j.ijdrr.2021.102101}
\showDOI{\tempurl}


\bibitem[Bennetot et~al\mbox{.}(2022)]%
        {Bennetot2022}
\bibfield{author}{\bibinfo{person}{Adrien Bennetot}, \bibinfo{person}{Gianni Franchi}, \bibinfo{person}{Javier~Del Ser}, {and} others.} \bibinfo{year}{2022}\natexlab{}.
\newblock \showarticletitle{{Greybox XAI: A Neural-Symbolic learning framework to produce interpretable predictions for image classification}}.
\newblock \bibinfo{journal}{\emph{Knowledge-Based Systems}}  \bibinfo{volume}{258} (\bibinfo{year}{2022}), \bibinfo{pages}{109947}.
\newblock
\showISSN{0950-7051}
\urldef\tempurl%
\url{https://doi.org/10.1016/j.knosys.2022.109947}
\showDOI{\tempurl}


\bibitem[Bollyky et~al\mbox{.}(2022)]%
        {Bollyky2022}
\bibfield{author}{\bibinfo{person}{T~J Bollyky}, \bibinfo{person}{E~N Hulland}, \bibinfo{person}{R~M Barber}, {and} others.} \bibinfo{year}{2022}\natexlab{}.
\newblock \showarticletitle{{Pandemic preparedness and COVID-19: an exploratory analysis of infection and fatality rates, and contextual factors associated with preparedness in 177 countries, from Jan 1, 2020, to Sept 30, 2021}}.
\newblock \bibinfo{journal}{\emph{The Lancet}}  \bibinfo{volume}{399} (\bibinfo{year}{2022}), \bibinfo{pages}{1489--1512}.
\newblock
Issue 10334.
\urldef\tempurl%
\url{https://doi.org/10.1016/S0140-6736(22)00172-6}
\showDOI{\tempurl}


\bibitem[Bowyer et~al\mbox{.}(2020)]%
        {Bowyer2020}
\bibfield{author}{\bibinfo{person}{K~W Bowyer}, \bibinfo{person}{M~C King}, \bibinfo{person}{W~J Scheirer}, {and} \bibinfo{person}{K Vangara}.} \bibinfo{year}{2020}\natexlab{}.
\newblock \showarticletitle{{The “Criminality From Face” Illusion}}.
\newblock \bibinfo{journal}{\emph{IEEE Transactions on Technology and Society}} \bibinfo{volume}{1}, \bibinfo{number}{4} (\bibinfo{year}{2020}), \bibinfo{pages}{175--183}.
\newblock
\showISSN{2637-6415 VO - 1}
\urldef\tempurl%
\url{https://doi.org/10.1109/TTS.2020.3032321}
\showDOI{\tempurl}


\bibitem[Bozorgpanah et~al\mbox{.}(2022)]%
        {Bozorgpanah2022}
\bibfield{author}{\bibinfo{person}{Aso Bozorgpanah}, \bibinfo{person}{Vicen{\c{c}} Torra}, {and} \bibinfo{person}{Laya Aliahmadipour}.} \bibinfo{year}{2022}\natexlab{}.
\newblock \bibinfo{title}{{Privacy and Explainability: The Effects of Data Protection on Shapley Values}}.
\newblock
\newblock
\showISBNx{2227-7080}
\urldef\tempurl%
\url{https://doi.org/10.3390/technologies10060125}
\showDOI{\tempurl}


\bibitem[Brito et~al\mbox{.}(2022)]%
        {Brito2022}
\bibfield{author}{\bibinfo{person}{L~C Brito}, \bibinfo{person}{G~A Susto}, \bibinfo{person}{J~N Brito}, {and} \bibinfo{person}{M~A~V Duarte}.} \bibinfo{year}{2022}\natexlab{}.
\newblock \showarticletitle{{An explainable artificial intelligence approach for unsupervised fault detection and diagnosis in rotating machinery}}.
\newblock \bibinfo{journal}{\emph{Mechanical Systems and Signal Processing}}  \bibinfo{volume}{163} (\bibinfo{year}{2022}), \bibinfo{pages}{108105}.
\newblock
\urldef\tempurl%
\url{https://doi.org/10.1016/j.ymssp.2021.108105}
\showDOI{\tempurl}


\bibitem[B{\"{u}}cker et~al\mbox{.}(2022)]%
        {Bucker2022}
\bibfield{author}{\bibinfo{person}{Michael B{\"{u}}cker}, \bibinfo{person}{Gero Szepannek}, \bibinfo{person}{Alicja Gosiewska}, {and} \bibinfo{person}{Przemyslaw Biecek}.} \bibinfo{year}{2022}\natexlab{}.
\newblock \showarticletitle{{Transparency, auditability, and explainability of machine learning models in credit scoring}}.
\newblock \bibinfo{journal}{\emph{Journal of the Operational Research Society}} \bibinfo{volume}{73}, \bibinfo{number}{1} (\bibinfo{date}{jan} \bibinfo{year}{2022}), \bibinfo{pages}{70--90}.
\newblock
\showISSN{0160-5682}
\urldef\tempurl%
\url{https://doi.org/10.1080/01605682.2021.1922098}
\showDOI{\tempurl}


\bibitem[Buolamwini and Gebru(2018)]%
        {buolamwini2018}
\bibfield{author}{\bibinfo{person}{Joy Buolamwini} {and} \bibinfo{person}{Timnit Gebru}.} \bibinfo{year}{2018}\natexlab{}.
\newblock \showarticletitle{Gender Shades: Intersectional Accuracy Disparities in Commercial Gender Classification}. In \bibinfo{booktitle}{\emph{Proceedings of the 1st Conference on Fairness, Accountability and Transparency}} \emph{(\bibinfo{series}{Proceedings of Machine Learning Research}, Vol.~\bibinfo{volume}{81})}. \bibinfo{publisher}{PMLR}, \bibinfo{pages}{77--91}.
\newblock
\urldef\tempurl%
\url{https://proceedings.mlr.press/v81/buolamwini18a.html}
\showURL{%
\tempurl}


\bibitem[Chefer et~al\mbox{.}(2021)]%
        {Chefer2021}
\bibfield{author}{\bibinfo{person}{H Chefer}, \bibinfo{person}{S Gur}, {and} \bibinfo{person}{L Wolf}.} \bibinfo{year}{2021}\natexlab{}.
\newblock \showarticletitle{{Transformer Interpretability Beyond Attention Visualization}}. In \bibinfo{booktitle}{\emph{2021 IEEE/CVF Conference on Computer Vision and Pattern Recognition (CVPR)}}. \bibinfo{publisher}{Curran Associates}, \bibinfo{address}{Montreal, Canada}, \bibinfo{pages}{782--791}.
\newblock
\showISBNx{2575-7075 VO}
\urldef\tempurl%
\url{https://doi.org/10.1109/CVPR46437.2021.00084}
\showDOI{\tempurl}


\bibitem[CHOICE(2022)]%
        {CHOICE2022}
\bibfield{author}{\bibinfo{person}{CHOICE}.} \bibinfo{year}{2022}\natexlab{}.
\newblock \bibinfo{title}{{Kmart, Bunnings and The Good Guys using facial recognition technology in stores}}.
\newblock
\newblock
\urldef\tempurl%
\url{https://www.choice.com.au/consumers-and-data/data-collection-and-use/how-your-data-is-used/articles/kmart-bunnings-and-the-good-guys-using-facial-recognition-technology-in-store}
\showURL{%
\tempurl}


\bibitem[Chun et~al\mbox{.}(2022)]%
        {Chun2022}
\bibfield{author}{\bibinfo{person}{P.-J. Chun}, \bibinfo{person}{T Yamane}, {and} \bibinfo{person}{Y Maemura}.} \bibinfo{year}{2022}\natexlab{}.
\newblock \showarticletitle{{A deep learning-based image captioning method to automatically generate comprehensive explanations of bridge damage}}.
\newblock \bibinfo{journal}{\emph{Computer-Aided Civil and Infrastructure Engineering}} \bibinfo{volume}{37}, \bibinfo{number}{11} (\bibinfo{year}{2022}), \bibinfo{pages}{1387--1401}.
\newblock
\urldef\tempurl%
\url{https://doi.org/10.1111/mice.12793}
\showDOI{\tempurl}


\bibitem[Chung et~al\mbox{.}(2022)]%
        {Chung2022cxr}
\bibfield{author}{\bibinfo{person}{Joowon Chung}, \bibinfo{person}{Doyun Kim}, \bibinfo{person}{Jongmun Choi}, {and} others.} \bibinfo{year}{2022}\natexlab{}.
\newblock \showarticletitle{{Prediction of oxygen requirement in patients with COVID-19 using a pre-trained chest radiograph xAI model: efficient development of auditable risk prediction models via a fine-tuning approach}}.
\newblock \bibinfo{journal}{\emph{Scientific Reports}} \bibinfo{volume}{12}, \bibinfo{number}{1} (\bibinfo{year}{2022}), \bibinfo{pages}{21164}.
\newblock
\showISSN{2045-2322}
\urldef\tempurl%
\url{https://doi.org/10.1038/s41598-022-24721-5}
\showDOI{\tempurl}


\bibitem[CSIRO(2023)]%
        {CSIRO2023}
\bibfield{author}{\bibinfo{person}{CSIRO}.} \bibinfo{year}{2023}\natexlab{}.
\newblock \bibinfo{title}{{Responsible artificial intelligence}}.
\newblock
\newblock
\urldef\tempurl%
\url{https://www.csiro.au/en/research/technology-space/ai/responsible-ai}
\showURL{%
\tempurl}


\bibitem[Dastin(2019)]%
        {Dastin2019}
\bibfield{author}{\bibinfo{person}{Jeffrey Dastin}.} \bibinfo{year}{2019}\natexlab{}.
\newblock \bibinfo{title}{{Amazon scraps secret AI recruiting tool that showed bias against women}}.
\newblock
\newblock
\urldef\tempurl%
\url{https://www.reuters.com/article/us-amazon-com-jobs-automation-insight/amazon-scraps-secret-ai-recruiting-tool-that-showed-bias-against-women-idUSKCN1MK08G}
\showURL{%
\tempurl}


\bibitem[Deshpande et~al\mbox{.}(2020)]%
        {Deshpande2020}
\bibfield{author}{\bibinfo{person}{Ketki~V. Deshpande}, \bibinfo{person}{Shimei Pan}, {and} \bibinfo{person}{James~R. Foulds}.} \bibinfo{year}{2020}\natexlab{}.
\newblock \showarticletitle{Mitigating Demographic Bias in AI-Based Resume Filtering}. In \bibinfo{booktitle}{\emph{Adjunct Publication of the 28th ACM Conference on User Modeling, Adaptation and Personalization}} (Genoa, Italy) \emph{(\bibinfo{series}{UMAP '20 Adjunct})}. \bibinfo{publisher}{ACM}, \bibinfo{address}{New York, NY, USA}, \bibinfo{pages}{268–275}.
\newblock
\showISBNx{9781450379502}
\urldef\tempurl%
\url{https://doi.org/10.1145/3386392.3399569}
\showDOI{\tempurl}


\bibitem[Ding et~al\mbox{.}(2022)]%
        {Ding2022}
\bibfield{author}{\bibinfo{person}{Weiping Ding}, \bibinfo{person}{Mohamed Abdel-Basset}, \bibinfo{person}{Hossam Hawash}, {and} \bibinfo{person}{Ahmed~M Ali}.} \bibinfo{year}{2022}\natexlab{}.
\newblock \showarticletitle{{Explainability of artificial intelligence methods, applications and challenges: A comprehensive survey}}.
\newblock \bibinfo{journal}{\emph{Information Sciences}}  \bibinfo{volume}{615} (\bibinfo{year}{2022}), \bibinfo{pages}{238--292}.
\newblock
\showISSN{0020-0255}
\urldef\tempurl%
\url{https://doi.org/10.1016/j.ins.2022.10.013}
\showDOI{\tempurl}


\bibitem[Draelos and Carin(2020)]%
        {Draelos2020}
\bibfield{author}{\bibinfo{person}{Rachel~Lea Draelos} {and} \bibinfo{person}{Lawrence Carin}.} \bibinfo{year}{2020}\natexlab{}.
\newblock \showarticletitle{{HiResCam: Faithful location representation in visual attention for explainable 3D medical image classification}}.
\newblock \bibinfo{journal}{\emph{arXiv preprint arXiv:2011.08891}}  \bibinfo{volume}{1} (\bibinfo{year}{2020}), \bibinfo{pages}{1}.
\newblock


\bibitem[Dwivedi et~al\mbox{.}(2023)]%
        {dwivedi2023}
\bibfield{author}{\bibinfo{person}{Rudresh Dwivedi}, \bibinfo{person}{Devam Dave}, \bibinfo{person}{Het Naik}, {and} others.} \bibinfo{year}{2023}\natexlab{}.
\newblock \showarticletitle{{Explainable AI (XAI): Core Ideas, Techniques, and Solutions}}.
\newblock \bibinfo{journal}{\emph{ACM Comput. Surv.}} \bibinfo{volume}{55}, \bibinfo{number}{9} (\bibinfo{date}{jan} \bibinfo{year}{2023}), \bibinfo{pages}{1--33}.
\newblock
\showISSN{0360-0300}
\urldef\tempurl%
\url{https://doi.org/10.1145/3561048}
\showDOI{\tempurl}


\bibitem[Ensign et~al\mbox{.}(2018)]%
        {Ensign2018}
\bibfield{author}{\bibinfo{person}{Danielle Ensign}, \bibinfo{person}{Sorelle~A Friedler}, \bibinfo{person}{Scott Neville}, {and} others.} \bibinfo{year}{2018}\natexlab{}.
\newblock \showarticletitle{{Runaway feedback loops in predictive policing}}. In \bibinfo{booktitle}{\emph{Proceedings of Machine Learning Research 2018}}. \bibinfo{publisher}{PMLR}, \bibinfo{address}{New York}, \bibinfo{pages}{160--171}.
\newblock
\showISBNx{2640-3498}


\bibitem[{European Commission}(2021)]%
        {EuropeanCommission2021}
\bibfield{author}{\bibinfo{person}{{European Commission}}.} \bibinfo{year}{2021}\natexlab{}.
\newblock \bibinfo{booktitle}{\emph{{Ethics Guidelines for Trustworthy AI}}}.
\newblock \bibinfo{type}{{T}echnical {R}eport}. \bibinfo{institution}{European Union}.
\newblock
\urldef\tempurl%
\url{https://ec.europa.eu/futurium/en/ai-alliance-consultation.1.html}
\showURL{%
\tempurl}


\bibitem[Fan et~al\mbox{.}(2022)]%
        {Fan2022}
\bibfield{author}{\bibinfo{person}{Y~V Fan}, \bibinfo{person}{P Jiang}, \bibinfo{person}{R~R Tan}, {and} others.} \bibinfo{year}{2022}\natexlab{}.
\newblock \showarticletitle{{Forecasting plastic waste generation and interventions for environmental hazard mitigation}}.
\newblock \bibinfo{journal}{\emph{Journal of Hazardous Materials}}  \bibinfo{volume}{424} (\bibinfo{year}{2022}), \bibinfo{pages}{127330}.
\newblock
\urldef\tempurl%
\url{https://doi.org/10.1016/j.jhazmat.2021.127330}
\showDOI{\tempurl}


\bibitem[Felzmann et~al\mbox{.}(2020)]%
        {felzmann2020towards}
\bibfield{author}{\bibinfo{person}{Heike Felzmann}, \bibinfo{person}{Eduard Fosch-Villaronga}, \bibinfo{person}{Christoph Lutz}, {and} \bibinfo{person}{Aurelia Tam{\`o}-Larrieux}.} \bibinfo{year}{2020}\natexlab{}.
\newblock \showarticletitle{Towards transparency by design for artificial intelligence}.
\newblock \bibinfo{journal}{\emph{Science and Engineering Ethics}} \bibinfo{volume}{26}, \bibinfo{number}{6} (\bibinfo{year}{2020}), \bibinfo{pages}{3333--3361}.
\newblock


\bibitem[Garb(2021)]%
        {Garb2021}
\bibfield{author}{\bibinfo{person}{Howard~N Garb}.} \bibinfo{year}{2021}\natexlab{}.
\newblock \showarticletitle{{Race bias and gender bias in the diagnosis of psychological disorders}}.
\newblock \bibinfo{journal}{\emph{Clinical Psychology Review}}  \bibinfo{volume}{90} (\bibinfo{year}{2021}), \bibinfo{pages}{102087}.
\newblock
\showISSN{0272-7358}
\urldef\tempurl%
\url{https://doi.org/10.1016/j.cpr.2021.102087}
\showDOI{\tempurl}


\bibitem[Gaudio et~al\mbox{.}(2023)]%
        {Gaudio2023}
\bibfield{author}{\bibinfo{person}{Alex Gaudio}, \bibinfo{person}{Asim Smailagic}, \bibinfo{person}{Christos Faloutsos}, {and} others.} \bibinfo{year}{2023}\natexlab{}.
\newblock \showarticletitle{{DeepFixCX: Explainable privacy-preserving image compression for medical image analysis}}.
\newblock \bibinfo{journal}{\emph{WIREs Data Mining and Knowledge Discovery}} \bibinfo{volume}{n/a}, \bibinfo{number}{n/a} (\bibinfo{date}{mar} \bibinfo{year}{2023}), \bibinfo{pages}{e1495}.
\newblock
\showISSN{1942-4787}
\urldef\tempurl%
\url{https://doi.org/10.1002/widm.1495}
\showDOI{\tempurl}


\bibitem[Gaur et~al\mbox{.}(2022)]%
        {Gaur2022}
\bibfield{author}{\bibinfo{person}{L Gaur}, \bibinfo{person}{M Bhandari}, \bibinfo{person}{T Razdan}, {and} others.} \bibinfo{year}{2022}\natexlab{}.
\newblock \showarticletitle{{Explanation-Driven Deep Learning Model for Prediction of Brain Tumour Status Using MRI Image Data}}.
\newblock \bibinfo{journal}{\emph{Frontiers in Genetics}}  \bibinfo{volume}{13} (\bibinfo{year}{2022}), \bibinfo{pages}{822666}.
\newblock
\urldef\tempurl%
\url{https://doi.org/10.3389/fgene.2022.822666}
\showDOI{\tempurl}


\bibitem[Ghosh et~al\mbox{.}(2021)]%
        {Ghosh2021}
\bibfield{author}{\bibinfo{person}{P Ghosh}, \bibinfo{person}{S Azam}, \bibinfo{person}{M Jonkman}, {and} others.} \bibinfo{year}{2021}\natexlab{}.
\newblock \showarticletitle{{Efficient prediction of cardiovascular disease using machine learning algorithms with relief and lasso feature selection techniques}}.
\newblock \bibinfo{journal}{\emph{IEEE Access}}  \bibinfo{volume}{9} (\bibinfo{year}{2021}), \bibinfo{pages}{19304--19326}.
\newblock
\urldef\tempurl%
\url{https://doi.org/10.1109/ACCESS.2021.3053759}
\showDOI{\tempurl}


\bibitem[Google(2023)]%
        {Google2023}
\bibfield{author}{\bibinfo{person}{Google}.} \bibinfo{year}{2023}\natexlab{}.
\newblock \bibinfo{title}{{Artificial Intelligence at Google: Our Principles}}.
\newblock
\newblock
\urldef\tempurl%
\url{https://ai.google/principles/}
\showURL{%
\tempurl}


\bibitem[Granata et~al\mbox{.}(2022)]%
        {Granata2022}
\bibfield{author}{\bibinfo{person}{V Granata}, \bibinfo{person}{R Fusco}, \bibinfo{person}{F {De Muzio}}, {and} others.} \bibinfo{year}{2022}\natexlab{}.
\newblock \showarticletitle{{EOB-MR Based Radiomics Analysis to Assess Clinical Outcomes following Liver Resection in Colorectal Liver Metastases}}.
\newblock \bibinfo{journal}{\emph{Cancers}} \bibinfo{volume}{14}, \bibinfo{number}{5} (\bibinfo{year}{2022}), \bibinfo{pages}{1239}.
\newblock
\urldef\tempurl%
\url{https://doi.org/10.3390/cancers14051239}
\showDOI{\tempurl}


\bibitem[Guidotti(2022)]%
        {Guidotti2022}
\bibfield{author}{\bibinfo{person}{R Guidotti}.} \bibinfo{year}{2022}\natexlab{}.
\newblock \showarticletitle{{Counterfactual explanations and how to find them: literature review and benchmarking}}.
\newblock \bibinfo{journal}{\emph{Data Mining and Knowledge Discovery}}  \bibinfo{volume}{36} (\bibinfo{year}{2022}), \bibinfo{pages}{1--55}.
\newblock
\urldef\tempurl%
\url{https://doi.org/10.1007/s10618-022-00831-6}
\showDOI{\tempurl}


\bibitem[Himeur et~al\mbox{.}(2021)]%
        {Himeur2021}
\bibfield{author}{\bibinfo{person}{Y Himeur}, \bibinfo{person}{K Ghanem}, \bibinfo{person}{A Alsalemi}, {and} others.} \bibinfo{year}{2021}\natexlab{}.
\newblock \showarticletitle{{Artificial intelligence based anomaly detection of energy consumption in buildings: A review, current trends and new perspectives}}.
\newblock \bibinfo{journal}{\emph{Applied Energy}}  \bibinfo{volume}{287} (\bibinfo{year}{2021}), \bibinfo{pages}{116601}.
\newblock
\urldef\tempurl%
\url{https://doi.org/10.1016/j.apenergy.2021.116601}
\showDOI{\tempurl}


\bibitem[{HM Government (United Kingdom)}(2021)]%
        {UKGovernment2021}
\bibfield{author}{\bibinfo{person}{{HM Government (United Kingdom)}}.} \bibinfo{year}{2021}\natexlab{}.
\newblock \bibinfo{booktitle}{\emph{{National AI Strategy}}}.
\newblock \bibinfo{type}{{T}echnical {R}eport}.
\newblock
\urldef\tempurl%
\url{https://assets.publishing.service.gov.uk/government/uploads/system/uploads/attachment_data/file/1020402/National_AI_Strategy_-_PDF_version.pdf}
\showURL{%
\tempurl}


\bibitem[Hosseinzadeh et~al\mbox{.}(2022)]%
        {Hosseinzadeh2022}
\bibfield{author}{\bibinfo{person}{A Hosseinzadeh}, \bibinfo{person}{J~L Zhou}, \bibinfo{person}{A Altaee}, {and} \bibinfo{person}{D Li}.} \bibinfo{year}{2022}\natexlab{}.
\newblock \showarticletitle{{Machine learning modeling and analysis of biohydrogen production from wastewater by dark fermentation process}}.
\newblock \bibinfo{journal}{\emph{Bioresource Technology}}  \bibinfo{volume}{343} (\bibinfo{year}{2022}), \bibinfo{pages}{126111}.
\newblock
\urldef\tempurl%
\url{https://doi.org/10.1016/j.biortech.2021.126111}
\showDOI{\tempurl}


\bibitem[Houda et~al\mbox{.}(2022)]%
        {Houda2022}
\bibfield{author}{\bibinfo{person}{Z~A~E Houda}, \bibinfo{person}{B Brik}, {and} \bibinfo{person}{L Khoukhi}.} \bibinfo{year}{2022}\natexlab{}.
\newblock \showarticletitle{{“Why Should I Trust Your IDS?”: An Explainable Deep Learning Framework for Intrusion Detection Systems in Internet of Things Networks}}.
\newblock \bibinfo{journal}{\emph{IEEE Open Journal of the Communications Society}}  \bibinfo{volume}{3} (\bibinfo{year}{2022}), \bibinfo{pages}{1164--1176}.
\newblock
\showISSN{2644-125X VO - 3}
\urldef\tempurl%
\url{https://doi.org/10.1109/OJCOMS.2022.3188750}
\showDOI{\tempurl}


\bibitem[Hsu and Thompson(2022)]%
        {Hsu2022}
\bibfield{author}{\bibinfo{person}{Tiffany Hsu} {and} \bibinfo{person}{Stuart~A Thompson}.} \bibinfo{year}{2022}\natexlab{}.
\newblock \bibinfo{title}{{Disinformation Researchers Raise Alarms About A.I. Chatbots}}.
\newblock
\newblock
\urldef\tempurl%
\url{https://www.nytimes.com/2023/02/08/technology/ai-chatbots-disinformation.html}
\showURL{%
\tempurl}


\bibitem[IBM(2023)]%
        {IBM2023}
\bibfield{author}{\bibinfo{person}{IBM}.} \bibinfo{year}{2023}\natexlab{}.
\newblock \bibinfo{title}{{AI Ethics}}.
\newblock
\newblock
\urldef\tempurl%
\url{https://www.ibm.com/topics/ai-ethics}
\showURL{%
\tempurl}


\bibitem[{India AI} and {National Association of Software and Service Companies (India)}(2022)]%
        {IndiaAI2022}
\bibfield{author}{\bibinfo{person}{{India AI}} {and} \bibinfo{person}{{National Association of Software and Service Companies (India)}}.} \bibinfo{year}{2022}\natexlab{}.
\newblock \bibinfo{title}{{Responsible AI Principles}}.
\newblock
\newblock
\urldef\tempurl%
\url{https://indiaai.gov.in/responsible-ai/homepage}
\showURL{%
\tempurl}


\bibitem[Jang et~al\mbox{.}(2022)]%
        {Jang2022}
\bibfield{author}{\bibinfo{person}{Yeonju Jang}, \bibinfo{person}{Seongyune Choi}, \bibinfo{person}{Heeseok Jung}, {and} \bibinfo{person}{Hyeoncheol Kim}.} \bibinfo{year}{2022}\natexlab{}.
\newblock \showarticletitle{{Practical early prediction of students' performance using machine learning and eXplainable AI}}.
\newblock \bibinfo{journal}{\emph{Education and Information Technologies}} \bibinfo{volume}{27}, \bibinfo{number}{9} (\bibinfo{year}{2022}), \bibinfo{pages}{12855--12889}.
\newblock
\showISSN{1573-7608}
\urldef\tempurl%
\url{https://doi.org/10.1007/s10639-022-11120-6}
\showDOI{\tempurl}


\bibitem[Jia et~al\mbox{.}(2022b)]%
        {Jia2022}
\bibfield{author}{\bibinfo{person}{X Jia}, \bibinfo{person}{K Shahzad}, \bibinfo{person}{J~J Kleme{\v{s}}}, {and} \bibinfo{person}{X Jia}.} \bibinfo{year}{2022}\natexlab{b}.
\newblock \showarticletitle{{Changes in water use and wastewater generation influenced by the COVID-19 pandemic: A case study of China}}.
\newblock \bibinfo{journal}{\emph{Journal of Environmental Management}}  \bibinfo{volume}{314} (\bibinfo{year}{2022}), \bibinfo{pages}{115024}.
\newblock
\urldef\tempurl%
\url{https://doi.org/10.1016/j.jenvman.2022.115024}
\showDOI{\tempurl}


\bibitem[Jia et~al\mbox{.}(2022a)]%
        {Jia2022xaihealthcare}
\bibfield{author}{\bibinfo{person}{Y Jia}, \bibinfo{person}{J McDermid}, \bibinfo{person}{T Lawton}, {and} \bibinfo{person}{I Habli}.} \bibinfo{year}{2022}\natexlab{a}.
\newblock \showarticletitle{{The Role of Explainability in Assuring Safety of Machine Learning in Healthcare}}.
\newblock \bibinfo{journal}{\emph{IEEE Transactions on Emerging Topics in Computing}} \bibinfo{volume}{10}, \bibinfo{number}{4} (\bibinfo{year}{2022}), \bibinfo{pages}{1746--1760}.
\newblock
\showISSN{2168-6750 VO - 10}
\urldef\tempurl%
\url{https://doi.org/10.1109/TETC.2022.3171314}
\showDOI{\tempurl}


\bibitem[Jiang et~al\mbox{.}(2023)]%
        {Jiang2023}
\bibfield{author}{\bibinfo{person}{Ke-Wen Jiang}, \bibinfo{person}{Yang Song}, \bibinfo{person}{Ying Hou}, {and} others.} \bibinfo{year}{2023}\natexlab{}.
\newblock \showarticletitle{{Performance of Artificial Intelligence-Aided Diagnosis System for Clinically Significant Prostate Cancer with MRI: A Diagnostic Comparison Study}}.
\newblock \bibinfo{journal}{\emph{Journal of Magnetic Resonance Imaging}} \bibinfo{volume}{57}, \bibinfo{number}{5} (\bibinfo{date}{may} \bibinfo{year}{2023}), \bibinfo{pages}{1352--1364}.
\newblock
\showISSN{1053-1807}
\urldef\tempurl%
\url{https://doi.org/10.1002/jmri.28427}
\showDOI{\tempurl}


\bibitem[Kaur et~al\mbox{.}(2022)]%
        {kaur2022}
\bibfield{author}{\bibinfo{person}{Davinder Kaur}, \bibinfo{person}{Suleyman Uslu}, \bibinfo{person}{Kaley~J Rittichier}, {and} \bibinfo{person}{Arjan Durresi}.} \bibinfo{year}{2022}\natexlab{}.
\newblock \showarticletitle{{Trustworthy Artificial Intelligence: A Review}}.
\newblock \bibinfo{journal}{\emph{ACM Comput. Surv.}} \bibinfo{volume}{55}, \bibinfo{number}{2} (\bibinfo{date}{jan} \bibinfo{year}{2022}), \bibinfo{pages}{1--38}.
\newblock
\showISSN{0360-0300}
\urldef\tempurl%
\url{https://doi.org/10.1145/3491209}
\showDOI{\tempurl}


\bibitem[Keras(2023)]%
        {Keras2023}
\bibfield{author}{\bibinfo{person}{Keras}.} \bibinfo{year}{2023}\natexlab{}.
\newblock \bibinfo{title}{{VGG16 and VGG19}}.
\newblock
\newblock
\urldef\tempurl%
\url{https://keras.io/api/applications/vgg/}
\showURL{%
\tempurl}


\bibitem[Khan et~al\mbox{.}(2022)]%
        {Khan2022}
\bibfield{author}{\bibinfo{person}{M Khan}, \bibinfo{person}{Z Ullah}, \bibinfo{person}{O Ma{\v{s}}ek}, {and} others.} \bibinfo{year}{2022}\natexlab{}.
\newblock \showarticletitle{{Artificial neural networks for the prediction of biochar yield: A comparative study of metaheuristic algorithms}}.
\newblock \bibinfo{journal}{\emph{Bioresource Technology}}  \bibinfo{volume}{355} (\bibinfo{year}{2022}), \bibinfo{pages}{127215}.
\newblock
\urldef\tempurl%
\url{https://doi.org/10.1016/j.biortech.2022.127215}
\showDOI{\tempurl}


\bibitem[Khodabandehloo et~al\mbox{.}(2021)]%
        {Khodabandehloo2021}
\bibfield{author}{\bibinfo{person}{Elham Khodabandehloo}, \bibinfo{person}{Daniele Riboni}, {and} \bibinfo{person}{Abbas Alimohammadi}.} \bibinfo{year}{2021}\natexlab{}.
\newblock \showarticletitle{{HealthXAI: Collaborative and explainable AI for supporting early diagnosis of cognitive decline}}.
\newblock \bibinfo{journal}{\emph{Future Generation Computer Systems}}  \bibinfo{volume}{116} (\bibinfo{year}{2021}), \bibinfo{pages}{168--189}.
\newblock
\showISSN{0167-739X}
\urldef\tempurl%
\url{https://doi.org/10.1016/j.future.2020.10.030}
\showDOI{\tempurl}


\bibitem[Khurana et~al\mbox{.}(2021)]%
        {Khurana2021}
\bibfield{author}{\bibinfo{person}{A Khurana}, \bibinfo{person}{P Alamzadeh}, {and} \bibinfo{person}{P~K Chilana}.} \bibinfo{year}{2021}\natexlab{}.
\newblock \showarticletitle{{ChatrEx: Designing Explainable Chatbot Interfaces for Enhancing Usefulness, Transparency, and Trust}}. In \bibinfo{booktitle}{\emph{2021 IEEE Symposium on Visual Languages and Human-Centric Computing (VL/HCC)}}. \bibinfo{pages}{1--11}.
\newblock
\showISBNx{1943-6106 VO -}
\urldef\tempurl%
\url{https://doi.org/10.1109/VL/HCC51201.2021.9576440}
\showDOI{\tempurl}


\bibitem[Kim et~al\mbox{.}(2021)]%
        {Kim2021}
\bibfield{author}{\bibinfo{person}{K Kim}, \bibinfo{person}{H Yang}, \bibinfo{person}{J Yi}, {and} others.} \bibinfo{year}{2021}\natexlab{}.
\newblock \showarticletitle{{Real-time clinical decision support based on recurrent neural networks for in-hospital acute kidney injury: External validation and model interpretation}}.
\newblock \bibinfo{journal}{\emph{Journal of Medical Internet Research}} \bibinfo{volume}{23}, \bibinfo{number}{4} (\bibinfo{year}{2021}), \bibinfo{pages}{e24120}.
\newblock
\urldef\tempurl%
\url{https://doi.org/10.2196/24120}
\showDOI{\tempurl}


\bibitem[{Kingdom of Saudi Arabia}(2022)]%
        {SaudiArabia2022}
\bibfield{author}{\bibinfo{person}{{Kingdom of Saudi Arabia}}.} \bibinfo{year}{2022}\natexlab{}.
\newblock \bibinfo{booktitle}{\emph{{AI Ethics Principles}}}.
\newblock \bibinfo{type}{{T}echnical {R}eport}.
\newblock
\urldef\tempurl%
\url{https://istitlaa.ncc.gov.sa/en/transportation/ndmo/aiethicsprinciples/Documents/AI Ethics Principles.pdf}
\showURL{%
\tempurl}


\bibitem[Kostopoulos et~al\mbox{.}(2021)]%
        {Kostopoulos2021}
\bibfield{author}{\bibinfo{person}{G Kostopoulos}, \bibinfo{person}{T Panagiotakopoulos}, \bibinfo{person}{S Kotsiantis}, {and} others.} \bibinfo{year}{2021}\natexlab{}.
\newblock \showarticletitle{{Interpretable Models for Early Prediction of Certification in MOOCs: A Case Study on a MOOC for Smart City Professionals}}.
\newblock \bibinfo{journal}{\emph{IEEE Access}}  \bibinfo{volume}{9} (\bibinfo{year}{2021}), \bibinfo{pages}{165881--165891}.
\newblock
\showISSN{2169-3536 VO - 9}
\urldef\tempurl%
\url{https://doi.org/10.1109/ACCESS.2021.3134787}
\showDOI{\tempurl}


\bibitem[Krizhevsky et~al\mbox{.}(2012)]%
        {AlexNet}
\bibfield{author}{\bibinfo{person}{Alex Krizhevsky}, \bibinfo{person}{Ilya Sutskever}, {and} \bibinfo{person}{Geoffrey~E. Hinton}.} \bibinfo{year}{2012}\natexlab{}.
\newblock \showarticletitle{ImageNet Classification with Deep Convolutional Neural Networks}. In \bibinfo{booktitle}{\emph{Proceedings of the 25th International Conference on Neural Information Processing Systems - Volume 1}} (Lake Tahoe, Nevada) \emph{(\bibinfo{series}{NIPS'12})}. \bibinfo{publisher}{Curran Associates Inc.}, \bibinfo{address}{Red Hook, NY, USA}, \bibinfo{pages}{1097–1105}.
\newblock


\bibitem[{La Gatta} et~al\mbox{.}(2021)]%
        {LaGatta2021}
\bibfield{author}{\bibinfo{person}{Valerio {La Gatta}}, \bibinfo{person}{Vincenzo Moscato}, \bibinfo{person}{Marco Postiglione}, {and} \bibinfo{person}{Giancarlo Sperl{\`{i}}}.} \bibinfo{year}{2021}\natexlab{}.
\newblock \showarticletitle{{CASTLE: Cluster-aided space transformation for local explanations}}.
\newblock \bibinfo{journal}{\emph{Expert Systems with Applications}}  \bibinfo{volume}{179} (\bibinfo{year}{2021}), \bibinfo{pages}{115045}.
\newblock
\showISSN{0957-4174}
\urldef\tempurl%
\url{https://doi.org/10.1016/j.eswa.2021.115045}
\showDOI{\tempurl}


\bibitem[Laestadius et~al\mbox{.}(2022)]%
        {Laestadius2022}
\bibfield{author}{\bibinfo{person}{Linnea Laestadius}, \bibinfo{person}{Andrea Bishop}, \bibinfo{person}{Michael Gonzalez}, {and} others.} \bibinfo{year}{2022}\natexlab{}.
\newblock \showarticletitle{{Too human and not human enough: A grounded theory analysis of mental health harms from emotional dependence on the social chatbot Replika}}.
\newblock \bibinfo{journal}{\emph{New Media \& Society}}  \bibinfo{volume}{26} (\bibinfo{date}{dec} \bibinfo{year}{2022}), \bibinfo{pages}{14614448221142007}.
\newblock
\showISSN{1461-4448}
\urldef\tempurl%
\url{https://doi.org/10.1177/14614448221142007}
\showDOI{\tempurl}


\bibitem[Li et~al\mbox{.}(2023)]%
        {Li2023}
\bibfield{author}{\bibinfo{person}{Bo Li}, \bibinfo{person}{Peng Qi}, \bibinfo{person}{Bo Liu}, {and} others.} \bibinfo{year}{2023}\natexlab{}.
\newblock \showarticletitle{{Trustworthy ai: From principles to practices}}.
\newblock \bibinfo{journal}{\emph{Comput. Surveys}} \bibinfo{volume}{55}, \bibinfo{number}{9} (\bibinfo{year}{2023}), \bibinfo{pages}{1--46}.
\newblock
\showISSN{0360-0300}


\bibitem[Liang et~al\mbox{.}(2022)]%
        {Liang2022}
\bibfield{author}{\bibinfo{person}{M Liang}, \bibinfo{person}{Z Chang}, \bibinfo{person}{Z Wan}, {and} others.} \bibinfo{year}{2022}\natexlab{}.
\newblock \showarticletitle{{Interpretable Ensemble-Machine-Learning models for predicting creep behavior of concrete}}.
\newblock \bibinfo{journal}{\emph{Cement and Concrete Composites}}  \bibinfo{volume}{125} (\bibinfo{year}{2022}), \bibinfo{pages}{104295}.
\newblock
\urldef\tempurl%
\url{https://doi.org/10.1016/j.cemconcomp.2021.104295}
\showDOI{\tempurl}


\bibitem[Loh et~al\mbox{.}(2022)]%
        {Loh2022}
\bibfield{author}{\bibinfo{person}{Hui~Wen Loh}, \bibinfo{person}{Chui~Ping Ooi}, \bibinfo{person}{Silvia Seoni}, {and} others.} \bibinfo{year}{2022}\natexlab{}.
\newblock \showarticletitle{{Application of explainable artificial intelligence for healthcare: A systematic review of the last decade (2011–2022)}}.
\newblock \bibinfo{journal}{\emph{Computer Methods and Programs in Biomedicine}}  \bibinfo{volume}{226} (\bibinfo{year}{2022}), \bibinfo{pages}{107161}.
\newblock
\showISSN{0169-2607}
\urldef\tempurl%
\url{https://doi.org/10.1016/j.cmpb.2022.107161}
\showDOI{\tempurl}


\bibitem[Lundberg and Lee(2017)]%
        {shap}
\bibfield{author}{\bibinfo{person}{Scott~M Lundberg} {and} \bibinfo{person}{Su-In Lee}.} \bibinfo{year}{2017}\natexlab{}.
\newblock \showarticletitle{A Unified Approach to Interpreting Model Predictions}.
\newblock In \bibinfo{booktitle}{\emph{Advances in Neural Information Processing Systems 30}}. \bibinfo{publisher}{Curran Associates, Inc.}, \bibinfo{address}{Long Beach, CA, USA}, \bibinfo{pages}{4765--4774}.
\newblock
\urldef\tempurl%
\url{http://papers.nips.cc/paper/7062-a-unified-approach-to-interpreting-model-predictions.pdf}
\showURL{%
\tempurl}


\bibitem[Malek(2022)]%
        {Malek2022}
\bibfield{author}{\bibinfo{person}{Md.~Abdul Malek}.} \bibinfo{year}{2022}\natexlab{}.
\newblock \showarticletitle{{Criminal courts' artificial intelligence: the way it reinforces bias and discrimination}}.
\newblock \bibinfo{journal}{\emph{AI and Ethics}} \bibinfo{volume}{2}, \bibinfo{number}{1} (\bibinfo{year}{2022}), \bibinfo{pages}{233--245}.
\newblock
\showISSN{2730-5961}
\urldef\tempurl%
\url{https://doi.org/10.1007/s43681-022-00137-9}
\showDOI{\tempurl}


\bibitem[Malik et~al\mbox{.}(2022)]%
        {malik2022xai}
\bibfield{author}{\bibinfo{person}{AL-Essa Malik}, \bibinfo{person}{Giuseppina Andresini}, \bibinfo{person}{Annalisa Appice}, {and} \bibinfo{person}{Donato Malerba}.} \bibinfo{year}{2022}\natexlab{}.
\newblock \showarticletitle{An XAI-based adversarial training approach for cyber-threat detection}. In \bibinfo{booktitle}{\emph{2022 IEEE Intl Conf on Dependable, Autonomic and Secure Computing, Intl Conf on Pervasive Intelligence and Computing, Intl Conf on Cloud and Big Data Computing, Intl Conf on Cyber Science and Technology Congress (DASC/PiCom/CBDCom/CyberSciTech)}}. IEEE, \bibinfo{pages}{1--8}.
\newblock


\bibitem[Maloney et~al\mbox{.}(2022)]%
        {Maloney2022}
\bibfield{author}{\bibinfo{person}{K~O Maloney}, \bibinfo{person}{C Buchanan}, \bibinfo{person}{R~D Jepsen}, {and} others.} \bibinfo{year}{2022}\natexlab{}.
\newblock \showarticletitle{{Explainable machine learning improves interpretability in the predictive modeling of biological stream conditions in the Chesapeake Bay Watershed, USA}}.
\newblock \bibinfo{journal}{\emph{Journal of Environmental Management}}  \bibinfo{volume}{322} (\bibinfo{year}{2022}), \bibinfo{pages}{116068}.
\newblock
\urldef\tempurl%
\url{https://doi.org/10.1016/j.jenvman.2022.116068}
\showDOI{\tempurl}


\bibitem[Marcillo et~al\mbox{.}(2021)]%
        {Marcillo2021}
\bibfield{author}{\bibinfo{person}{Guillermo~S Marcillo}, \bibinfo{person}{Nicolas~F Martin}, \bibinfo{person}{Brian~W Diers}, {and} others.} \bibinfo{year}{2021}\natexlab{}.
\newblock \bibinfo{title}{{Implementation of a Generalized Additive Model (GAM) for Soybean Maturity Prediction in African Environments}}.
\newblock
\newblock
\showISBNx{2073-4395}
\urldef\tempurl%
\url{https://doi.org/10.3390/agronomy11061043}
\showDOI{\tempurl}


\bibitem[Martinengo et~al\mbox{.}(2022)]%
        {Martinengo2022}
\bibfield{author}{\bibinfo{person}{Laura Martinengo}, \bibinfo{person}{Elaine Lum}, {and} \bibinfo{person}{Josip Car}.} \bibinfo{year}{2022}\natexlab{}.
\newblock \showarticletitle{{Evaluation of chatbot-delivered interventions for self-management of depression: Content analysis}}.
\newblock \bibinfo{journal}{\emph{Journal of Affective Disorders}}  \bibinfo{volume}{319} (\bibinfo{year}{2022}), \bibinfo{pages}{598--607}.
\newblock
\showISSN{0165-0327}
\urldef\tempurl%
\url{https://doi.org/10.1016/j.jad.2022.09.028}
\showDOI{\tempurl}


\bibitem[Mehrabi et~al\mbox{.}(2021)]%
        {Mehrabi2021}
\bibfield{author}{\bibinfo{person}{Ninareh Mehrabi}, \bibinfo{person}{Fred Morstatter}, \bibinfo{person}{Nripsuta Saxena}, {and} others.} \bibinfo{year}{2021}\natexlab{}.
\newblock \showarticletitle{A Survey on Bias and Fairness in Machine Learning}.
\newblock \bibinfo{journal}{\emph{ACM Comput. Surv.}} \bibinfo{volume}{54}, \bibinfo{number}{6}, Article \bibinfo{articleno}{115} (\bibinfo{date}{jul} \bibinfo{year}{2021}), \bibinfo{numpages}{35}~pages.
\newblock
\showISSN{0360-0300}
\urldef\tempurl%
\url{https://doi.org/10.1145/3457607}
\showDOI{\tempurl}


\bibitem[{Meta AI}(2021)]%
        {MetaAI2021}
\bibfield{author}{\bibinfo{person}{{Meta AI}}.} \bibinfo{year}{2021}\natexlab{}.
\newblock \bibinfo{title}{{Facebook's five pillars of Responsible AI}}.
\newblock
\newblock
\urldef\tempurl%
\url{https://ai.facebook.com/blog/facebooks-five-pillars-of-responsible-ai/}
\showURL{%
\tempurl}


\bibitem[Mi et~al\mbox{.}(2021)]%
        {Mi2021}
\bibfield{author}{\bibinfo{person}{X Mi}, \bibinfo{person}{B Zou}, \bibinfo{person}{F Zou}, {and} \bibinfo{person}{J Hu}.} \bibinfo{year}{2021}\natexlab{}.
\newblock \showarticletitle{{Permutation-based identification of important biomarkers for complex diseases via machine learning models}}.
\newblock \bibinfo{journal}{\emph{Nature Communications}} \bibinfo{volume}{12}, \bibinfo{number}{1} (\bibinfo{year}{2021}), \bibinfo{pages}{3008}.
\newblock
\urldef\tempurl%
\url{https://doi.org/10.1038/s41467-021-22756-2}
\showDOI{\tempurl}


\bibitem[Microsoft(2023)]%
        {Microsoft2023}
\bibfield{author}{\bibinfo{person}{Microsoft}.} \bibinfo{year}{2023}\natexlab{}.
\newblock \bibinfo{title}{{Responsible AI}}.
\newblock
\newblock
\urldef\tempurl%
\url{https://www.microsoft.com/en-us/ai/responsible-ai?activetab=pivot1%3Aprimaryr6}
\showURL{%
\tempurl}


\bibitem[Minh et~al\mbox{.}(2022)]%
        {Minh2022}
\bibfield{author}{\bibinfo{person}{Dang Minh}, \bibinfo{person}{H~Xiang Wang}, \bibinfo{person}{Y~Fen Li}, {and} \bibinfo{person}{Tan~N Nguyen}.} \bibinfo{year}{2022}\natexlab{}.
\newblock \showarticletitle{{Explainable artificial intelligence: a comprehensive review}}.
\newblock \bibinfo{journal}{\emph{Artificial Intelligence Review}} \bibinfo{volume}{55}, \bibinfo{number}{5} (\bibinfo{year}{2022}), \bibinfo{pages}{3503--3568}.
\newblock
\showISSN{1573-7462}
\urldef\tempurl%
\url{https://doi.org/10.1007/s10462-021-10088-y}
\showDOI{\tempurl}


\bibitem[{Ministry of Economy Trade and Industry (Japan)}(2022)]%
        {JapanMIST2022}
\bibfield{author}{\bibinfo{person}{{Ministry of Economy Trade and Industry (Japan)}}.} \bibinfo{year}{2022}\natexlab{}.
\newblock \bibinfo{booktitle}{\emph{{AI Governance in Japan}}}.
\newblock \bibinfo{type}{{T}echnical {R}eport}.
\newblock
\urldef\tempurl%
\url{https://www.meti.go.jp/english/press/2022/0128_003.html}
\showURL{%
\tempurl}


\bibitem[{Ministry of Science and Technology of the People's Republic of China}(2022)]%
        {ChinaRAI2022}
\bibfield{author}{\bibinfo{person}{{Ministry of Science and Technology of the People's Republic of China}}.} \bibinfo{year}{2022}\natexlab{}.
\newblock \bibinfo{booktitle}{\emph{{New Generation Artificial Intelligence Code of Ethics}}}.
\newblock \bibinfo{type}{{T}echnical {R}eport}.
\newblock
\urldef\tempurl%
\url{https://www.most.gov.cn/kjbgz/202109/t20210926_177063.html}
\showURL{%
\tempurl}


\bibitem[Miron et~al\mbox{.}(2021)]%
        {Miron2021}
\bibfield{author}{\bibinfo{person}{M Miron}, \bibinfo{person}{S Tolan}, \bibinfo{person}{E G{\'{o}}mez}, {and} \bibinfo{person}{C Castillo}.} \bibinfo{year}{2021}\natexlab{}.
\newblock \showarticletitle{{Evaluating causes of algorithmic bias in juvenile criminal recidivism}}.
\newblock \bibinfo{journal}{\emph{Artificial Intelligence and Law}} \bibinfo{volume}{29}, \bibinfo{number}{2} (\bibinfo{year}{2021}), \bibinfo{pages}{111--147}.
\newblock
\urldef\tempurl%
\url{https://doi.org/10.1007/s10506-020-09268-y}
\showDOI{\tempurl}


\bibitem[Mondal et~al\mbox{.}(2022)]%
        {Mondal2022}
\bibfield{author}{\bibinfo{person}{A~K Mondal}, \bibinfo{person}{A Bhattacharjee}, \bibinfo{person}{P Singla}, {and} \bibinfo{person}{A~P Prathosh}.} \bibinfo{year}{2022}\natexlab{}.
\newblock \showarticletitle{{xViTCOS: Explainable Vision Transformer Based COVID-19 Screening Using Radiography}}.
\newblock \bibinfo{journal}{\emph{IEEE Journal of Translational Engineering in Health and Medicine}}  \bibinfo{volume}{10} (\bibinfo{year}{2022}), \bibinfo{pages}{1--10}.
\newblock
\showISSN{2168-2372 VO - 10}
\urldef\tempurl%
\url{https://doi.org/10.1109/JTEHM.2021.3134096}
\showDOI{\tempurl}


\bibitem[Muddamsetty et~al\mbox{.}(2022)]%
        {Muddamsetty2022}
\bibfield{author}{\bibinfo{person}{S~M Muddamsetty}, \bibinfo{person}{M~N~S Jahromi}, \bibinfo{person}{A~E Ciontos}, {and} others.} \bibinfo{year}{2022}\natexlab{}.
\newblock \showarticletitle{{Visual explanation of black-box model: Similarity Difference and Uniqueness (SIDU) method}}.
\newblock \bibinfo{journal}{\emph{Pattern Recognition}}  \bibinfo{volume}{127} (\bibinfo{year}{2022}).
\newblock
\urldef\tempurl%
\url{https://doi.org/10.1016/j.patcog.2022.108604}
\showDOI{\tempurl}


\bibitem[Muhammad et~al\mbox{.}(2021)]%
        {Muhammad2021}
\bibfield{author}{\bibinfo{person}{K Muhammad}, \bibinfo{person}{A Ullah}, \bibinfo{person}{J Lloret}, {and} others.} \bibinfo{year}{2021}\natexlab{}.
\newblock \showarticletitle{{Deep Learning for Safe Autonomous Driving: Current Challenges and Future Directions}}.
\newblock \bibinfo{journal}{\emph{IEEE Transactions on Intelligent Transportation Systems}} \bibinfo{volume}{22}, \bibinfo{number}{7} (\bibinfo{year}{2021}), \bibinfo{pages}{4316--4336}.
\newblock
\urldef\tempurl%
\url{https://doi.org/10.1109/TITS.2020.3032227}
\showDOI{\tempurl}


\bibitem[Nakao et~al\mbox{.}(2022)]%
        {Nakao2022}
\bibfield{author}{\bibinfo{person}{Yuri Nakao}, \bibinfo{person}{Simone Stumpf}, \bibinfo{person}{Subeida Ahmed}, {and} others.} \bibinfo{year}{2022}\natexlab{}.
\newblock \showarticletitle{Toward Involving End-Users in Interactive Human-in-the-Loop AI Fairness}.
\newblock \bibinfo{journal}{\emph{ACM Trans. Interact. Intell. Syst.}} \bibinfo{volume}{12}, \bibinfo{number}{3}, Article \bibinfo{articleno}{18} (\bibinfo{date}{jul} \bibinfo{year}{2022}), \bibinfo{numpages}{30}~pages.
\newblock
\showISSN{2160-6455}
\urldef\tempurl%
\url{https://doi.org/10.1145/3514258}
\showDOI{\tempurl}


\bibitem[{National Institute of Standards and Technology} and {(U.S. Department of Commerce)}(2022)]%
        {USGovRAI2022}
\bibfield{author}{\bibinfo{person}{{National Institute of Standards and Technology}} {and} \bibinfo{person}{{(U.S. Department of Commerce)}}.} \bibinfo{year}{2022}\natexlab{}.
\newblock \bibinfo{booktitle}{\emph{{AI Risk Management Framework: Second Draf}}}.
\newblock \bibinfo{type}{{T}echnical {R}eport}.
\newblock
\urldef\tempurl%
\url{https://www.nist.gov/system/files/documents/2022/08/18/AI_RMF_2nd_draft.pdf}
\showURL{%
\tempurl}


\bibitem[Neves et~al\mbox{.}(2021)]%
        {Neves2021}
\bibfield{author}{\bibinfo{person}{In{\^{e}}s Neves}, \bibinfo{person}{Duarte Folgado}, \bibinfo{person}{Sara Santos}, {and} others.} \bibinfo{year}{2021}\natexlab{}.
\newblock \showarticletitle{{Interpretable heartbeat classification using local model-agnostic explanations on ECGs}}.
\newblock \bibinfo{journal}{\emph{Computers in Biology and Medicine}}  \bibinfo{volume}{133} (\bibinfo{year}{2021}), \bibinfo{pages}{104393}.
\newblock
\showISSN{0010-4825}
\urldef\tempurl%
\url{https://doi.org/10.1016/j.compbiomed.2021.104393}
\showDOI{\tempurl}


\bibitem[{New South Wales Police Force}(2023)]%
        {NewSouthWalesPoliceForce2023}
\bibfield{author}{\bibinfo{person}{{New South Wales Police Force}}.} \bibinfo{year}{2023}\natexlab{}.
\newblock \bibinfo{title}{{Facial Recognition}}.
\newblock
\newblock
\urldef\tempurl%
\url{https://www.police.nsw.gov.au/crime/terrorism/terrorism_categories/facial_recognition}
\showURL{%
\tempurl}


\bibitem[Noack et~al\mbox{.}(2021)]%
        {Noack2021}
\bibfield{author}{\bibinfo{person}{Adam Noack}, \bibinfo{person}{Isaac Ahern}, \bibinfo{person}{Dejing Dou}, {and} \bibinfo{person}{Boyang Li}.} \bibinfo{year}{2021}\natexlab{}.
\newblock \showarticletitle{{An Empirical Study on the Relation Between Network Interpretability and Adversarial Robustness}}.
\newblock \bibinfo{journal}{\emph{SN Computer Science}} \bibinfo{volume}{2}, \bibinfo{number}{1} (\bibinfo{year}{2021}), \bibinfo{pages}{32}.
\newblock
\showISSN{2661-8907}
\urldef\tempurl%
\url{https://doi.org/10.1007/s42979-020-00390-x}
\showDOI{\tempurl}


\bibitem[Norori et~al\mbox{.}(2021)]%
        {Norori2021}
\bibfield{author}{\bibinfo{person}{Natalia Norori}, \bibinfo{person}{Qiyang Hu}, \bibinfo{person}{Florence~Marcelle Aellen}, {and} others.} \bibinfo{year}{2021}\natexlab{}.
\newblock \showarticletitle{{Addressing bias in big data and AI for health care: A call for open science}}.
\newblock \bibinfo{journal}{\emph{Patterns}} \bibinfo{volume}{2}, \bibinfo{number}{10} (\bibinfo{year}{2021}), \bibinfo{pages}{100347}.
\newblock
\showISSN{2666-3899}
\urldef\tempurl%
\url{https://doi.org/10.1016/j.patter.2021.100347}
\showDOI{\tempurl}


\bibitem[Obeid et~al\mbox{.}(2020)]%
        {Obeid2020}
\bibfield{author}{\bibinfo{person}{Jihad~S Obeid}, \bibinfo{person}{Matthew Davis}, \bibinfo{person}{Matthew Turner}, {and} others.} \bibinfo{year}{2020}\natexlab{}.
\newblock \showarticletitle{{An artificial intelligence approach to COVID-19 infection risk assessment in virtual visits: A case report}}.
\newblock \bibinfo{journal}{\emph{Journal of the American Medical Informatics Association}} \bibinfo{volume}{27}, \bibinfo{number}{8} (\bibinfo{date}{aug} \bibinfo{year}{2020}), \bibinfo{pages}{1321--1325}.
\newblock
\showISSN{1527-974X}
\urldef\tempurl%
\url{https://doi.org/10.1093/jamia/ocaa105}
\showDOI{\tempurl}


\bibitem[Onchis and Gillich(2021)]%
        {Onchis2021}
\bibfield{author}{\bibinfo{person}{Darian~M Onchis} {and} \bibinfo{person}{Gilbert-Rainer Gillich}.} \bibinfo{year}{2021}\natexlab{}.
\newblock \showarticletitle{{Stable and explainable deep learning damage prediction for prismatic cantilever steel beam}}.
\newblock \bibinfo{journal}{\emph{Computers in Industry}}  \bibinfo{volume}{125} (\bibinfo{year}{2021}), \bibinfo{pages}{103359}.
\newblock
\showISSN{0166-3615}
\urldef\tempurl%
\url{https://doi.org/10.1016/j.compind.2020.103359}
\showDOI{\tempurl}


\bibitem[Onishi and Ise(2021)]%
        {Onishi2021}
\bibfield{author}{\bibinfo{person}{M Onishi} {and} \bibinfo{person}{T Ise}.} \bibinfo{year}{2021}\natexlab{}.
\newblock \showarticletitle{{Explainable identification and mapping of trees using UAV RGB image and deep learning}}.
\newblock \bibinfo{journal}{\emph{Scientific Reports}} \bibinfo{volume}{11}, \bibinfo{number}{1} (\bibinfo{year}{2021}), \bibinfo{pages}{903}.
\newblock
\urldef\tempurl%
\url{https://doi.org/10.1038/s41598-020-79653-9}
\showDOI{\tempurl}


\bibitem[Pachauri and Ahn(2022)]%
        {Pachauri2022}
\bibfield{author}{\bibinfo{person}{N Pachauri} {and} \bibinfo{person}{C~W Ahn}.} \bibinfo{year}{2022}\natexlab{}.
\newblock \showarticletitle{{Electrical Energy Prediction of Combined Cycle Power Plant Using Gradient Boosted Generalized Additive Model}}.
\newblock \bibinfo{journal}{\emph{IEEE Access}}  \bibinfo{volume}{10} (\bibinfo{year}{2022}), \bibinfo{pages}{24566--24577}.
\newblock
\showISSN{2169-3536 VO - 10}
\urldef\tempurl%
\url{https://doi.org/10.1109/ACCESS.2022.3153720}
\showDOI{\tempurl}


\bibitem[Padovan et~al\mbox{.}(2023)]%
        {Padovan2023}
\bibfield{author}{\bibinfo{person}{Paulo~Henrique Padovan}, \bibinfo{person}{Clarice~Marinho Martins}, {and} \bibinfo{person}{Chris Reed}.} \bibinfo{year}{2023}\natexlab{}.
\newblock \showarticletitle{{Black is the new orange: how to determine AI liability}}.
\newblock \bibinfo{journal}{\emph{Artificial Intelligence and Law}} \bibinfo{volume}{31}, \bibinfo{number}{1} (\bibinfo{year}{2023}), \bibinfo{pages}{133--167}.
\newblock
\showISSN{1572-8382}
\urldef\tempurl%
\url{https://doi.org/10.1007/s10506-022-09308-9}
\showDOI{\tempurl}


\bibitem[Panigutti et~al\mbox{.}(2021)]%
        {Panigutti2021}
\bibfield{author}{\bibinfo{person}{Cecilia Panigutti}, \bibinfo{person}{Alan Perotti}, \bibinfo{person}{Andr{\'{e}} Panisson}, {and} others.} \bibinfo{year}{2021}\natexlab{}.
\newblock \showarticletitle{{FairLens: Auditing black-box clinical decision support systems}}.
\newblock \bibinfo{journal}{\emph{Information Processing \& Management}} \bibinfo{volume}{58}, \bibinfo{number}{5} (\bibinfo{year}{2021}), \bibinfo{pages}{102657}.
\newblock
\showISSN{0306-4573}
\urldef\tempurl%
\url{https://doi.org/10.1016/j.ipm.2021.102657}
\showDOI{\tempurl}


\bibitem[Park et~al\mbox{.}(2021)]%
        {Park2021}
\bibfield{author}{\bibinfo{person}{M~S Park}, \bibinfo{person}{H Son}, \bibinfo{person}{C Hyun}, {and} \bibinfo{person}{H~J Hwang}.} \bibinfo{year}{2021}\natexlab{}.
\newblock \showarticletitle{{Explainability of Machine Learning Models for Bankruptcy Prediction}}.
\newblock \bibinfo{journal}{\emph{IEEE Access}}  \bibinfo{volume}{9} (\bibinfo{year}{2021}), \bibinfo{pages}{124887--124899}.
\newblock
\urldef\tempurl%
\url{https://doi.org/10.1109/ACCESS.2021.3110270}
\showDOI{\tempurl}


\bibitem[Peng et~al\mbox{.}(2021)]%
        {Peng2021}
\bibfield{author}{\bibinfo{person}{J Peng}, \bibinfo{person}{K Zou}, \bibinfo{person}{M Zhou}, {and} others.} \bibinfo{year}{2021}\natexlab{}.
\newblock \showarticletitle{{An Explainable Artificial Intelligence Framework for the Deterioration Risk Prediction of Hepatitis Patients}}.
\newblock \bibinfo{journal}{\emph{Journal of Medical Systems}} \bibinfo{volume}{45}, \bibinfo{number}{5} (\bibinfo{year}{2021}), \bibinfo{pages}{61}.
\newblock
\urldef\tempurl%
\url{https://doi.org/10.1007/s10916-021-01736-5}
\showDOI{\tempurl}


\bibitem[Phelps and Woolford(2021)]%
        {Phelps2021}
\bibfield{author}{\bibinfo{person}{Nathan Phelps} {and} \bibinfo{person}{Douglas~G Woolford}.} \bibinfo{year}{2021}\natexlab{}.
\newblock \showarticletitle{{Comparing calibrated statistical and machine learning methods for wildland fire occurrence prediction: a case study of human-caused fires in Lac La Biche, Alberta, Canada}}.
\newblock \bibinfo{journal}{\emph{International Journal of Wildland Fire}} \bibinfo{volume}{30}, \bibinfo{number}{11} (\bibinfo{year}{2021}), \bibinfo{pages}{850--870}.
\newblock
\urldef\tempurl%
\url{https://doi.org/10.1071/WF20139}
\showURL{%
\tempurl}


\bibitem[Piccialli et~al\mbox{.}(2021)]%
        {Piccialli2021}
\bibfield{author}{\bibinfo{person}{Francesco Piccialli}, \bibinfo{person}{Vittorio~Di Somma}, \bibinfo{person}{Fabio Giampaolo}, {and} others.} \bibinfo{year}{2021}\natexlab{}.
\newblock \showarticletitle{{A survey on deep learning in medicine: Why, how and when?}}
\newblock \bibinfo{journal}{\emph{Information Fusion}}  \bibinfo{volume}{66} (\bibinfo{year}{2021}), \bibinfo{pages}{111--137}.
\newblock
\showISSN{1566-2535}
\urldef\tempurl%
\url{https://doi.org/10.1016/j.inffus.2020.09.006}
\showDOI{\tempurl}


\bibitem[Pradhan et~al\mbox{.}(2022)]%
        {Pradhan2022}
\bibfield{author}{\bibinfo{person}{Romila Pradhan}, \bibinfo{person}{Jiongli Zhu}, \bibinfo{person}{Boris Glavic}, {and} \bibinfo{person}{Babak Salimi}.} \bibinfo{year}{2022}\natexlab{}.
\newblock \showarticletitle{Interpretable Data-Based Explanations for Fairness Debugging}. In \bibinfo{booktitle}{\emph{Proceedings of the 2022 International Conference on Management of Data}} (Philadelphia, PA, USA) \emph{(\bibinfo{series}{SIGMOD '22})}. \bibinfo{publisher}{ACM}, \bibinfo{address}{New York, NY, USA}, \bibinfo{pages}{247–261}.
\newblock
\showISBNx{9781450392495}
\urldef\tempurl%
\url{https://doi.org/10.1145/3514221.3517886}
\showDOI{\tempurl}


\bibitem[Qiu et~al\mbox{.}(2022)]%
        {Qiu2022}
\bibfield{author}{\bibinfo{person}{Y Qiu}, \bibinfo{person}{J Wang}, \bibinfo{person}{Z Jin}, {and} others.} \bibinfo{year}{2022}\natexlab{}.
\newblock \showarticletitle{{Pose-guided matching based on deep learning for assessing quality of action on rehabilitation training}}.
\newblock \bibinfo{journal}{\emph{Biomedical Signal Processing and Control}}  \bibinfo{volume}{72} (\bibinfo{year}{2022}), \bibinfo{pages}{103323}.
\newblock
\urldef\tempurl%
\url{https://doi.org/10.1016/j.bspc.2021.103323}
\showDOI{\tempurl}


\bibitem[{Quan Tran} et~al\mbox{.}(2022)]%
        {QuanTran2022}
\bibfield{author}{\bibinfo{person}{V {Quan Tran}}, \bibinfo{person}{V {Quoc Dang}}, {and} \bibinfo{person}{L {Si Ho}}.} \bibinfo{year}{2022}\natexlab{}.
\newblock \showarticletitle{{Evaluating compressive strength of concrete made with recycled concrete aggregates using machine learning approach}}.
\newblock \bibinfo{journal}{\emph{Construction and Building Materials}}  \bibinfo{volume}{323} (\bibinfo{year}{2022}), \bibinfo{pages}{126578}.
\newblock
\urldef\tempurl%
\url{https://doi.org/10.1016/j.conbuildmat.2022.126578}
\showDOI{\tempurl}


\bibitem[Raji and Buolamwini(2022)]%
        {raji2022}
\bibfield{author}{\bibinfo{person}{Inioluwa~Deborah Raji} {and} \bibinfo{person}{Joy Buolamwini}.} \bibinfo{year}{2022}\natexlab{}.
\newblock \showarticletitle{Actionable Auditing Revisited: Investigating the Impact of Publicly Naming Biased Performance Results of Commercial AI Products}.
\newblock \bibinfo{journal}{\emph{Commun. ACM}} \bibinfo{volume}{66}, \bibinfo{number}{1} (\bibinfo{date}{dec} \bibinfo{year}{2022}), \bibinfo{pages}{101–108}.
\newblock
\showISSN{0001-0782}
\urldef\tempurl%
\url{https://doi.org/10.1145/3571151}
\showDOI{\tempurl}


\bibitem[Raza et~al\mbox{.}(2022)]%
        {Raza2022}
\bibfield{author}{\bibinfo{person}{Ali Raza}, \bibinfo{person}{Kim~Phuc Tran}, \bibinfo{person}{Ludovic Koehl}, {and} \bibinfo{person}{Shujun Li}.} \bibinfo{year}{2022}\natexlab{}.
\newblock \showarticletitle{{Designing ECG monitoring healthcare system with federated transfer learning and explainable AI}}.
\newblock \bibinfo{journal}{\emph{Knowledge-Based Systems}}  \bibinfo{volume}{236} (\bibinfo{year}{2022}), \bibinfo{pages}{107763}.
\newblock
\showISSN{0950-7051}
\urldef\tempurl%
\url{https://doi.org/10.1016/j.knosys.2021.107763}
\showDOI{\tempurl}


\bibitem[Ribeiro et~al\mbox{.}(2016)]%
        {lime}
\bibfield{author}{\bibinfo{person}{Marco~Tulio Ribeiro}, \bibinfo{person}{Sameer Singh}, {and} \bibinfo{person}{Carlos Guestrin}.} \bibinfo{year}{2016}\natexlab{}.
\newblock \showarticletitle{"Why Should {I} Trust You?": Explaining the Predictions of Any Classifier}. In \bibinfo{booktitle}{\emph{Proceedings of the 22nd {ACM} {SIGKDD} International Conference on Knowledge Discovery and Data Mining, August 13-17, 2016}}. \bibinfo{publisher}{ACM}, \bibinfo{address}{San Francisco, CA, USA}, \bibinfo{pages}{1135--1144}.
\newblock


\bibitem[Ribeiro et~al\mbox{.}(2018)]%
        {Ribeiro2018}
\bibfield{author}{\bibinfo{person}{Marco~Tulio Ribeiro}, \bibinfo{person}{Sameer Singh}, {and} \bibinfo{person}{Carlos Guestrin}.} \bibinfo{year}{2018}\natexlab{}.
\newblock \showarticletitle{{Anchors: High-precision model-agnostic explanations}}. In \bibinfo{booktitle}{\emph{Proceedings of the AAAI conference on artificial intelligence}}, Vol.~\bibinfo{volume}{32}. \bibinfo{publisher}{AAAI Press}, \bibinfo{address}{New Orleans, Lousiana, USA}, \bibinfo{pages}{1}.
\newblock
\showISBNx{2374-3468}


\bibitem[Rinta-Kahila et~al\mbox{.}(2022)]%
        {Rinta-Kahila2022}
\bibfield{author}{\bibinfo{person}{Tapani Rinta-Kahila}, \bibinfo{person}{Ida Someh}, \bibinfo{person}{Nicole Gillespie}, {and} others.} \bibinfo{year}{2022}\natexlab{}.
\newblock \showarticletitle{{Algorithmic decision-making and system destructiveness: A case of automatic debt recovery}}.
\newblock \bibinfo{journal}{\emph{European Journal of Information Systems}} \bibinfo{volume}{31}, \bibinfo{number}{3} (\bibinfo{date}{may} \bibinfo{year}{2022}), \bibinfo{pages}{313--338}.
\newblock
\showISSN{0960-085X}
\urldef\tempurl%
\url{https://doi.org/10.1080/0960085X.2021.1960905}
\showDOI{\tempurl}


\bibitem[Robert et~al\mbox{.}(2020)]%
        {Robert2020}
\bibfield{author}{\bibinfo{person}{Lionel~P Robert}, \bibinfo{person}{Casey Pierce}, \bibinfo{person}{Liz Marquis}, {and} others.} \bibinfo{year}{2020}\natexlab{}.
\newblock \showarticletitle{{Designing fair AI for managing employees in organizations: a review, critique, and design agenda}}.
\newblock \bibinfo{journal}{\emph{Human–Computer Interaction}} \bibinfo{volume}{35}, \bibinfo{number}{5-6} (\bibinfo{date}{nov} \bibinfo{year}{2020}), \bibinfo{pages}{545--575}.
\newblock
\showISSN{0737-0024}
\urldef\tempurl%
\url{https://doi.org/10.1080/07370024.2020.1735391}
\showDOI{\tempurl}


\bibitem[Sado et~al\mbox{.}(2023)]%
        {Sado2023}
\bibfield{author}{\bibinfo{person}{Fatai Sado}, \bibinfo{person}{Chu~Kiong Loo}, \bibinfo{person}{Wei~Shiung Liew}, {and} others.} \bibinfo{year}{2023}\natexlab{}.
\newblock \showarticletitle{Explainable Goal-Driven Agents and Robots - A Comprehensive Review}.
\newblock \bibinfo{journal}{\emph{ACM Comput. Surv.}} \bibinfo{volume}{55}, \bibinfo{number}{10}, Article \bibinfo{articleno}{211} (\bibinfo{date}{feb} \bibinfo{year}{2023}), \bibinfo{numpages}{41}~pages.
\newblock
\showISSN{0360-0300}
\urldef\tempurl%
\url{https://doi.org/10.1145/3564240}
\showDOI{\tempurl}


\bibitem[Saeed and Omlin(2023)]%
        {Saeed2023}
\bibfield{author}{\bibinfo{person}{Waddah Saeed} {and} \bibinfo{person}{Christian Omlin}.} \bibinfo{year}{2023}\natexlab{}.
\newblock \showarticletitle{{Explainable AI (XAI): A systematic meta-survey of current challenges and future opportunities}}.
\newblock \bibinfo{journal}{\emph{Knowledge-Based Systems}}  \bibinfo{volume}{263} (\bibinfo{year}{2023}), \bibinfo{pages}{110273}.
\newblock
\showISSN{0950-7051}
\urldef\tempurl%
\url{https://doi.org/10.1016/j.knosys.2023.110273}
\showDOI{\tempurl}


\bibitem[Safaei et~al\mbox{.}(2022)]%
        {Safaei2022}
\bibfield{author}{\bibinfo{person}{N Safaei}, \bibinfo{person}{B Safaei}, \bibinfo{person}{S Seyedekrami}, {and} others.} \bibinfo{year}{2022}\natexlab{}.
\newblock \showarticletitle{{E-CatBoost: An efficient machine learning framework for predicting ICU mortality using the eICU Collaborative Research Database}}.
\newblock \bibinfo{journal}{\emph{PLoS ONE}} \bibinfo{volume}{17}, \bibinfo{number}{5 May} (\bibinfo{year}{2022}), \bibinfo{pages}{0262895}.
\newblock
\urldef\tempurl%
\url{https://doi.org/10.1371/journal.pone.0262895}
\showDOI{\tempurl}


\bibitem[Saiz-Rubio and Rovira-M{\'{a}}s(2020)]%
        {Saiz-Rubio2020}
\bibfield{author}{\bibinfo{person}{V Saiz-Rubio} {and} \bibinfo{person}{F Rovira-M{\'{a}}s}.} \bibinfo{year}{2020}\natexlab{}.
\newblock \showarticletitle{{From smart farming towards agriculture 5.0: A review on crop data management}}.
\newblock \bibinfo{journal}{\emph{Agronomy}} \bibinfo{volume}{10}, \bibinfo{number}{2} (\bibinfo{year}{2020}).
\newblock
\urldef\tempurl%
\url{https://doi.org/10.3390/agronomy10020207}
\showDOI{\tempurl}


\bibitem[Saldanha et~al\mbox{.}(2022)]%
        {Saldanha2022}
\bibfield{author}{\bibinfo{person}{Oliver~Lester Saldanha}, \bibinfo{person}{Philip Quirke}, \bibinfo{person}{Nicholas~P West}, {and} others.} \bibinfo{year}{2022}\natexlab{}.
\newblock \showarticletitle{{Swarm learning for decentralized artificial intelligence in cancer histopathology}}.
\newblock \bibinfo{journal}{\emph{Nature Medicine}} \bibinfo{volume}{28}, \bibinfo{number}{6} (\bibinfo{year}{2022}), \bibinfo{pages}{1232--1239}.
\newblock
\showISSN{1546-170X}
\urldef\tempurl%
\url{https://doi.org/10.1038/s41591-022-01768-5}
\showDOI{\tempurl}


\bibitem[Samsung(2023)]%
        {Samsung2023}
\bibfield{author}{\bibinfo{person}{Samsung}.} \bibinfo{year}{2023}\natexlab{}.
\newblock \bibinfo{title}{{Samsung AI Principles}}.
\newblock
\newblock
\urldef\tempurl%
\url{https://www.samsung.com/latin_en/sustainability/digital-responsibility/ai-ethics/}
\showURL{%
\tempurl}


\bibitem[Selvaraju et~al\mbox{.}(2017)]%
        {Selvaraju2017}
\bibfield{author}{\bibinfo{person}{Ramprasaath~R Selvaraju}, \bibinfo{person}{Michael Cogswell}, \bibinfo{person}{Abhishek Das}, {and} others.} \bibinfo{year}{2017}\natexlab{}.
\newblock \showarticletitle{Grad-CAM: Visual Explanations From Deep Networks via Gradient-Based Localization}. In \bibinfo{booktitle}{\emph{Proceedings of the IEEE International Conference on Computer Vision (ICCV)}}. \bibinfo{publisher}{Curran Associates, Inc.}, \bibinfo{address}{Venice, Italy}, \bibinfo{pages}{618--626}.
\newblock


\bibitem[Sharma et~al\mbox{.}(2020)]%
        {SharmaCERTIFAI}
\bibfield{author}{\bibinfo{person}{Shubham Sharma}, \bibinfo{person}{Jette Henderson}, {and} \bibinfo{person}{Joydeep Ghosh}.} \bibinfo{year}{2020}\natexlab{}.
\newblock \showarticletitle{CERTIFAI: A Common Framework to Provide Explanations and Analyse the Fairness and Robustness of Black-Box Models}. In \bibinfo{booktitle}{\emph{Proceedings of the AAAI/ACM Conference on AI, Ethics, and Society}} (New York, NY, USA) \emph{(\bibinfo{series}{AIES '20})}. \bibinfo{publisher}{ACM}, \bibinfo{address}{New York, NY, USA}, \bibinfo{pages}{166–172}.
\newblock
\showISBNx{9781450371100}
\urldef\tempurl%
\url{https://doi.org/10.1145/3375627.3375812}
\showDOI{\tempurl}


\bibitem[Shin(2021)]%
        {Shin2021}
\bibfield{author}{\bibinfo{person}{Donghee Shin}.} \bibinfo{year}{2021}\natexlab{}.
\newblock \showarticletitle{{The effects of explainability and causability on perception, trust, and acceptance: Implications for explainable AI}}.
\newblock \bibinfo{journal}{\emph{International Journal of Human-Computer Studies}}  \bibinfo{volume}{146} (\bibinfo{year}{2021}), \bibinfo{pages}{102551}.
\newblock
\showISSN{1071-5819}
\urldef\tempurl%
\url{https://doi.org/10.1016/j.ijhcs.2020.102551}
\showDOI{\tempurl}


\bibitem[Singla et~al\mbox{.}(2023)]%
        {Singla2023}
\bibfield{author}{\bibinfo{person}{Sumedha Singla}, \bibinfo{person}{Motahhare Eslami}, \bibinfo{person}{Brian Pollack}, {and} others.} \bibinfo{year}{2023}\natexlab{}.
\newblock \showarticletitle{{Explaining the black-box smoothly—A counterfactual approach}}.
\newblock \bibinfo{journal}{\emph{Medical Image Analysis}}  \bibinfo{volume}{84} (\bibinfo{year}{2023}), \bibinfo{pages}{102721}.
\newblock
\showISSN{1361-8415}
\urldef\tempurl%
\url{https://doi.org/10.1016/j.media.2022.102721}
\showDOI{\tempurl}


\bibitem[Song et~al\mbox{.}(2021)]%
        {Song2021}
\bibfield{author}{\bibinfo{person}{X.-F. Song}, \bibinfo{person}{Y Zhang}, \bibinfo{person}{D.-W. Gong}, {and} \bibinfo{person}{X.-Y. Sun}.} \bibinfo{year}{2021}\natexlab{}.
\newblock \showarticletitle{{Feature selection using bare-bones particle swarm optimization with mutual information}}.
\newblock \bibinfo{journal}{\emph{Pattern Recognition}}  \bibinfo{volume}{112} (\bibinfo{year}{2021}), \bibinfo{pages}{107804}.
\newblock
\urldef\tempurl%
\url{https://doi.org/10.1016/j.patcog.2020.107804}
\showDOI{\tempurl}


\bibitem[{Stanford Vision Lab}(2020)]%
        {StanfordVisionLab2020}
\bibfield{author}{\bibinfo{person}{{Stanford Vision Lab}}.} \bibinfo{year}{2020}\natexlab{}.
\newblock \bibinfo{title}{{ImageNet}}.
\newblock
\newblock
\urldef\tempurl%
\url{https://www.image-net.org/}
\showURL{%
\tempurl}


\bibitem[Straw(2020)]%
        {Straw2020}
\bibfield{author}{\bibinfo{person}{Isabel Straw}.} \bibinfo{year}{2020}\natexlab{}.
\newblock \showarticletitle{{The automation of bias in medical Artificial Intelligence (AI): Decoding the past to create a better future}}.
\newblock \bibinfo{journal}{\emph{Artificial Intelligence in Medicine}}  \bibinfo{volume}{110} (\bibinfo{year}{2020}), \bibinfo{pages}{101965}.
\newblock
\showISSN{0933-3657}
\urldef\tempurl%
\url{https://doi.org/10.1016/j.artmed.2020.101965}
\showDOI{\tempurl}


\bibitem[Taghizadeh-Mehrjardi et~al\mbox{.}(2021)]%
        {Taghizadeh-Mehrjardi2021}
\bibfield{author}{\bibinfo{person}{R Taghizadeh-Mehrjardi}, \bibinfo{person}{N Hamzehpour}, \bibinfo{person}{M Hassanzadeh}, {and} others.} \bibinfo{year}{2021}\natexlab{}.
\newblock \showarticletitle{{Enhancing the accuracy of machine learning models using the super learner technique in digital soil mapping}}.
\newblock \bibinfo{journal}{\emph{Geoderma}}  \bibinfo{volume}{399} (\bibinfo{year}{2021}), \bibinfo{pages}{115108}.
\newblock
\urldef\tempurl%
\url{https://doi.org/10.1016/j.geoderma.2021.115108}
\showDOI{\tempurl}


\bibitem[Ti et~al\mbox{.}(2021)]%
        {Ti2021}
\bibfield{author}{\bibinfo{person}{Lianping Ti}, \bibinfo{person}{Anita Ho}, {and} \bibinfo{person}{Rod Knight}.} \bibinfo{year}{2021}\natexlab{}.
\newblock \showarticletitle{{Towards Equitable AI Interventions for People Who Use Drugs: Key Areas That Require Ethical Investment}}.
\newblock \bibinfo{journal}{\emph{Journal of Addiction Medicine}} \bibinfo{volume}{15}, \bibinfo{number}{2} (\bibinfo{year}{2021}).
\newblock
\showISSN{1932-0620}
\urldef\tempurl%
\url{https://journals.lww.com/journaladdictionmedicine/Fulltext/2021/04000/Towards_Equitable_AI_Interventions_for_People_Who.3.aspx}
\showURL{%
\tempurl}


\bibitem[Tilmes(2022)]%
        {Tilmes2022}
\bibfield{author}{\bibinfo{person}{Nicholas Tilmes}.} \bibinfo{year}{2022}\natexlab{}.
\newblock \showarticletitle{{Disability, fairness, and algorithmic bias in AI recruitment}}.
\newblock \bibinfo{journal}{\emph{Ethics and Information Technology}} \bibinfo{volume}{24}, \bibinfo{number}{2} (\bibinfo{year}{2022}), \bibinfo{pages}{21}.
\newblock
\showISSN{1572-8439}
\urldef\tempurl%
\url{https://doi.org/10.1007/s10676-022-09633-2}
\showDOI{\tempurl}


\bibitem[Tomlein et~al\mbox{.}(2021)]%
        {Tomlein2021}
\bibfield{author}{\bibinfo{person}{Matus Tomlein}, \bibinfo{person}{Branislav Pecher}, \bibinfo{person}{Jakub Simko}, {and} others.} \bibinfo{year}{2021}\natexlab{}.
\newblock \showarticletitle{An Audit of Misinformation Filter Bubbles on YouTube: Bubble Bursting and Recent Behavior Changes}. In \bibinfo{booktitle}{\emph{Proceedings of the 15th ACM Conference on Recommender Systems}} (Amsterdam, Netherlands) \emph{(\bibinfo{series}{RecSys '21})}. \bibinfo{publisher}{ACM}, \bibinfo{address}{New York, NY, USA}, \bibinfo{pages}{1–11}.
\newblock
\showISBNx{9781450384582}
\urldef\tempurl%
\url{https://doi.org/10.1145/3460231.3474241}
\showDOI{\tempurl}


\bibitem[Trocin et~al\mbox{.}(2021)]%
        {Trocin2021}
\bibfield{author}{\bibinfo{person}{Cristina Trocin}, \bibinfo{person}{Patrick Mikalef}, \bibinfo{person}{Zacharoula Papamitsiou}, {and} \bibinfo{person}{Kieran Conboy}.} \bibinfo{year}{2021}\natexlab{}.
\newblock \showarticletitle{{Responsible AI for Digital Health: a Synthesis and a Research Agenda}}.
\newblock \bibinfo{journal}{\emph{Information Systems Frontiers}}  \bibinfo{volume}{23} (\bibinfo{year}{2021}), \bibinfo{pages}{1--19}.
\newblock
Issue 3.
\showISSN{1572-9419}
\urldef\tempurl%
\url{https://doi.org/10.1007/s10796-021-10146-4}
\showDOI{\tempurl}


\bibitem[Tubishat et~al\mbox{.}(2021)]%
        {Tubishat2021}
\bibfield{author}{\bibinfo{person}{M Tubishat}, \bibinfo{person}{S Ja'afar}, \bibinfo{person}{M Alswaitti}, {and} others.} \bibinfo{year}{2021}\natexlab{}.
\newblock \showarticletitle{{Dynamic Salp swarm algorithm for feature selection}}.
\newblock \bibinfo{journal}{\emph{Expert Systems with Applications}}  \bibinfo{volume}{164} (\bibinfo{year}{2021}), \bibinfo{pages}{113873}.
\newblock
\urldef\tempurl%
\url{https://doi.org/10.1016/j.eswa.2020.113873}
\showDOI{\tempurl}


\bibitem[Uddin et~al\mbox{.}(2022a)]%
        {Uddin2022depression}
\bibfield{author}{\bibinfo{person}{Md~Zia Uddin}, \bibinfo{person}{Kim~Kristoffer Dysthe}, \bibinfo{person}{Asbj{\o}rn F{\o}lstad}, {and} \bibinfo{person}{Petter~Bae Brandtzaeg}.} \bibinfo{year}{2022}\natexlab{a}.
\newblock \showarticletitle{{Deep learning for prediction of depressive symptoms in a large textual dataset}}.
\newblock \bibinfo{journal}{\emph{Neural Computing and Applications}} \bibinfo{volume}{34}, \bibinfo{number}{1} (\bibinfo{year}{2022}), \bibinfo{pages}{721--744}.
\newblock
\showISSN{1433-3058}
\urldef\tempurl%
\url{https://doi.org/10.1007/s00521-021-06426-4}
\showDOI{\tempurl}


\bibitem[Uddin and Soylu(2021)]%
        {Uddin2021}
\bibfield{author}{\bibinfo{person}{Md~Zia Uddin} {and} \bibinfo{person}{Ahmet Soylu}.} \bibinfo{year}{2021}\natexlab{}.
\newblock \showarticletitle{{Human activity recognition using wearable sensors, discriminant analysis, and long short-term memory-based neural structured learning}}.
\newblock \bibinfo{journal}{\emph{Scientific Reports}} \bibinfo{volume}{11}, \bibinfo{number}{1} (\bibinfo{year}{2021}), \bibinfo{pages}{16455}.
\newblock
\showISSN{2045-2322}
\urldef\tempurl%
\url{https://doi.org/10.1038/s41598-021-95947-y}
\showDOI{\tempurl}


\bibitem[Uddin et~al\mbox{.}(2022b)]%
        {Uddin2022}
\bibfield{author}{\bibinfo{person}{Shahadat Uddin}, \bibinfo{person}{Ibtisham Haque}, \bibinfo{person}{Haohui Lu}, {and} others.} \bibinfo{year}{2022}\natexlab{b}.
\newblock \showarticletitle{{Comparative performance analysis of K-nearest neighbour (KNN) algorithm and its different variants for disease prediction}}.
\newblock \bibinfo{journal}{\emph{Scientific Reports}} \bibinfo{volume}{12}, \bibinfo{number}{1} (\bibinfo{year}{2022}), \bibinfo{pages}{1--11}.
\newblock
\showISSN{2045-2322}


\bibitem[Ullah et~al\mbox{.}(2022)]%
        {Ullah2022}
\bibfield{author}{\bibinfo{person}{I Ullah}, \bibinfo{person}{K Liu}, \bibinfo{person}{T Yamamoto}, {and} others.} \bibinfo{year}{2022}\natexlab{}.
\newblock \showarticletitle{{Prediction of electric vehicle charging duration time using ensemble machine learning algorithm and Shapley additive explanations}}.
\newblock \bibinfo{journal}{\emph{International Journal of Energy Research}} \bibinfo{volume}{46}, \bibinfo{number}{11} (\bibinfo{year}{2022}), \bibinfo{pages}{15211--15230}.
\newblock
\urldef\tempurl%
\url{https://doi.org/10.1002/er.8219}
\showDOI{\tempurl}


\bibitem[van~der Waa et~al\mbox{.}(2020)]%
        {Waa2020}
\bibfield{author}{\bibinfo{person}{Jasper van~der Waa}, \bibinfo{person}{Tjeerd Schoonderwoerd}, \bibinfo{person}{Jurriaan van Diggelen}, {and} \bibinfo{person}{Mark Neerincx}.} \bibinfo{year}{2020}\natexlab{}.
\newblock \showarticletitle{{Interpretable confidence measures for decision support systems}}.
\newblock \bibinfo{journal}{\emph{International Journal of Human-Computer Studies}}  \bibinfo{volume}{144} (\bibinfo{year}{2020}), \bibinfo{pages}{102493}.
\newblock
\showISSN{1071-5819}
\urldef\tempurl%
\url{https://doi.org/10.1016/j.ijhcs.2020.102493}
\showDOI{\tempurl}


\bibitem[van Dis et~al\mbox{.}(2023)]%
        {VanDis2023}
\bibfield{author}{\bibinfo{person}{Eva A~M van Dis}, \bibinfo{person}{Johan Bollen}, \bibinfo{person}{Willem Zuidema}, {and} others.} \bibinfo{year}{2023}\natexlab{}.
\newblock \showarticletitle{{ChatGPT: five priorities for research}}.
\newblock \bibinfo{journal}{\emph{Nature}} \bibinfo{volume}{614}, \bibinfo{number}{7947} (\bibinfo{year}{2023}), \bibinfo{pages}{224--226}.
\newblock
\showISSN{0028-0836}


\bibitem[Wakefield(2016)]%
        {Wakefield2016}
\bibfield{author}{\bibinfo{person}{Jane Wakefield}.} \bibinfo{year}{2016}\natexlab{}.
\newblock \bibinfo{title}{{Microsoft chatbot is taught to swear on Twitter}}.
\newblock
\newblock
\urldef\tempurl%
\url{https://www.bbc.com/news/technology-35890188}
\showURL{%
\tempurl}


\bibitem[Wang et~al\mbox{.}(2022)]%
        {Wang2022}
\bibfield{author}{\bibinfo{person}{D Wang}, \bibinfo{person}{S Thun{\'{e}}ll}, \bibinfo{person}{U Lindberg}, {and} others.} \bibinfo{year}{2022}\natexlab{}.
\newblock \showarticletitle{{Towards better process management in wastewater treatment plants: Process analytics based on SHAP values for tree-based machine learning methods}}.
\newblock \bibinfo{journal}{\emph{Journal of Environmental Management}}  \bibinfo{volume}{301} (\bibinfo{year}{2022}), \bibinfo{pages}{113941}.
\newblock
\urldef\tempurl%
\url{https://doi.org/10.1016/j.jenvman.2021.113941}
\showDOI{\tempurl}


\bibitem[Wang et~al\mbox{.}(2020)]%
        {Wang2020}
\bibfield{author}{\bibinfo{person}{Haofan Wang}, \bibinfo{person}{Zifan Wang}, \bibinfo{person}{Mengnan Du}, {and} others.} \bibinfo{year}{2020}\natexlab{}.
\newblock \showarticletitle{{Score-CAM: Score-weighted visual explanations for convolutional neural networks}}. In \bibinfo{booktitle}{\emph{Proceedings of the IEEE/CVF conference on computer vision and pattern recognition workshops}}. \bibinfo{publisher}{Curran Associates}, \bibinfo{address}{Nashville, USA}, \bibinfo{pages}{24--25}.
\newblock


\bibitem[Wang et~al\mbox{.}(2021)]%
        {Wang2021}
\bibfield{author}{\bibinfo{person}{R Wang}, \bibinfo{person}{J.-H. Kim}, {and} \bibinfo{person}{M.-H. Li}.} \bibinfo{year}{2021}\natexlab{}.
\newblock \showarticletitle{{Predicting stream water quality under different urban development pattern scenarios with an interpretable machine learning approach}}.
\newblock \bibinfo{journal}{\emph{Science of the Total Environment}}  \bibinfo{volume}{761} (\bibinfo{year}{2021}), \bibinfo{pages}{144057}.
\newblock
\urldef\tempurl%
\url{https://doi.org/10.1016/j.scitotenv.2020.144057}
\showDOI{\tempurl}


\bibitem[Wu et~al\mbox{.}(2022)]%
        {Wu2022}
\bibfield{author}{\bibinfo{person}{Dingming Wu}, \bibinfo{person}{Xiaolong Wang}, {and} \bibinfo{person}{Shaocong Wu}.} \bibinfo{year}{2022}\natexlab{}.
\newblock \showarticletitle{{Jointly modeling transfer learning of industrial chain information and deep learning for stock prediction}}.
\newblock \bibinfo{journal}{\emph{Expert Systems with Applications}}  \bibinfo{volume}{191} (\bibinfo{year}{2022}), \bibinfo{pages}{116257}.
\newblock
\showISSN{0957-4174}
\urldef\tempurl%
\url{https://doi.org/10.1016/j.eswa.2021.116257}
\showDOI{\tempurl}


\bibitem[Yam and Skorburg(2021)]%
        {Yam2021}
\bibfield{author}{\bibinfo{person}{Josephine Yam} {and} \bibinfo{person}{Joshua~August Skorburg}.} \bibinfo{year}{2021}\natexlab{}.
\newblock \showarticletitle{{From human resources to human rights: Impact assessments for hiring algorithms}}.
\newblock \bibinfo{journal}{\emph{Ethics and Information Technology}} \bibinfo{volume}{23}, \bibinfo{number}{4} (\bibinfo{year}{2021}), \bibinfo{pages}{611--623}.
\newblock
\showISSN{1572-8439}
\urldef\tempurl%
\url{https://doi.org/10.1007/s10676-021-09599-7}
\showDOI{\tempurl}


\bibitem[Yan et~al\mbox{.}(2023)]%
        {Yan2023}
\bibfield{author}{\bibinfo{person}{Anli Yan}, \bibinfo{person}{Teng Huang}, \bibinfo{person}{Lishan Ke}, {and} others.} \bibinfo{year}{2023}\natexlab{}.
\newblock \showarticletitle{{Explanation leaks: Explanation-guided model extraction attacks}}.
\newblock \bibinfo{journal}{\emph{Information Sciences}}  \bibinfo{volume}{632} (\bibinfo{year}{2023}), \bibinfo{pages}{269--284}.
\newblock
\showISSN{0020-0255}
\urldef\tempurl%
\url{https://doi.org/10.1016/j.ins.2023.03.020}
\showDOI{\tempurl}


\bibitem[Yeo et~al\mbox{.}(2023)]%
        {Yeo2023}
\bibfield{author}{\bibinfo{person}{Yee~Hui Yeo}, \bibinfo{person}{Jamil~S Samaan}, \bibinfo{person}{Wee~Han Ng}, {and} others.} \bibinfo{year}{2023}\natexlab{}.
\newblock \showarticletitle{{Assessing the performance of ChatGPT in answering questions regarding cirrhosis and hepatocellular carcinoma}}.
\newblock \bibinfo{journal}{\emph{medRxiv}} (\bibinfo{year}{2023}), \bibinfo{pages}{2002--2023}.
\newblock


\bibitem[Yesilada and Lewandowsky(2022)]%
        {Yesilada2022Systematic}
\bibfield{author}{\bibinfo{person}{Muhsin Yesilada} {and} \bibinfo{person}{Stephan Lewandowsky}.} \bibinfo{year}{2022}\natexlab{}.
\newblock \showarticletitle{{Systematic review: YouTube recommendations and problematic content}}.
\newblock \bibinfo{journal}{\emph{Internet Policy Review}} \bibinfo{volume}{11}, \bibinfo{number}{1} (\bibinfo{year}{2022}), \bibinfo{pages}{1--22}.
\newblock
\showISSN{2197-6775}
\urldef\tempurl%
\url{https://doi.org/10.14763/2022.1.1652}
\showDOI{\tempurl}


\bibitem[Yogarajan et~al\mbox{.}(2022)]%
        {Yogarajan2022}
\bibfield{author}{\bibinfo{person}{Vithya Yogarajan}, \bibinfo{person}{Gillian Dobbie}, \bibinfo{person}{Sharon Leitch}, {and} others.} \bibinfo{year}{2022}\natexlab{}.
\newblock \bibinfo{title}{{Data and model bias in artificial intelligence for healthcare applications in New Zealand }}.
\newblock
\newblock
\showISBNx{2624-9898}
\urldef\tempurl%
\url{https://www.frontiersin.org/articles/10.3389/fcomp.2022.1070493}
\showURL{%
\tempurl}


\bibitem[Yoo and Kang(2021)]%
        {Yoo2021}
\bibfield{author}{\bibinfo{person}{S Yoo} {and} \bibinfo{person}{N Kang}.} \bibinfo{year}{2021}\natexlab{}.
\newblock \showarticletitle{{Explainable artificial intelligence for manufacturing cost estimation and machining feature visualization}}.
\newblock \bibinfo{journal}{\emph{Expert Systems with Applications}}  \bibinfo{volume}{183} (\bibinfo{year}{2021}), \bibinfo{pages}{115430}.
\newblock
\urldef\tempurl%
\url{https://doi.org/10.1016/j.eswa.2021.115430}
\showDOI{\tempurl}


\bibitem[Zemel et~al\mbox{.}(2013)]%
        {Zemel2013}
\bibfield{author}{\bibinfo{person}{Rich Zemel}, \bibinfo{person}{Yu Wu}, \bibinfo{person}{Kevin Swersky}, {and} others.} \bibinfo{year}{2013}\natexlab{}.
\newblock \bibinfo{title}{{Learning Fair Representations}}.
\newblock , \bibinfo{numpages}{325--333}~pages.
\newblock
\urldef\tempurl%
\url{http://proceedings.mlr.press/v28/zemel13.pdf https://proceedings.mlr.press/v28/zemel13.html}
\showURL{%
\tempurl}


\bibitem[Zhan et~al\mbox{.}(2020)]%
        {Zhan2020}
\bibfield{author}{\bibinfo{person}{Chen Zhan}, \bibinfo{person}{Elizabeth Roughead}, \bibinfo{person}{Lin Liu}, {and} others.} \bibinfo{year}{2020}\natexlab{}.
\newblock \showarticletitle{{Detecting potential signals of adverse drug events from prescription data}}.
\newblock \bibinfo{journal}{\emph{Artificial Intelligence in Medicine}}  \bibinfo{volume}{104} (\bibinfo{year}{2020}), \bibinfo{pages}{101839}.
\newblock
\showISSN{0933-3657}
\urldef\tempurl%
\url{https://doi.org/10.1016/j.artmed.2020.101839}
\showDOI{\tempurl}


\bibitem[Zhang et~al\mbox{.}(2021)]%
        {Zhang2021}
\bibfield{author}{\bibinfo{person}{G Zhang}, \bibinfo{person}{M Wang}, {and} \bibinfo{person}{K Liu}.} \bibinfo{year}{2021}\natexlab{}.
\newblock \showarticletitle{{Deep neural networks for global wildfire susceptibility modelling}}.
\newblock \bibinfo{journal}{\emph{Ecological Indicators}}  \bibinfo{volume}{127} (\bibinfo{year}{2021}), \bibinfo{pages}{107735}.
\newblock
\urldef\tempurl%
\url{https://doi.org/10.1016/j.ecolind.2021.107735}
\showDOI{\tempurl}


\bibitem[Zhang et~al\mbox{.}(2022b)]%
        {Zhang2022cam}
\bibfield{author}{\bibinfo{person}{X Zhang}, \bibinfo{person}{L Han}, \bibinfo{person}{W Zhu}, {and} others.} \bibinfo{year}{2022}\natexlab{b}.
\newblock \showarticletitle{{An Explainable 3D Residual Self-Attention Deep Neural Network for Joint Atrophy Localization and Alzheimer's Disease Diagnosis Using Structural MRI}}.
\newblock \bibinfo{journal}{\emph{IEEE Journal of Biomedical and Health Informatics}} \bibinfo{volume}{26}, \bibinfo{number}{11} (\bibinfo{year}{2022}), \bibinfo{pages}{5289--5297}.
\newblock
\urldef\tempurl%
\url{https://doi.org/10.1109/JBHI.2021.3066832}
\showDOI{\tempurl}


\bibitem[Zhang et~al\mbox{.}(2022c)]%
        {Zhang2022}
\bibfield{author}{\bibinfo{person}{Y Zhang}, \bibinfo{person}{T Zhou}, \bibinfo{person}{W Wu}, {and} others.} \bibinfo{year}{2022}\natexlab{c}.
\newblock \showarticletitle{{Improving EEG Decoding via Clustering-Based Multitask Feature Learning}}.
\newblock \bibinfo{journal}{\emph{IEEE Transactions on Neural Networks and Learning Systems}} \bibinfo{volume}{33}, \bibinfo{number}{8} (\bibinfo{year}{2022}), \bibinfo{pages}{3587--3597}.
\newblock
\showISSN{2162-2388 VO - 33}
\urldef\tempurl%
\url{https://doi.org/10.1109/TNNLS.2021.3053576}
\showDOI{\tempurl}


\bibitem[Zhang et~al\mbox{.}(2022a)]%
        {Zhang2022Spam}
\bibfield{author}{\bibinfo{person}{Z Zhang}, \bibinfo{person}{E Damiani}, \bibinfo{person}{H~A Hamadi}, {and} others.} \bibinfo{year}{2022}\natexlab{a}.
\newblock \showarticletitle{{Explainable Artificial Intelligence to Detect Image Spam Using Convolutional Neural Network}}. In \bibinfo{booktitle}{\emph{2022 International Conference on Cyber Resilience (ICCR)}}. \bibinfo{publisher}{Curran Associates}, \bibinfo{address}{Dubai, UAE}, \bibinfo{pages}{1--5}.
\newblock
\showISBNx{VO -}
\urldef\tempurl%
\url{https://doi.org/10.1109/ICCR56254.2022.9995839}
\showDOI{\tempurl}


\bibitem[Zhao et~al\mbox{.}(2022)]%
        {Zhao2022}
\bibfield{author}{\bibinfo{person}{Huan Zhao}, \bibinfo{person}{Ruixue Wang}, \bibinfo{person}{Yaguo Lei}, {and} others.} \bibinfo{year}{2022}\natexlab{}.
\newblock \showarticletitle{{Severity level diagnosis of Parkinson's disease by ensemble K-nearest neighbor under imbalanced data}}.
\newblock \bibinfo{journal}{\emph{Expert Systems with Applications}}  \bibinfo{volume}{189} (\bibinfo{year}{2022}), \bibinfo{pages}{116113}.
\newblock
\showISSN{0957-4174}
\urldef\tempurl%
\url{https://doi.org/10.1016/j.eswa.2021.116113}
\showDOI{\tempurl}


\bibitem[Zhao et~al\mbox{.}(2021)]%
        {zhao2021}
\bibfield{author}{\bibinfo{person}{X. Zhao}, \bibinfo{person}{W. Zhang}, \bibinfo{person}{X. Xiao}, {and} \bibinfo{person}{B. Lim}.} \bibinfo{year}{2021}\natexlab{}.
\newblock \showarticletitle{Exploiting Explanations for Model Inversion Attacks}. In \bibinfo{booktitle}{\emph{2021 IEEE/CVF International Conference on Computer Vision (ICCV)}}. \bibinfo{publisher}{IEEE Computer Society}, \bibinfo{address}{Los Alamitos, CA, USA}, \bibinfo{pages}{662--672}.
\newblock
\urldef\tempurl%
\url{https://doi.org/10.1109/ICCV48922.2021.00072}
\showDOI{\tempurl}


\bibitem[Zheng et~al\mbox{.}(2022)]%
        {Zheng2022}
\bibfield{author}{\bibinfo{person}{Quan Zheng}, \bibinfo{person}{Ziwei Wang}, \bibinfo{person}{Jie Zhou}, {and} \bibinfo{person}{Jiwen Lu}.} \bibinfo{year}{2022}\natexlab{}.
\newblock \showarticletitle{{Shap-CAM: Visual Explanations for Convolutional Neural Networks Based on Shapley Value}}. In \bibinfo{booktitle}{\emph{Computer Vision–ECCV 2022: 17th European Conference}}. \bibinfo{publisher}{Springer}, \bibinfo{address}{Tel Aviv, Israel}, \bibinfo{pages}{459--474}.
\newblock


\bibitem[Zhi et~al\mbox{.}(2022)]%
        {Zhi2022}
\bibfield{author}{\bibinfo{person}{J Zhi}, \bibinfo{person}{X Cao}, \bibinfo{person}{Z Zhang}, {and} others.} \bibinfo{year}{2022}\natexlab{}.
\newblock \showarticletitle{{Identifying the determinants of crop yields in China since 1952 and its policy implications}}.
\newblock \bibinfo{journal}{\emph{Agricultural and Forest Meteorology}}  \bibinfo{volume}{327} (\bibinfo{year}{2022}), \bibinfo{pages}{109216}.
\newblock
\urldef\tempurl%
\url{https://doi.org/10.1016/j.agrformet.2022.109216}
\showDOI{\tempurl}


\bibitem[Zhu et~al\mbox{.}(2017)]%
        {Zhu2017}
\bibfield{author}{\bibinfo{person}{J~Y. Zhu}, \bibinfo{person}{T Park}, \bibinfo{person}{P Isola}, {and} \bibinfo{person}{A~A Efros}.} \bibinfo{year}{2017}\natexlab{}.
\newblock \showarticletitle{{Unpaired Image-to-Image Translation Using Cycle-Consistent Adversarial Networks}}. In \bibinfo{booktitle}{\emph{2017 IEEE International Conference on Computer Vision (ICCV)}}. \bibinfo{pages}{2242--2251}.
\newblock
\showISBNx{2380-7504 VO -}
\urldef\tempurl%
\url{https://doi.org/10.1109/ICCV.2017.244}
\showDOI{\tempurl}


\bibitem[Zhu et~al\mbox{.}(2022)]%
        {Zhu2022}
\bibfield{author}{\bibinfo{person}{M Zhu}, \bibinfo{person}{B Zang}, \bibinfo{person}{L Ding}, {and} others.} \bibinfo{year}{2022}\natexlab{}.
\newblock \showarticletitle{{LIME-Based Data Selection Method for SAR Images Generation Using GAN}}.
\newblock \bibinfo{journal}{\emph{Remote Sensing}} \bibinfo{volume}{14}, \bibinfo{number}{1} (\bibinfo{year}{2022}), \bibinfo{pages}{204}.
\newblock
\urldef\tempurl%
\url{https://doi.org/10.3390/rs14010204}
\showDOI{\tempurl}


\bibitem[Zou and Khern-am nuai(2022)]%
        {Zou2022}
\bibfield{author}{\bibinfo{person}{Leying Zou} {and} \bibinfo{person}{Warut Khern-am nuai}.} \bibinfo{year}{2022}\natexlab{}.
\newblock \showarticletitle{{AI and housing discrimination: the case of mortgage applications}}.
\newblock \bibinfo{journal}{\emph{AI and Ethics}} (\bibinfo{year}{2022}).
\newblock
\showISSN{2730-5961}
\urldef\tempurl%
\url{https://doi.org/10.1007/s43681-022-00234-9}
\showDOI{\tempurl}


\end{thebibliography}
\end{document}